\documentclass[review]{elsarticle}
\usepackage{setspace}
\usepackage{lineno,hyperref}
\usepackage{graphicx}
\graphicspath{{figures/}} 
\usepackage{xcolor,colortbl}
\usepackage{booktabs}
\usepackage{subcaption}
\usepackage{amsmath,amssymb,mathtools}
\usepackage{multirow}
\usepackage{arydshln}
\usepackage{tabularx, makecell}
\usepackage{xspace}
\usepackage{stackrel}

\def\eg{\emph{e.g.}}

\def\ie{\emph{i.e.}}

\def\etc{\emph{etc.}}

\def\etal{\emph{et al.}}
\def\iid{i.i.d.}

\makeatletter
\def\ps@pprintTitle{%
  \let\@oddhead\@empty
  \let\@evenhead\@empty
  \def\@oddfoot{\reset@font\hfil\thepage\hfil}
  \let\@evenfoot\@oddfoot
}
\makeatother

\DeclareMathOperator*{\argmax}{arg\,max}

\makeatletter
\def\ps@pprintTitle{%
 \let\@oddhead\@empty
 \let\@evenhead\@empty
 \def\@oddfoot{\centerline{\thepage}}%
 \let\@evenfoot\@oddfoot}
\makeatother

\begin{document}

\begin{frontmatter}
\title{Continual Coarse-to-Fine Domain Adaptation in Semantic Segmentation}
\author{Donald Shenaj}
\author{Francesco Barbato}
\author{Umberto Michieli}
\author{Pietro Zanuttigh}
\address{Department of Information Engineering, University of Padova}
\address{\{donald.shenaj, francesco.barbato, umberto.michieli, zanuttigh\}@dei.unipd.it}
\begin{abstract}
Deep neural networks are typically trained in a single shot for a specific task and data distribution, but 
in real world settings both the task and the domain of application can change. 
The problem becomes even more challenging 
in dense predictive tasks, such as semantic segmentation, and furthermore most approaches tackle the two problems  separately.
In this paper we introduce the novel task of coarse-to-fine learning of semantic segmentation architectures in presence of domain shift. We consider subsequent learning stages  progressively  refining the task at the semantic level; \ie, the \textit{finer} set of semantic labels at each learning step is hierarchically derived from the \textit{coarser} set of the previous step.
We propose a new approach (CCDA) to tackle this scenario. 
First, we employ the maximum squares loss to align source and target domains and, at the same time, to balance the gradients between well-classified and harder samples. Second, we introduce a novel coarse-to-fine knowledge distillation constraint to transfer network capabilities acquired on a coarser set of labels to a set of finer labels.
Finally, we design a coarse-to-fine weight initialization rule to spread the importance from each coarse class to the respective finer classes.
To evaluate our approach, we design two benchmarks where source knowledge is extracted from the GTA5 dataset and it is transferred to either the Cityscapes or the IDD datasets, and we show how it outperforms the main competitors. 
\end{abstract}

\end{frontmatter}

\section{Introduction}

A fundamental aspect for any intelligent system is the ability to understand the surrounding environment. Semantic segmentation, \ie,  the dense (pixel-level) classification of an input image, is a key tool to achieve this target. 
Deep Convolutional Neural Networks, starting with the seminal work on FCNs by Long \etal~\cite{long2015} achieve impressive performance on this challenging task. 
Such architectures are very effective when trained on \iid\ samples coming from a single domain, however, they suffer from an important shortcoming: they tend to fail when even a small distribution shift occurs over the input domain distribution or over the output label space. Particularly, semantic segmentation solutions trained on a source domain show clear signs of \textit{domain shift} when tested on a target test domain, limiting the performance \cite{toldo2020unsupervised}. A similar effect is found when training architectures in incremental steps, where \textit{catastrophic forgetting} leads to a clear performance degradation \cite{michieli2019incremental}. Such effects impair significantly the employment of deep neural networks in real-world settings, where they are expected to adapt their knowledge to the specific encountered environment, while at the same time maintaining the ability of achieving subsequent levels of understanding of the scene incrementally. 
To address these undesirable effects, a multitude of Unsupervised Domain Adaptation (UDA) and Continual Learning (CL) approaches have been recently investigated to tackle domain shift and catastrophic forgetting, respectively.

\begin{figure}[t]
    \centering
    \includegraphics[width=\textwidth]{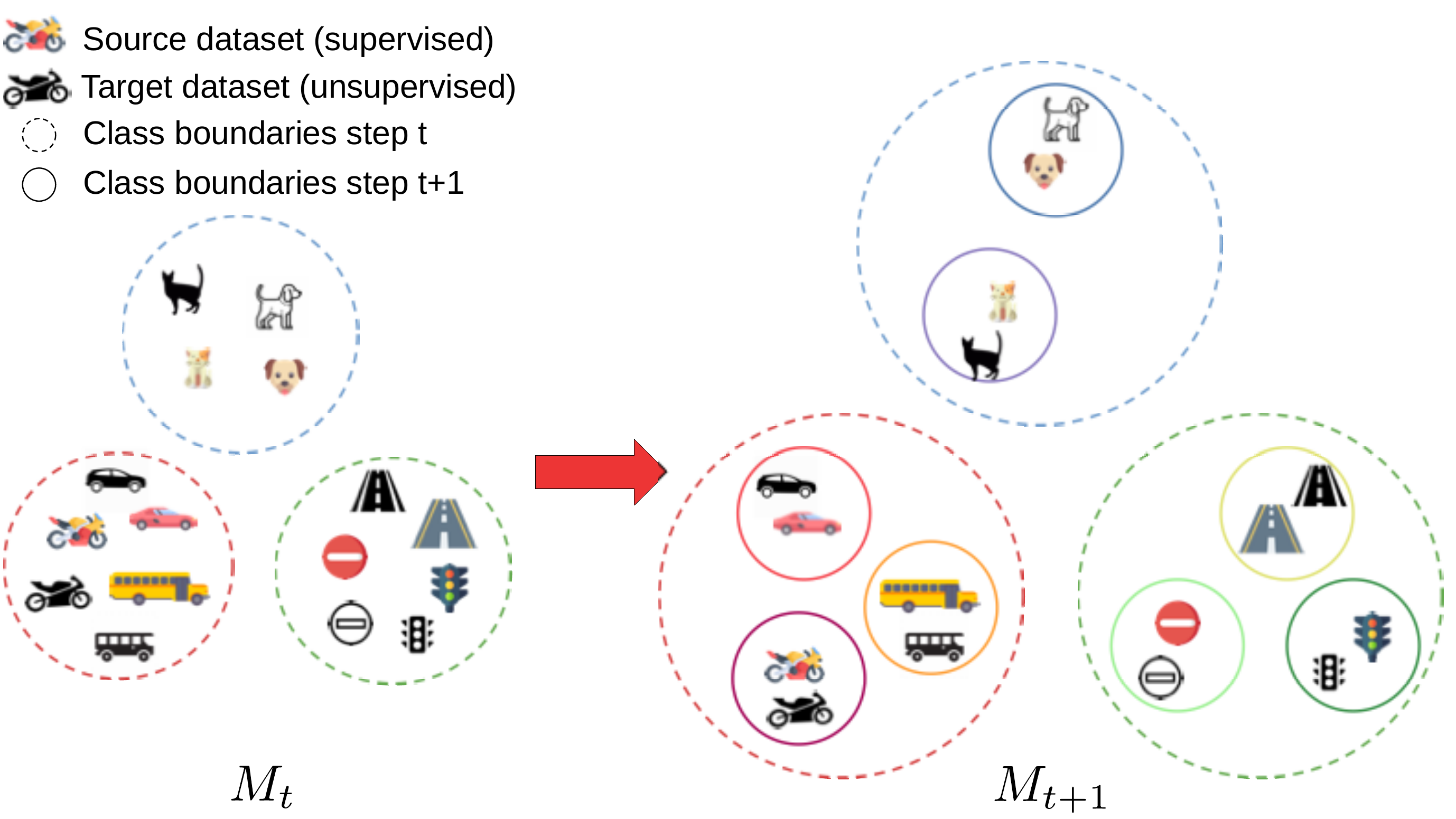}
    \caption{In semantic coarse-to-fine learning we train a model in subsequent steps to recognize incremental sets of finer classes over time, that are derived hierarchically from previous coarse ones. Similarly to UDA, labeled samples come from a source domain, which is different from the (testing) target domain, whose labels are unknown at training time.}
    \label{fig:c2f_sets}
\end{figure}

In this work, we propose a joint continual and domain adaptive approach, placing an additional brick towards the development of  deep learning architectures which can be deployed with intelligent agents. We consider coarse-to-fine learning, where the label sets are incrementally refined and classes are hierarchically derived from the previous ones, as it often occurs during the manual labeling by humans. A first, rough, segmentation comprised of few coarse classes is provided; then, it is further refined by splitting the coarse classes into finer categories. Moreover, similarly to continual learning, we assume that the set of classes to be recognized can change during training, as it is often unfeasible to retrain a segmentation architecture on the whole class set each time a new fine class is introduced. This incremental behavior is applied by masking classes that have reached their maximum detail level in previous class set refinements, forcing the architecture to preserve previously acquired knowledge. Finally, to better mimic the deployment of an actual intelligent system, we evaluate our setup in the presence of domain shift. An overview of the considered task is shown in Figure~\ref{fig:c2f_sets}, where we appreciate how the coarser classes are refined into finer categories by hierarchically splitting them, while adapting the network at both steps. 
The considered task, together with the mathematical notation needed to describe it, is reported in detail in Section~\ref{sec:task}.
To tackle this task we introduce a novel approach named CCDA (Continual Coarse-to-fine Domain Adaptation), detailed in Section~\ref{sec:method}. In CCDA we employ a custom-designed coarse-to-fine knowledge distillation loss (see Section~\ref{subsec:c2fkd}) and an unbiased coarse-to-fine weight initialization technique (see Section~\ref{subsec:c2fuwi}) together with an entropy minimization-based approach for unsupervised domain adaptation \cite{Chen2019} which, can regularize the gradient evolution and better preserve less common classes during the adaptation procedure (see Section~\ref{subsec:uda_method}). Importantly, we apply the domain adaptation approach in all the incremental training steps, obtaining a network with improved generalization ability. This is due to the effect of self-supervision in earlier steps, which is very effective when a restricted number of classes  
is considered.

Given the novelty of the task, in order to evaluate our approach we design two new benchmarks, derived from common semantic segmentation synthetic-to-real unsupervised domain adaptation tasks. In particular, we reframed the GTA5 \cite{Richter2016}, Cityscapes \cite{Cordts2016} and IDD \cite{varma2019idd} datasets into class-hierarchical coarse-to-fine labeling datasets. These benchmarks (\textit{GTA5}$\rightarrow$\textit{Cityscapes} and {\textit{GTA5}$\rightarrow$~\textit{IDD}}) are then employed to evaluate our approach and compare it with other competing methods. The implementation details are discussed in Section \ref{sec:impl}, while the quantitative and qualitative results obtained are reported in Section~\ref{sec:results}.

\section{Related Work}
\textit{\textbf{Semantic Segmentation}} 
has gained wide popularity over the last few years.
A multitude of Convolutional Neural Network-based (CNN) segmentation architectures has been proposed for this task starting from FCN~\cite{long2015} to more recent approaches like SegNet \cite{badrinarayanan2017segnet}, PSPNet~\cite{zhao2017} and DeepLab~\cite{chen2017rethinking,chen2018deeplab,chen2018encoder}.
These approaches are supported by a fast-growing number of datasets such as Cityscapes~\cite{Cordts2016}, IDD~\cite{varma2019idd}, GTA5~\cite{Richter2016}, \etc\ 
Such architectures are very effective when trained on i.i.d.\ samples coming from a single domain distribution, however they tend to fail when distribution shifts occur over the input domain distribution or the output label space.

\textit{\textbf{Coarse-to-Fine Learning}: }
hierarchical composition of semantic representations has been explored in few previous work for part-based regularization \cite{li2021prototypical,yang2021part} and coarse-to-fine approaches where coarse-level classes are refined into finer categories \cite{mel2020incremental,stretcu2020coarse,stretcu2021coarse,michieli2020gmnet}. In particular, coarse-to-fine learning  in semantic segmentation can be addressed by resorting either to a direct concatenation of features \cite{mel2020incremental} or using a semantic embedding network to transfer object-level predictions during part-level training \cite{michieli2020gmnet}.
However, approaches for this task all assume that the domain distribution remains unaltered at subsequent refinement stages and they are comprised of one single refinement stage.

\textit{\textbf{Continual Learning (CL)}} is a more mature field where previous knowledge acquired on a subset of data is further expanded to learn new tasks (\eg, class labels) \cite{parisi2019continual,lesort2019continual,michieli2021unsupervised}. Continual learning has recently been investigated also in semantic segmentation \cite{michieli2019incremental,klingner2020class,michieli2021continual,cermelli2020modeling,douillard2021plop,maracani2021recall}.
The most popular strategy to retain knowledge from the previous learning step is the usage of knowledge distillation constraints at either the output \cite{michieli2019incremental,michieli2021continual,cermelli2020modeling,klingner2020class} or feature level \cite{michieli2019incremental,douillard2021plop}. Another promising research direction involves feature-level regularization to increase separation among features from different classes \cite{michieli2021continual}. Additionally, replay samples of past classes can also be either generated via a GAN or downloaded from the Web \cite{maracani2021recall}.
However, all these approaches do not consider classes hierarchically derived from macro-classes.
The strongest similarity with most of these works lies in the special \textit{background} class from which the novel classes are extracted in these approaches. 
Indeed, the \textit{background} at a certain learning step could contain future classes, therefore at subsequent steps classes are derived from this unique macro-class.
However, all these approaches deal only with the semantic shift of one class (\ie, the background) and they do not tackle data domain shift (\ie, source and target domains).

\textit{\textbf{Unsupervised Domain Adaptation}} is a more challenging variant of domain adaptation \cite{csurka2017domain, toldo2020unsupervised},  where no labeled samples coming from the target domain are exploited for the adaptation procedure.
These techniques deal with the reduction of the domain shift effect and, therefore, lead to a performance improvement in the target domain. While originally proposed to tackle such problems in other settings, like image classification, the rise of segmentation architectures made the transposition of adaptation techniques to such task a very desirable objective.
Following the encoder-decoder structure of common semantic segmentation networks, a multitude of UDA techniques has been proposed acting on different network levels: \ie, at the \textit{input}, \textit{feature} and \textit{output} levels \cite{toldo2020unsupervised}.

Works at input level usually resort to image-to-image translation \cite{chen2019crdoco,hoffman2018,hoffman2016,MurezKKRK18,toldo2020,pizzati2020domain}, or, more recently, to image-inpainting \cite{tranheden2021dacs}. The former aims to reduce the domain shift by acting directly on the input images, either via style transfer or by generative architectures, while the latter exploits carefully designed cross-domain samples. Particularly, in \cite{tranheden2021dacs} the network is adapted via a mixture of self-training and image inpainting. 

Feature-level techniques reduce the domain shift by aligning the statistical distributions of the two domains exploiting the encoder-decoder latent representations. Recent works in this category \cite{toldo2020clustering, barbato2021latent, barbato2021road} investigated various ways towards the goal, such as self-supervised and unsupervised orthogonal embeddings, source-target alignment objectives, prototype-driven clustering, vector sparsity and norm matching.

Finally, works at the output level fall into two major categories: entropy minimization and self-supervised training, both via direct approaches \cite{zou2018, Zou2019} or by exploiting curriculum-learning techniques \cite{zhang2017, zhang2020curriculum}. In \cite{zou2018, Zou2019} the network is forced to produce confident predictions in the target domain, similarly to the behavior on the source domain, by acting on the logits produced by the segmentation architecture. In \cite{zhang2017, zhang2020curriculum} the network predictions on the target domain reinforce the training and prediction confidence. 
Other recent and high-performing techniques, such as \cite{cardace2021shallow}, use a mixture of all the families of techniques. Here, the authors exploit image-inpainting, feature-level adaptation, self-supervised training and a modified segmentation architecture to surpass previous state-of-the-art approaches.


\section{Task Formulation}
\label{sec:task}
In this section we provide a detailed overview of the proposed task along with the mathematical notation needed to describe it.
We start by introducing the employed notation in Section~\ref{subsec:prob}, then
in Section~\ref{subsec:dsets} we report an overview of the considered benchmarks describing their adaptation to our task; and in Section~\ref{subsec:splits} we detail the hierarchical semantic class splits identified in such benchmarks.

\subsection{Problem Setup}
\label{subsec:prob}
In this work we consider semantic segmentation, which performs a pixel-level labeling of input images.
Let us define the (input) image space as $\mathcal{X} \subset \mathbb{R}^{H \times W \times 3}$, which represents the set of RGB images of width $W$ and height $H$. Together with the input we identify the corresponding pixel-level (output) labels, which belong to the space $\mathcal{Y}_t \subset \mathcal{C}_t^{H \times W}$, where $t$ is a task index accounting for the changes in the label set $\mathcal{C}$ over time.
The semantic segmentation task can be expressed as the problem of finding a map $M: \mathcal{X} \ni \mathbf{X} \mapsto \mathbf{P}_t = M_t(\mathbf{X}) \in [0,1]^{H \times W \times |C_t|}$ such that $\hat{\mathbf{Y}} \coloneqq \argmax\limits_{c \in \mathcal{C}_t} {\mathbf{P}_t} = \mathbf{Y} \in \mathcal{Y}$. In recent semantic segmentation architectures the map $M$ takes the form of an encoder-decoder network, \ie, $M_t = D_t \circ E_t$. This allows to identify the latent features extracted by the encoder and used for the final predictions as $\mathbf{F} = E_t(\mathbf{X})$ for any image $\mathbf{X} \in \mathcal{X}$.
Indeed, as opposed to the standard supervised training where the whole class set is available from the beginning, in a continual setting the final class set $\mathcal{C} = \bigcup\limits_{t} {\mathcal{C}_t}$ is divided into subsets that are learned separately as tasks $t = \{0, 1, \dots\}$.
In a coarse-to-fine setting, we can make a further  separation: the class sets of each task can be divided as $\mathcal{C}_t = \mathcal{C}_t^c \cup \mathcal{C}_t^f$, where $\mathcal{C}_t^c$ contains the classes that will be split into finer ones in the future and $\mathcal{C}_t^f$ contains classes that have already reached the maximum level of refinement (\ie, they will not be split anymore). The split between  $\mathcal{C}_t^c$ and $\mathcal{C}_t^f$ is unknown at stage $t-1$, but it becomes available when the subsequent training stage begins.
At that time (task $t$), the coarse set of classes of the previous task $t-1$ are mapped to a set of derived finer classes with the map ${S_t: \mathcal{C}_{t-1} ^c \ni c \mapsto S_t (c) \subset \mathcal{C}_t}$. Therefore, at each step the last layer of the decoder $D_t$ must change in order to accommodate the new classes, as we will detail in Section~\ref{sec:method}.
Then, we can define the training set of task $t$ as $\mathcal{T}_t = \mathcal{T}_t^S \cup \mathcal{T}_t^T$, where $\mathcal{T}_t^S = \{(\mathbf{X}^S,Y^S) \stackrel{\mathcal{D}^S}{\sim}
(\mathcal{X} \times \mathcal{Y}_t)\}$ is the (supervised) source dataset, while $\mathcal{T}_t^T = \{\mathbf{X}^T \stackrel{\mathcal{D}^T}{\sim}  
\mathcal{X}\}$ is the (unsupervised) target one.
Note that, while the contents of $\mathcal{T}_t^S$ and $\mathcal{T}_t^T$ are 
relative to the same image space, the distributions according to which they are sampled varies (namely, $\mathcal{D}^S$ for the source domain distribution and $\mathcal{D}^T$ for the target one), leading to the presence of domain shift.
\subsection{Datasets}
\label{subsec:dsets}
For the evaluation of our strategy we use three different datasets, \ie, \textit{GTA5}, used as source dataset for supervised training and \textit{Cityscapes} and \textit{IDD}, used as target ones. This allows to build two synthetic-to-real semantic segmentation benchmarks: \textit{GTA5}$\rightarrow$\textit{Cityscapes} and \textit{GTA5}$\rightarrow$\textit{IDD}. In these benchmarks the segmentation architecture is trained in a supervised manner using samples exclusively from the \textit{GTA5}~\cite{Richter2016} dataset and its performance is evaluated on \textit{Cityscapes}~\cite{Cordts2016} and \textit{IDD}~\cite{varma2019idd}, respectively. 

The \textbf{GTA5}~\cite{Richter2016} dataset provides $25000$ synthetic images generated using the homonym video game, which is set in American cities. The images have varying resolutions close to $1920 \times 1080$, an aspect ratio of $16:9$ and are paired with semantic segmentation labels with the same class set as Cityscapes (total of $35$ classes, $19$ of which are used for training). 

The \textbf{Cityscapes}~\cite{Cordts2016} dataset, instead, provides $2500$ finely-labeled real images captured in multiple European cities and an additional set of $20000$ coarsely-labeled samples from the same cities. The images have resolution $2048\times1024$ (aspect ratio of $2:1$)  and are annotated using $35$ classes.  

Finally, the \textbf{IDD}~\cite{varma2019idd} dataset provides $10004$ finely-labeled samples captured in India. The images have a smaller resolution than the previous datasets ($1280\times720$, aspect ratio of $16:9$) and the labels are provided in a different class set ($36$ classes, $27$ of which are used in the training) leading to a different number of classes when performing closed-set domain adaptation: in particular the \emph{train} and \emph{terrain} classes are missing.

\begin{figure}[h!]
\centering
\includegraphics[width=\textwidth]{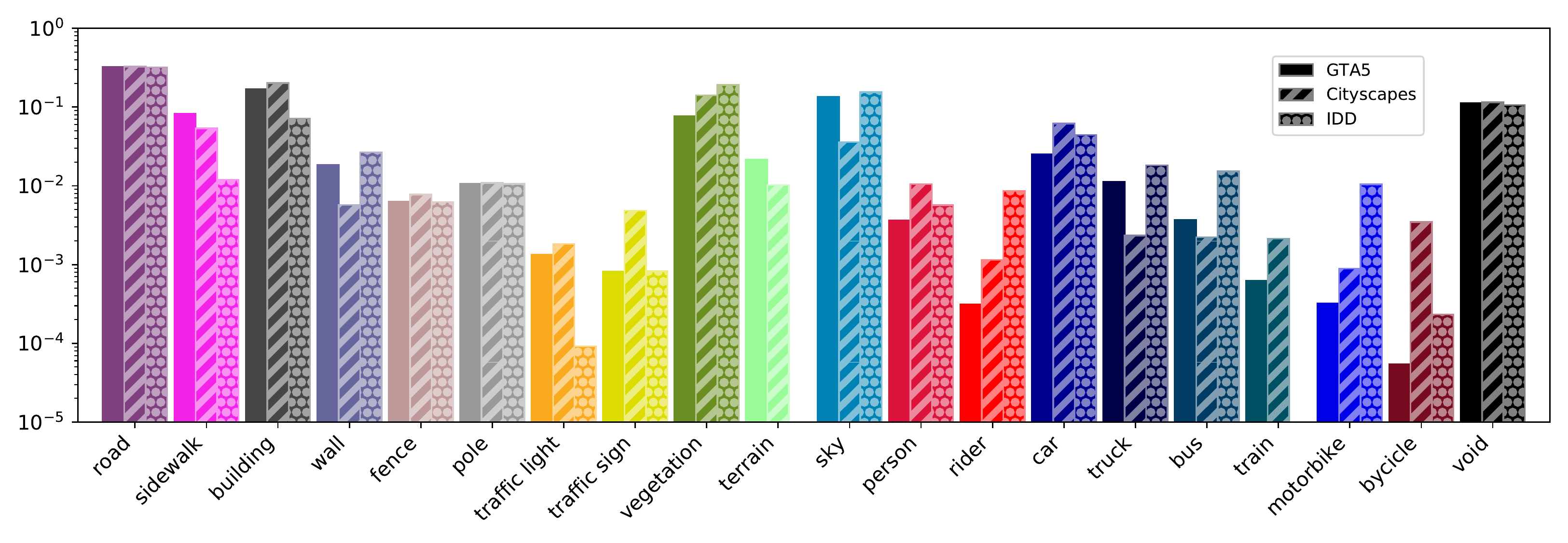}
\caption{Relative pixel-wise class frequencies in each considered dataset (the plot uses a logarithmic scale). GTA5 in solid colors, Cityscapes in dashed colors, IDD in dotted colors (\emph{train} and \emph{terrain} are not present in IDD).}
\label{fig:gen_dset_freqs}
\end{figure}
\begin{figure}[h!]
\centering
\begin{subfigure}{.3275\textwidth}
\centering
\textbf{GTA5\vphantom{y}}
\includegraphics[width=\textwidth, trim={0 3pt 0 3pt}, clip]{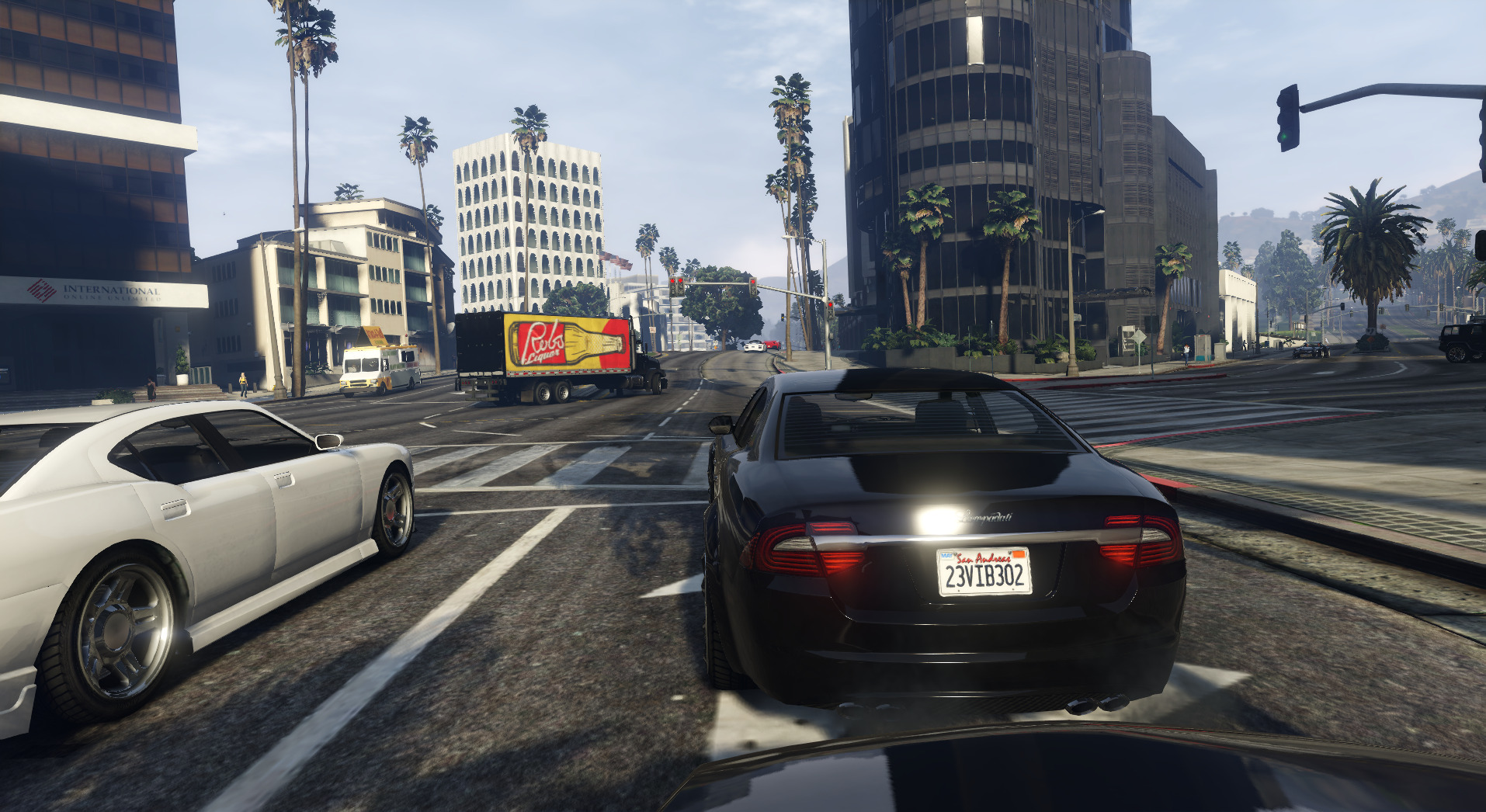}
\end{subfigure}%
\begin{subfigure}{.3165\textwidth}
\centering
\textbf{IDD\vphantom{y}}
\includegraphics[width=\textwidth]{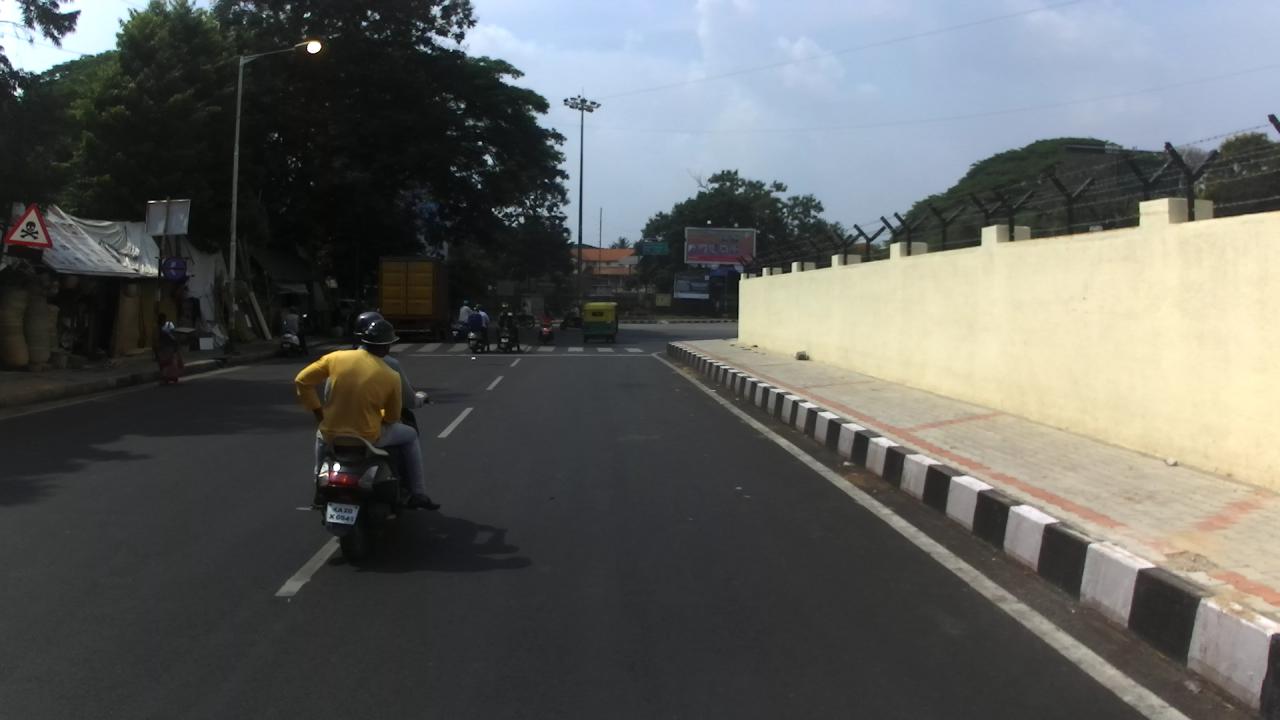}
\end{subfigure}%
\begin{subfigure}{.356\textwidth}
\centering
\textbf{Cityscapes}
\includegraphics[width=\textwidth]{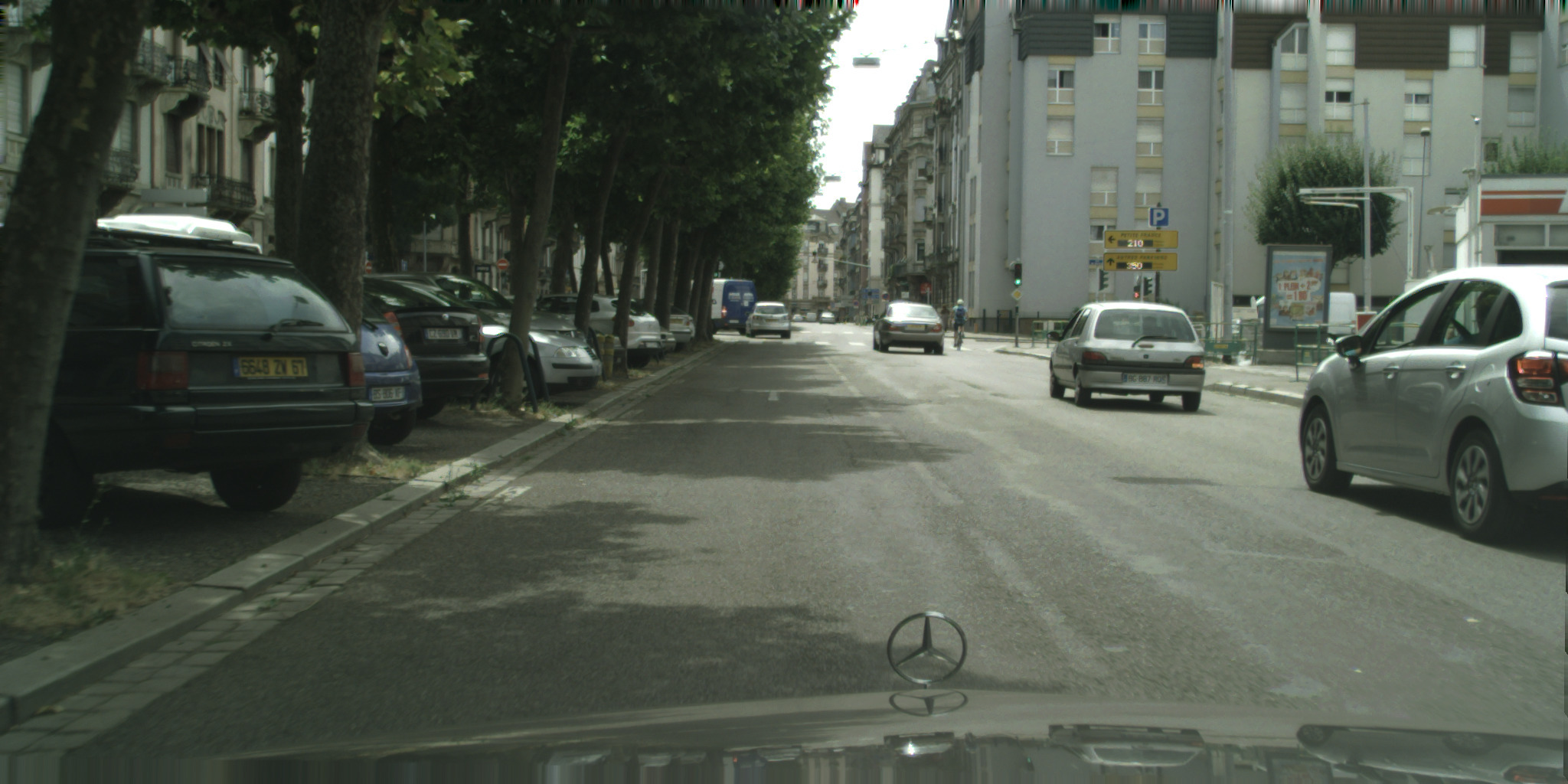}
\end{subfigure}
\begin{subfigure}{.3275\textwidth}
\includegraphics[width=\textwidth, trim={0 5pt 0 5pt}, clip]{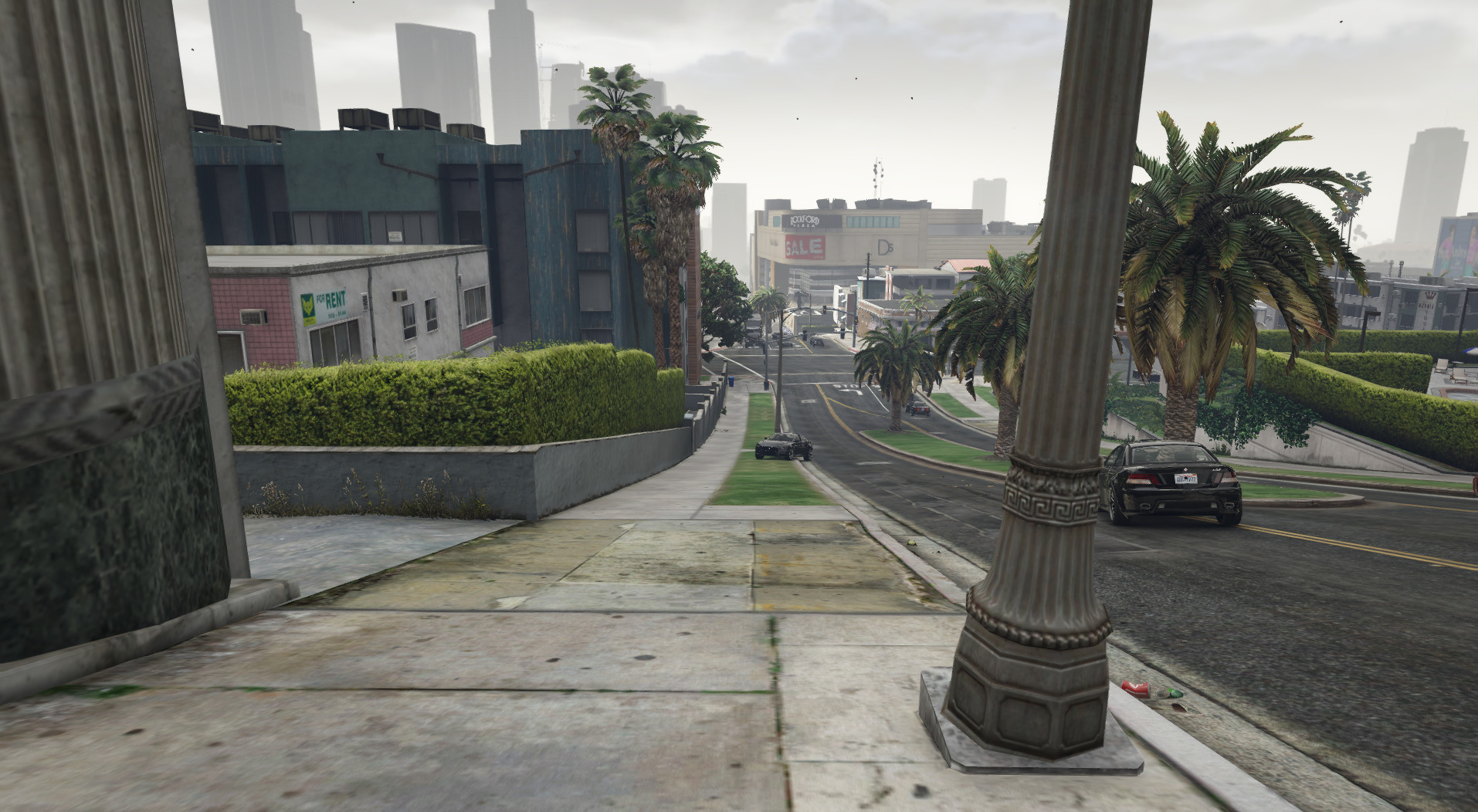}
\end{subfigure}%
\begin{subfigure}{.3165\textwidth}
\includegraphics[width=\textwidth]{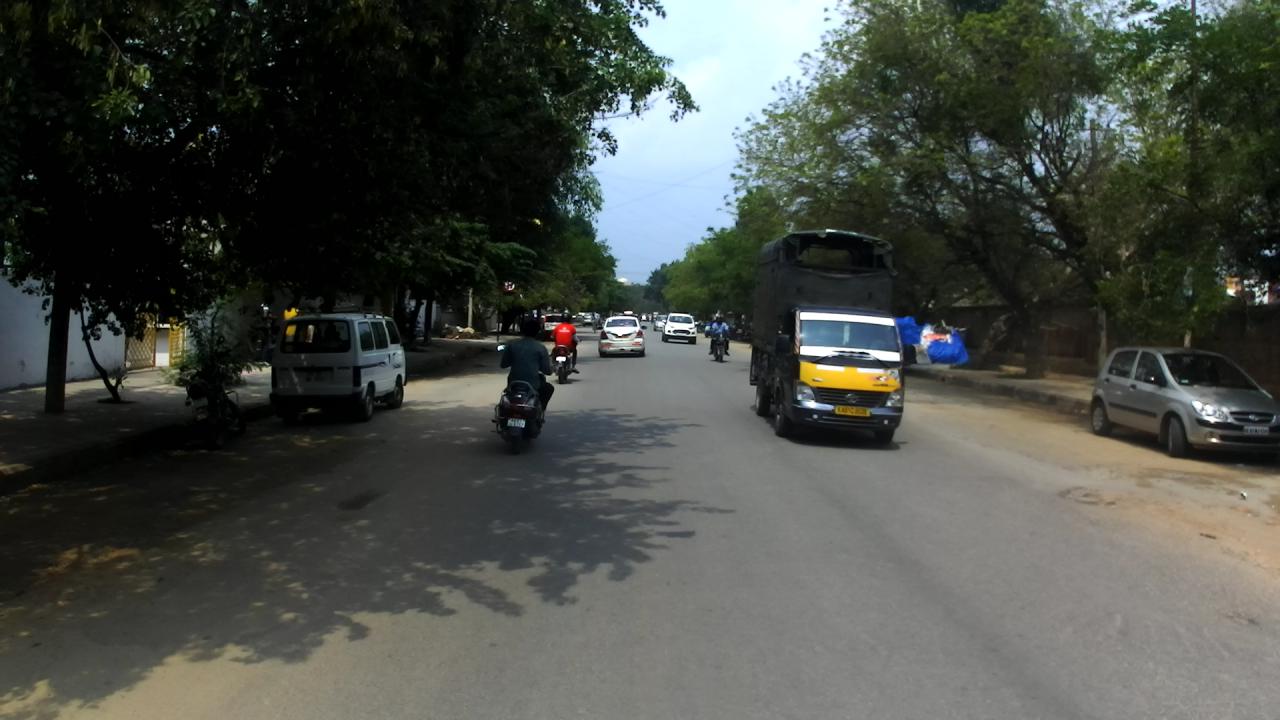}
\end{subfigure}%
\begin{subfigure}{.356\textwidth}
\includegraphics[width=\textwidth]{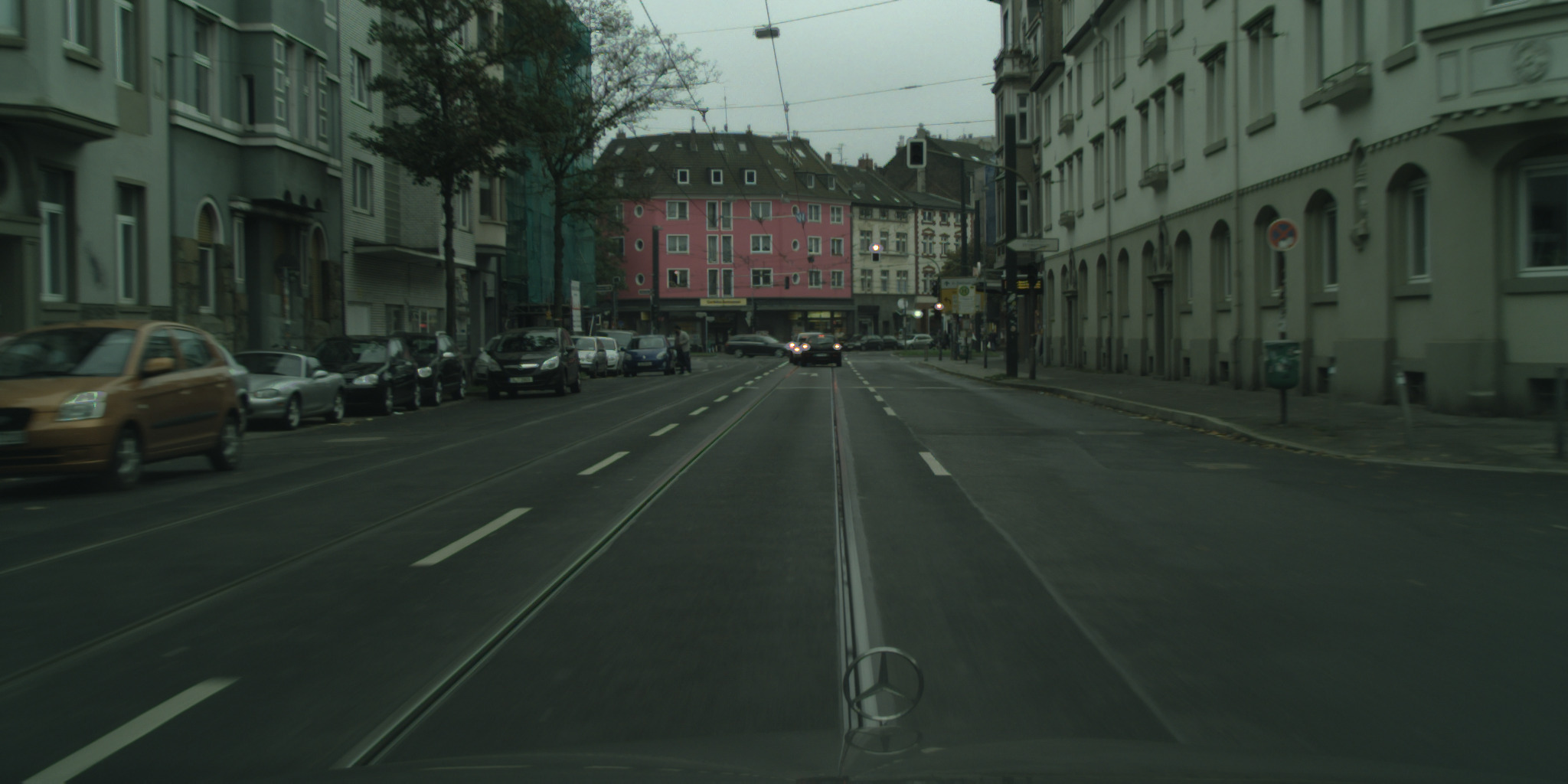}
\end{subfigure}
\caption{Samples from the three considered datasets highlight the large visual domain shift among them.}
\label{fig:data_samples}
\end{figure}

Autonomous driving datasets are inherently highly class-imbalanced due to the variable amount of pixels each class can have. We show the relative pixel class frequency of the three datasets in Figure \ref{fig:gen_dset_freqs}, where the significant distribution shift and the missing IDD classes are clearly visible.

Even more, datasets acquired in different geographical regions and experimental scenarios have profound domain distribution shifts. In Figure~\ref{fig:data_samples} we report a couple of samples from each dataset, noticing how domain shift is evident from the visual aspect of the images. The images from the GTA5 dataset have a very crisp look, typical of synthetic datasets, even when the weather is cloudy (second row of first column); while the images coming from the real datasets are slightly blurry, side effect of the motion blur and of the real camera optics, which cannot focus perfectly on the whole scene. Even more evident is the difference in coloring between the datasets: GTA5 offers visually-pleasing well-saturated colors (which are inherited from the game engine used to generate the samples) while in IDD (second column) the colors are much dimmer. Even worse is the Cityscapes dataset (third column), where the white point is unbalanced and the resulting samples appear greenish and grayish.

\subsection{Coarse-to-Fine Hierarchical Splits}
\label{subsec:splits}

\begin{figure}[t]
\centering
\includegraphics[width=\textwidth]{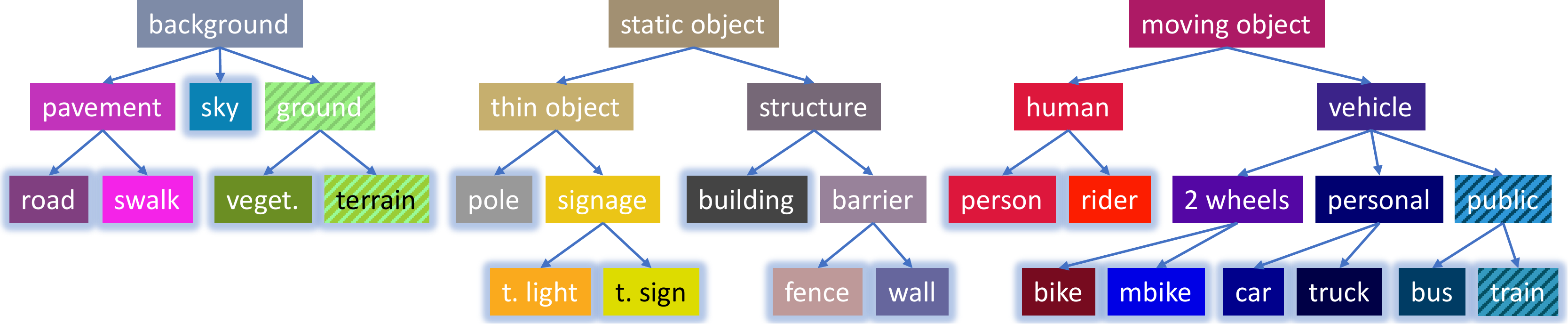}
\caption{Hierarchical label splits: each level of depth in the tree corresponds to a different learning step. At each step, the provided ground truth segmentation maps contain only the labels of the given step, while all the other pixels are unlabeled. Boxes highlighted in blue contain the original Cityscapes classes. Classes in dashed boxes are not present in the IDD dataset (\emph{terrain} and \emph{train} are missing leading to the collapse of \emph{ground} in \emph{vegetation} and of \emph{public transport} in \emph{bus} at the second step).}
\label{fig:label_splits}
\end{figure}

To validate our setup, we devise a way of extending the benchmarks into class-hierarchical coarse-to-fine datasets. In particular, we focus on the $19$ trainable classes of the Cityscapes dataset, developing a semantic grouping based both on the similarity of the classes and on common semantic concepts  (\ie, the \emph{vehicle} class can be naturally split into finer classes like \emph{two-wheels}, while simpler classes like \emph{sky} reach the maximum level of detail earlier and cannot be further split).
A graphical representation of our hierarchical split is reported in Figure~\ref{fig:label_splits}, where different depths on the tree represent different incremental steps during the training procedure. The boxes highlighted with a blue border represent the original $19$ classes from Cityscapes and, therefore, will be masked in subsequent steps, having reached their maximum refinement level. Classes in dashed boxes, instead, are not present in the IDD split (\ie, \emph{ground}, \emph{terrain}, \emph{public transport} and \emph{train}). Such difference with respect to Cityscapes is due to the fact that \emph{terrain} and \emph{train} are missing, thus leading to the collapse of \emph{ground} in \emph{vegetation} and of \emph{public transport} in \emph{bus} at the second and third steps, respectively.

\newcommand{\imgWidth}{0.2\textwidth}
\begin{figure}[h!]
\centering
\vspace{-.75em}

\begin{subfigure}{\textwidth}
\begin{subfigure}{1em}
\rotatebox[origin=c]{90}{\hspace*{-1em}Cityscapes}
\end{subfigure}
\begin{subfigure}{\textwidth -1em}
\begin{subfigure}{\textwidth -1em}
\begin{subfigure}{\imgWidth}
\centering
\caption*{RGB}
\vspace{-.75em}
\includegraphics[width=\textwidth]{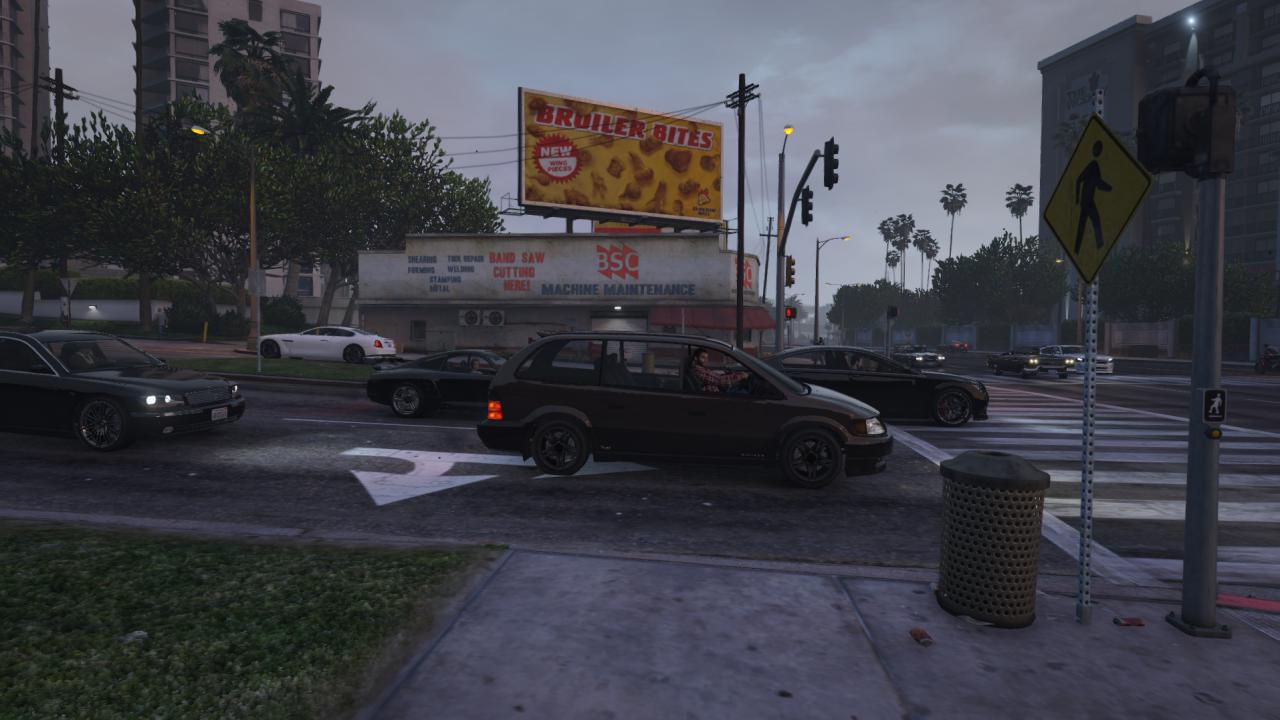}
\end{subfigure}%
\begin{subfigure}{\imgWidth}
\centering
\caption*{Step 0}
\vspace{-.75em}
\includegraphics[width=\textwidth]{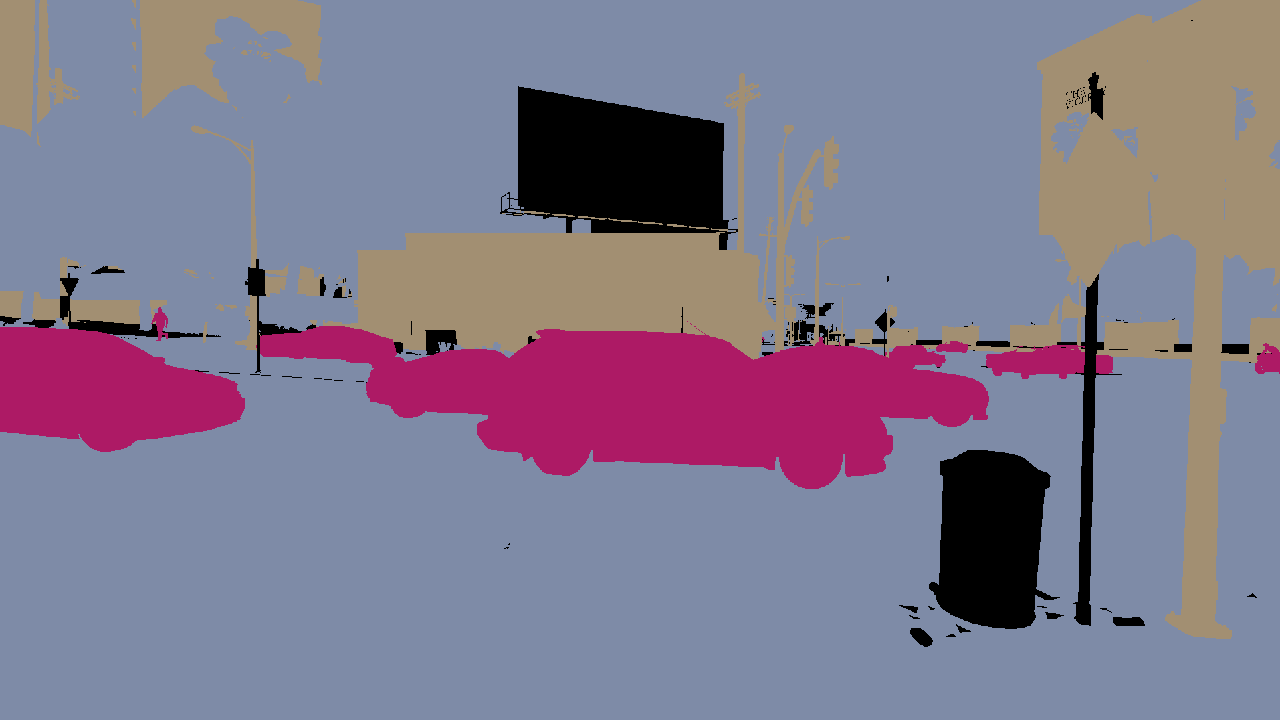}
\end{subfigure}%
\begin{subfigure}{\imgWidth}
\centering
\caption*{Step 1}
\vspace{-.75em}
\includegraphics[width=\textwidth]{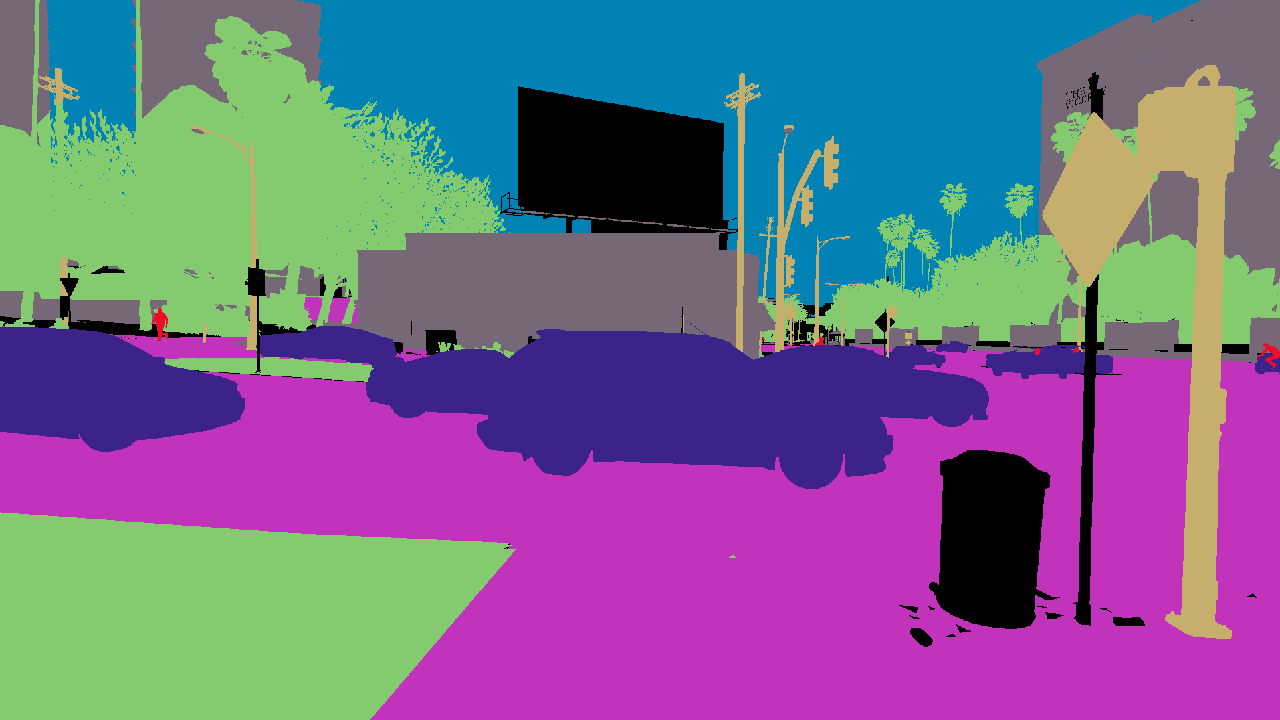}
\end{subfigure}%
\begin{subfigure}{\imgWidth}
\centering
\caption*{Step 2}
\vspace{-.75em}
\includegraphics[width=\textwidth]{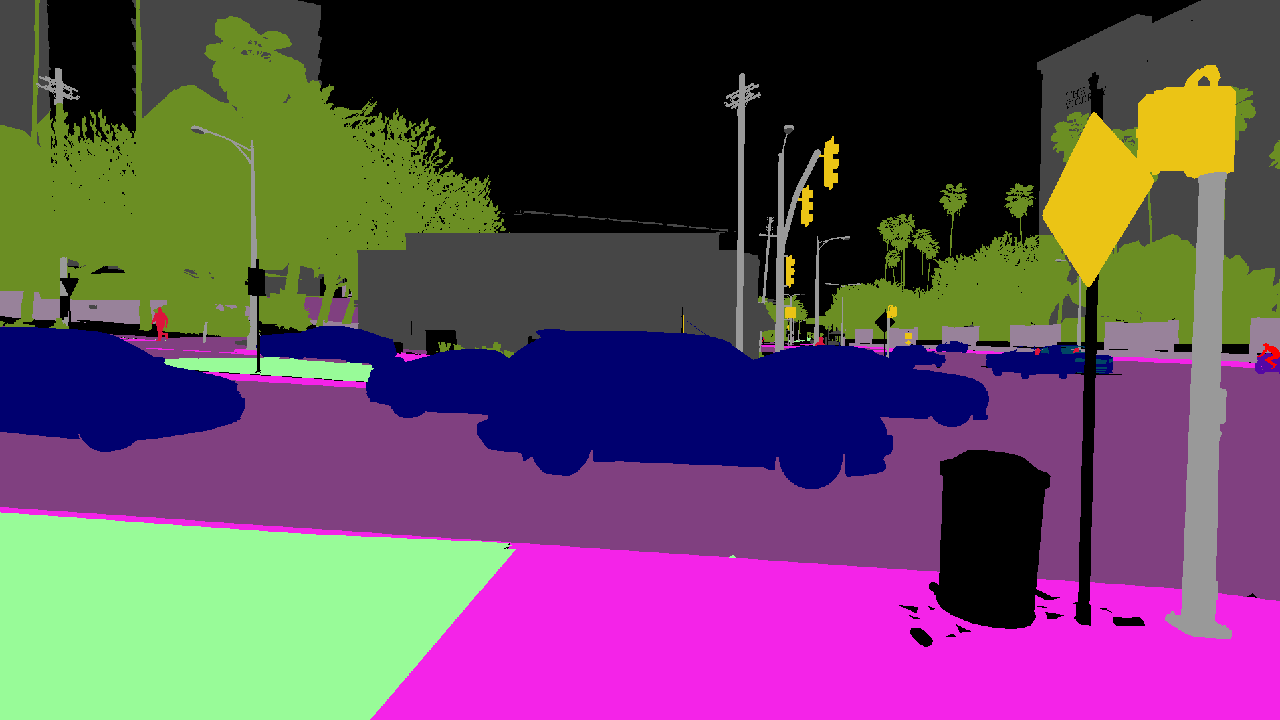}
\end{subfigure}%
\begin{subfigure}{\imgWidth}
\centering
\caption*{Step 3}
\vspace{-.75em}
\includegraphics[width=\textwidth]{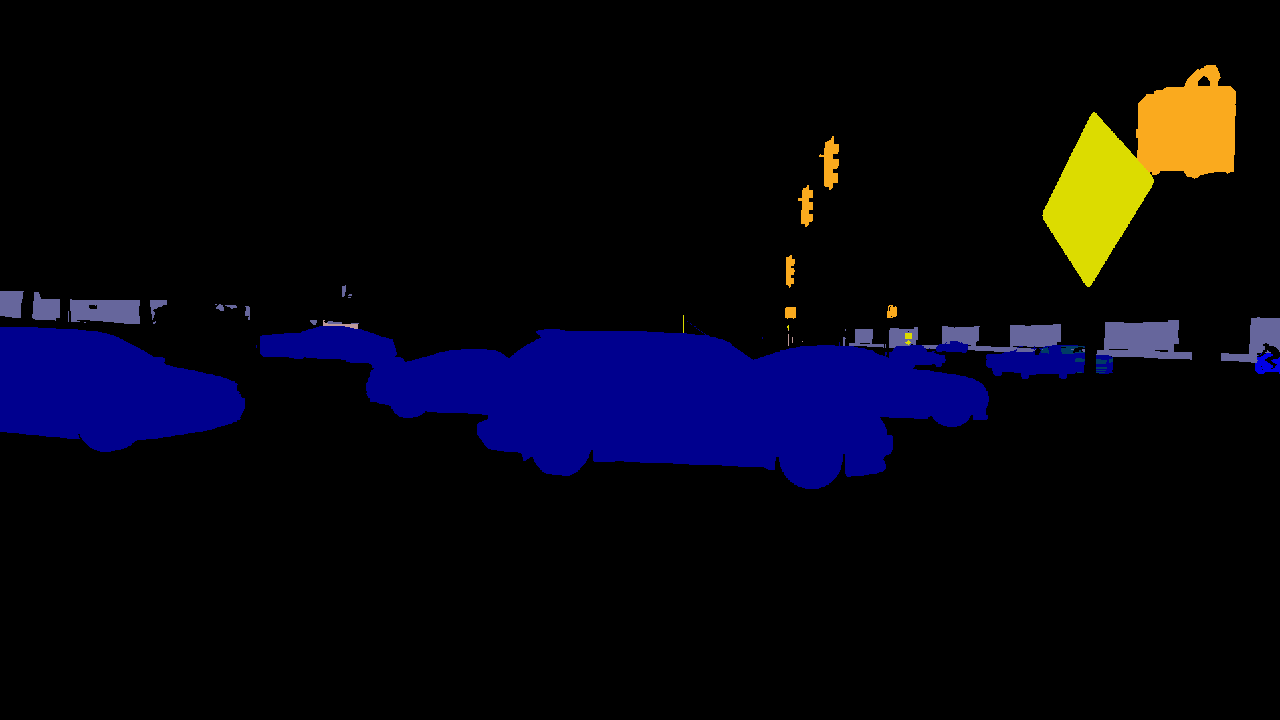}
\end{subfigure}
\end{subfigure}

\begin{subfigure}{\textwidth -1em}
\begin{subfigure}{\imgWidth}
\centering
\includegraphics[width=\textwidth]{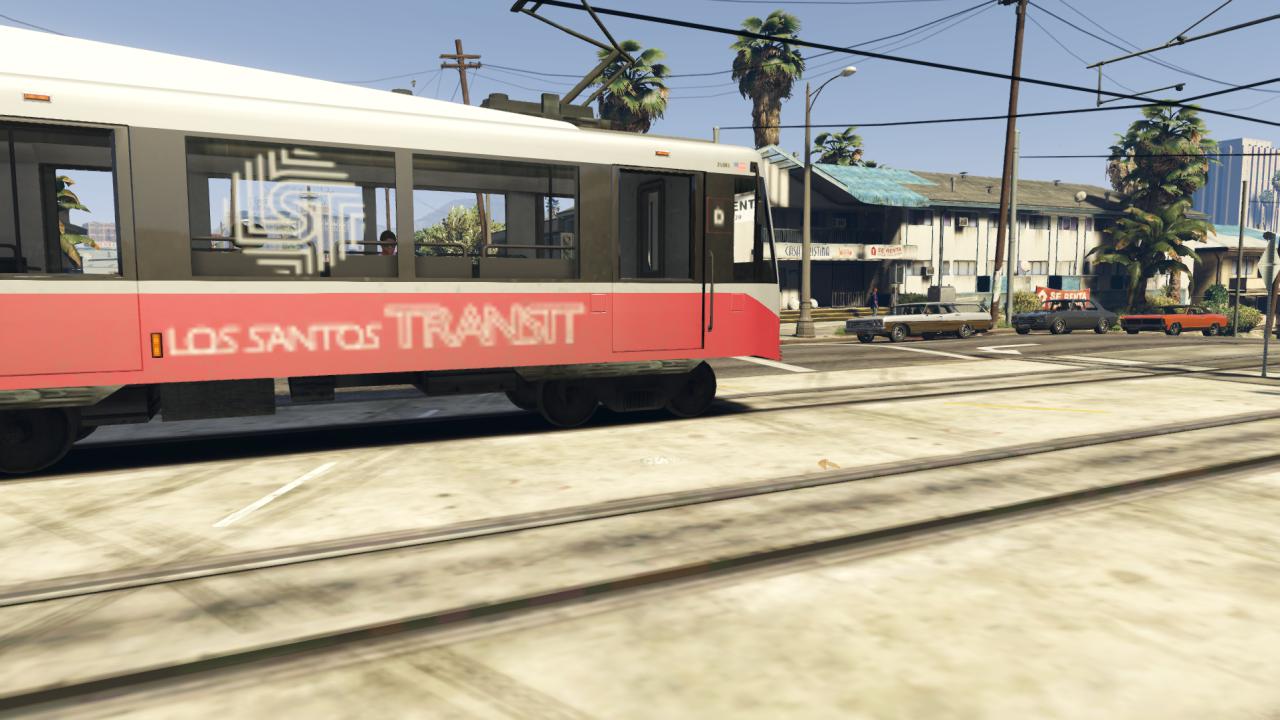}
\end{subfigure}%
\begin{subfigure}{\imgWidth}
\centering
\includegraphics[width=\textwidth]{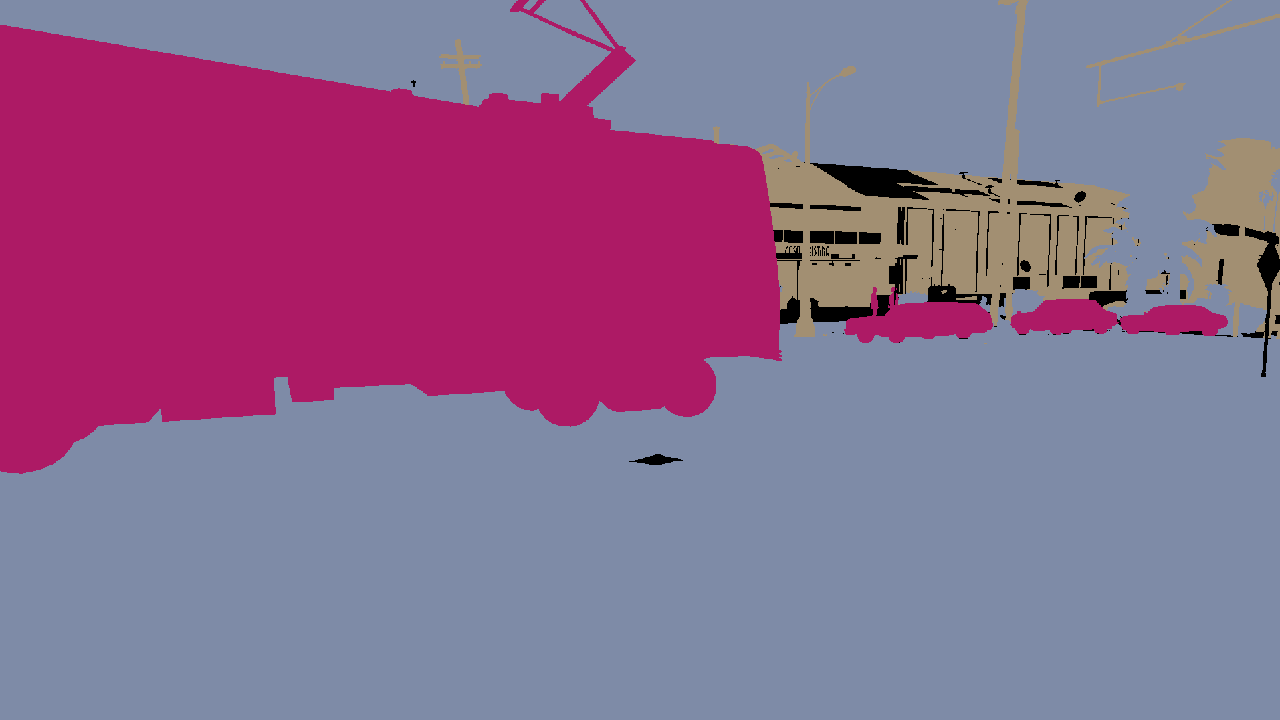}
\end{subfigure}%
\begin{subfigure}{\imgWidth}
\centering
\includegraphics[width=\textwidth]{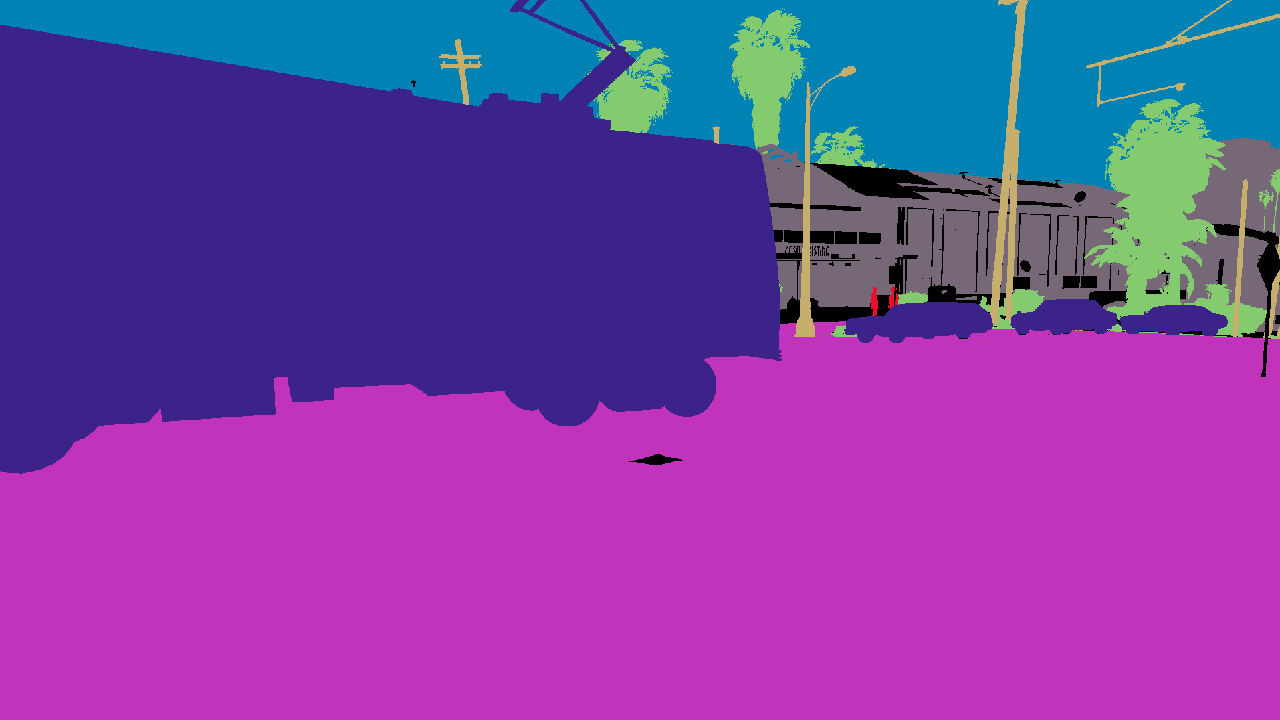}
\end{subfigure}%
\begin{subfigure}{\imgWidth}
\centering
\includegraphics[width=\textwidth]{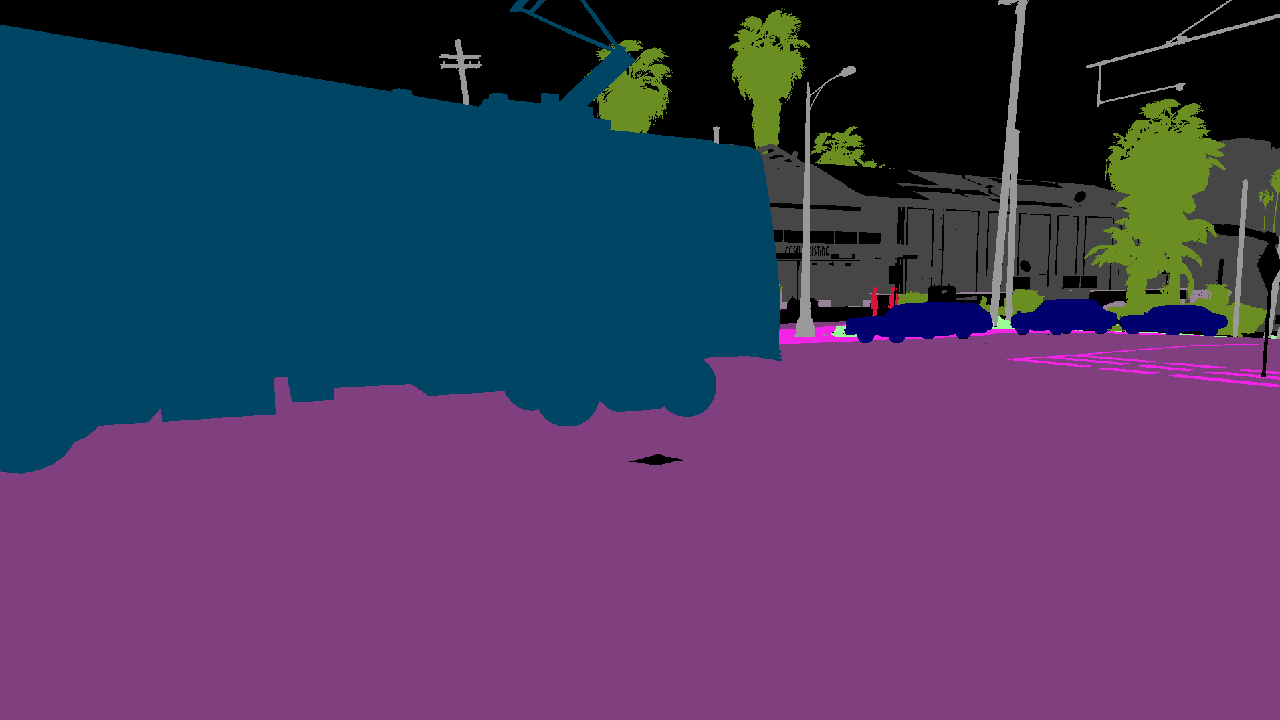}
\end{subfigure}%
\begin{subfigure}{\imgWidth}
\centering
\includegraphics[width=\textwidth]{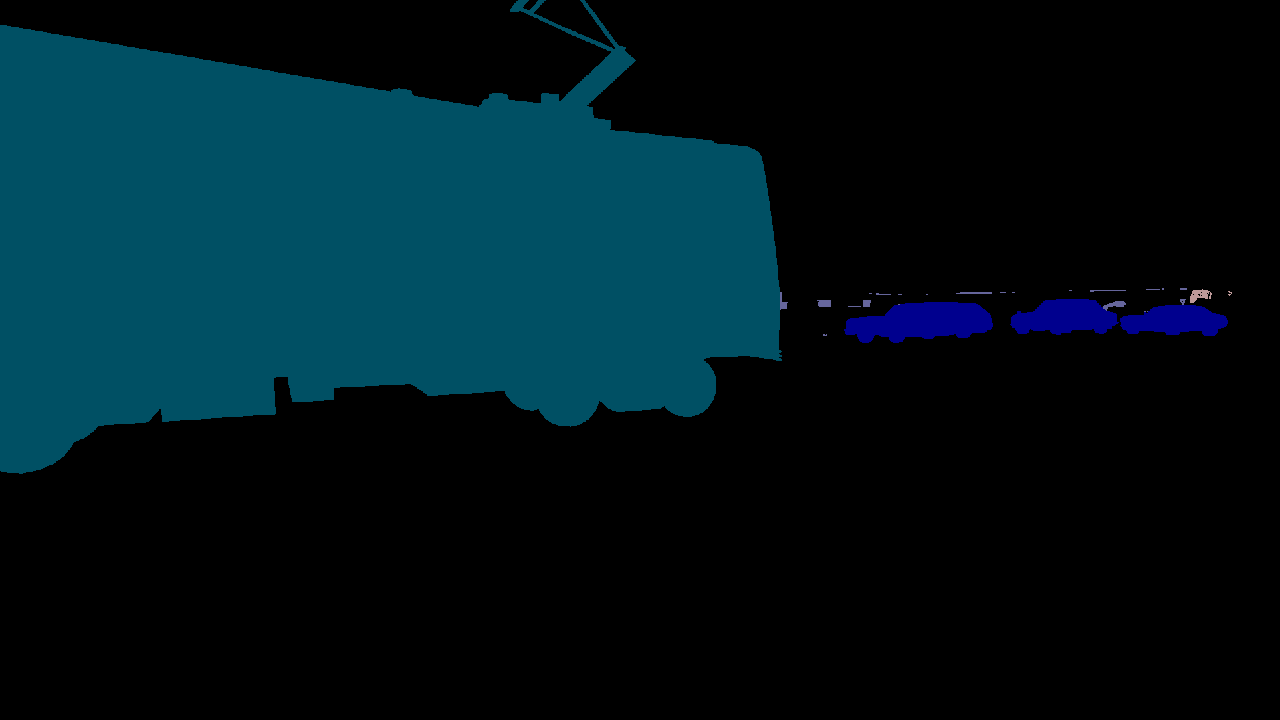}
\end{subfigure}
\end{subfigure}
\end{subfigure}
\end{subfigure}

\begin{subfigure}{\textwidth}
\begin{subfigure}{1em}
\rotatebox[origin=c]{90}{IDD}
\end{subfigure}
\begin{subfigure}{\textwidth -1em}
\begin{subfigure}{\textwidth -1em}
\begin{subfigure}{\imgWidth}
\centering
\includegraphics[width=\textwidth]{annotation_gta/151.jpeg}
\end{subfigure}%
\begin{subfigure}{\imgWidth}
\centering
\includegraphics[width=\textwidth]{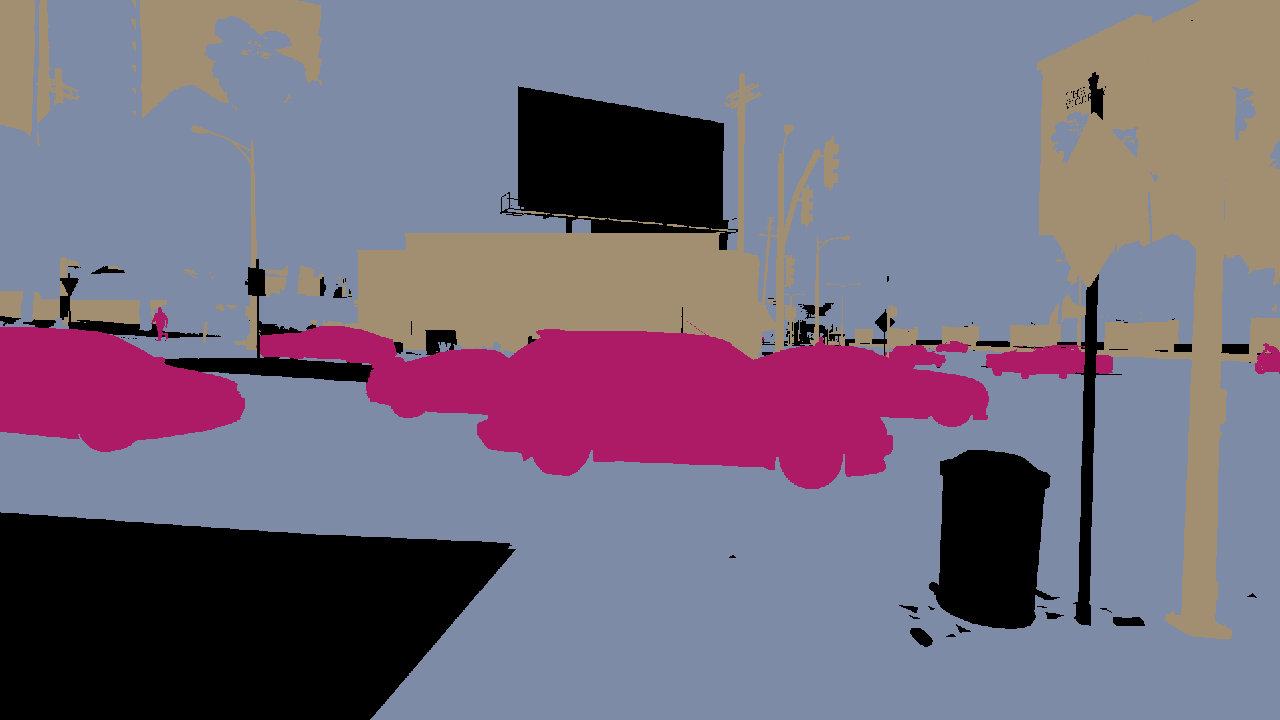}
\end{subfigure}%
\begin{subfigure}{\imgWidth}
\centering
\includegraphics[width=\textwidth]{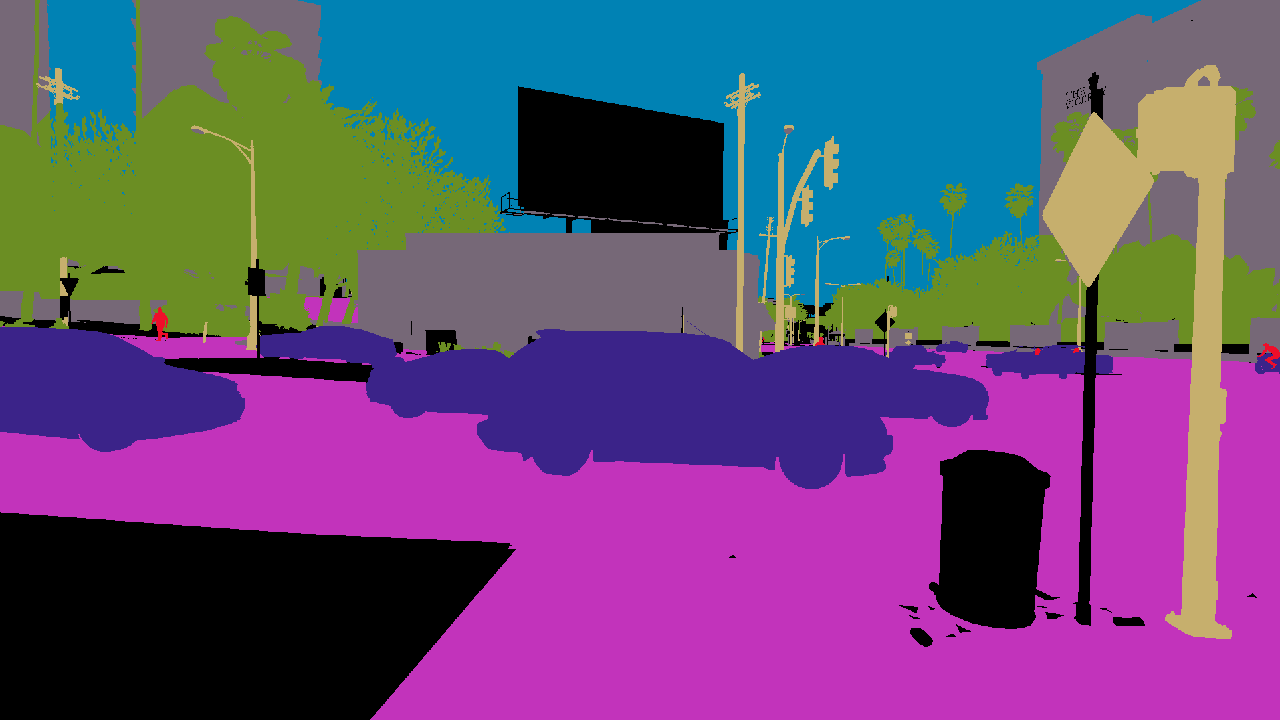}
\end{subfigure}%
\begin{subfigure}{\imgWidth}
\centering
\includegraphics[width=\textwidth]{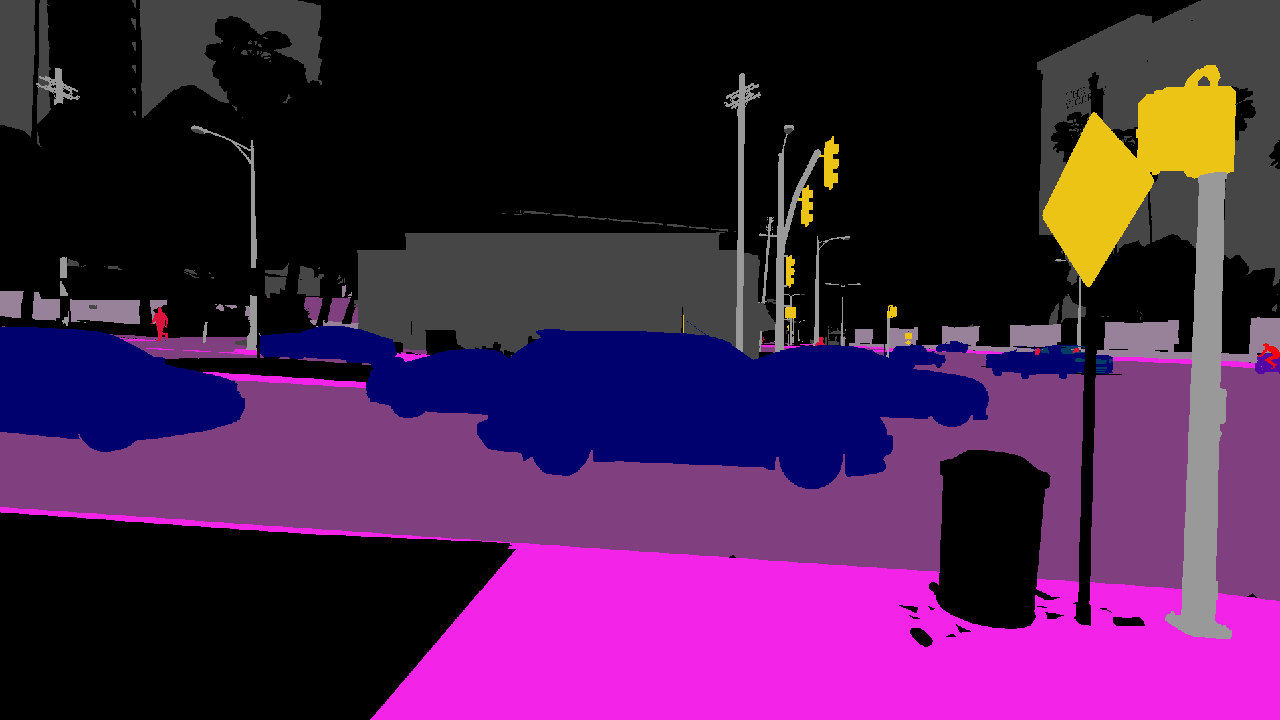}
\end{subfigure}%
\begin{subfigure}{\imgWidth}
\centering
\includegraphics[width=\textwidth]{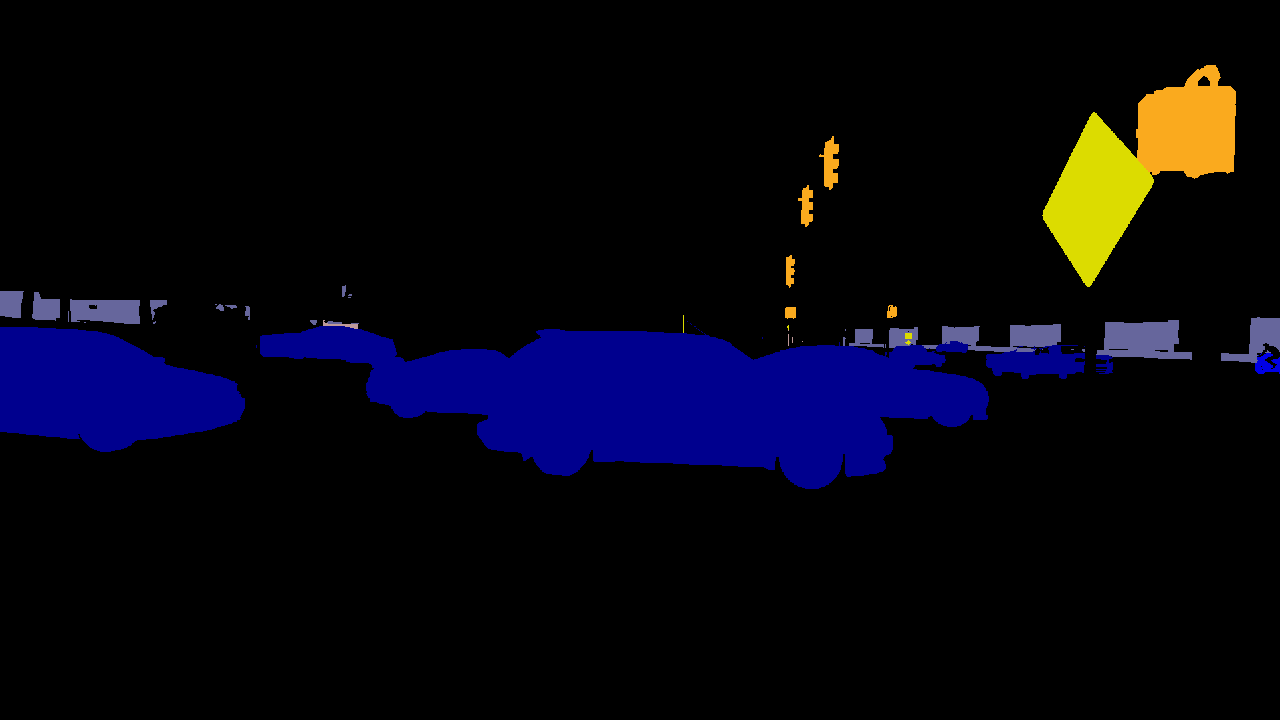}
\end{subfigure}
\end{subfigure}

\begin{subfigure}{\textwidth -1em}
\begin{subfigure}{\imgWidth}
\centering
\includegraphics[width=\textwidth]{annotation_gta/126.jpeg}
\end{subfigure}%
\begin{subfigure}{\imgWidth}
\centering
\includegraphics[width=\textwidth]{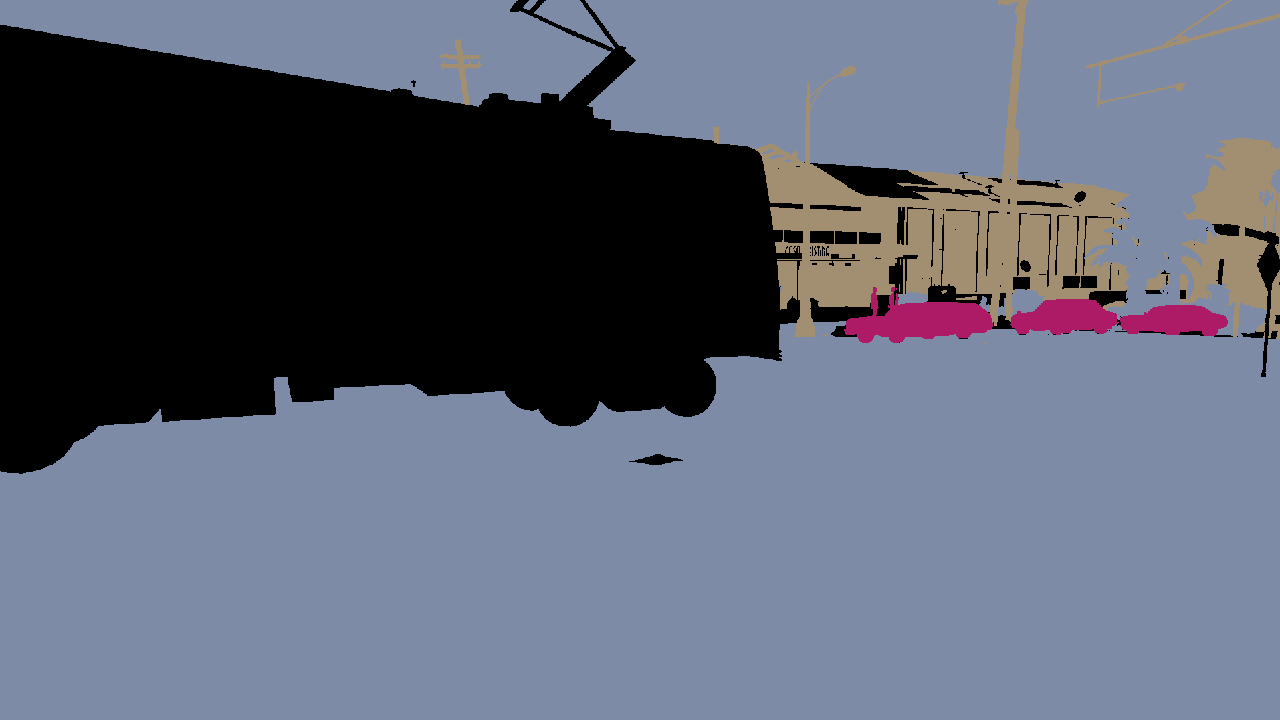}
\end{subfigure}%
\begin{subfigure}{\imgWidth}
\centering
\includegraphics[width=\textwidth]{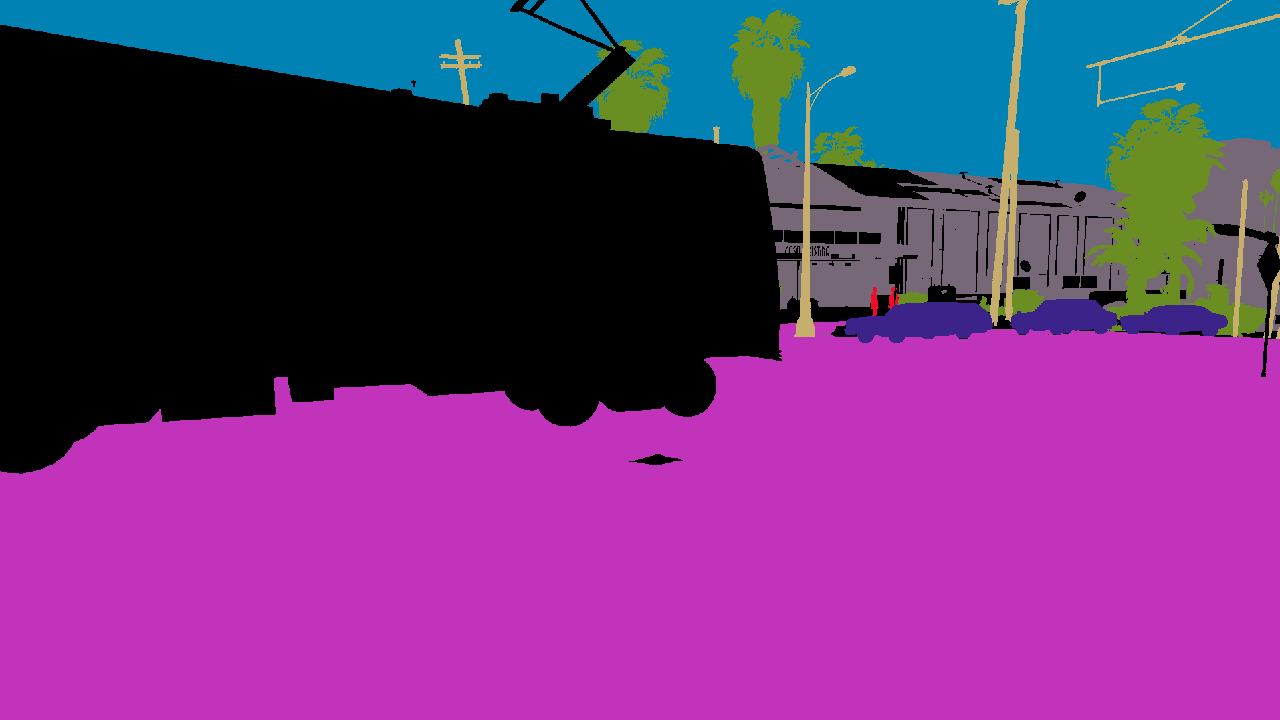}
\end{subfigure}%
\begin{subfigure}{\imgWidth}
\centering
\includegraphics[width=\textwidth]{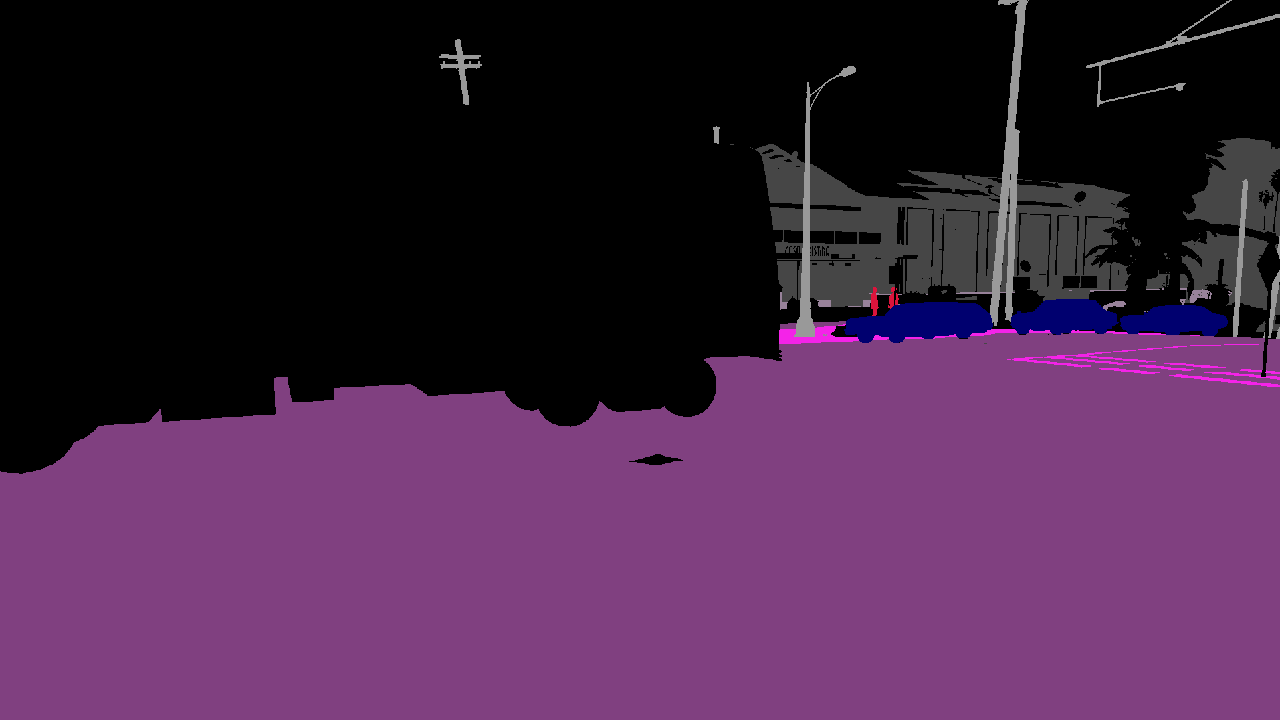}
\end{subfigure}%
\begin{subfigure}{\imgWidth}
\centering
\includegraphics[width=\textwidth]{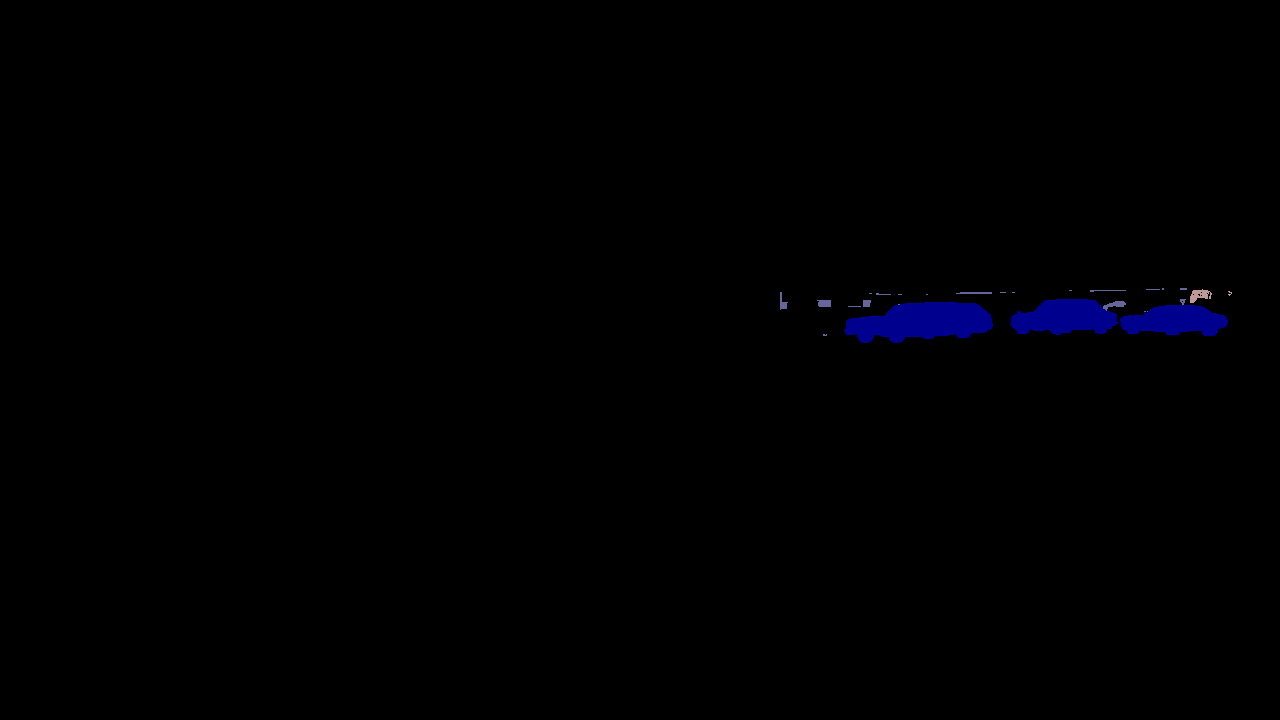}
\end{subfigure}
\end{subfigure}
\end{subfigure}
\end{subfigure}

\caption{Continual coarse-to-fine annotation maps on the GTA5 dataset over different learning steps. These are the images used for supervised training. First two images reported in the Cityscapes class-split, second two in the IDD one.}
\label{fig:annotation_maps_gta}
\end{figure}

In Figure~\ref{fig:annotation_maps_gta} we show a qualitative example of the per-step labels provided to the architecture during training, notice how the fine classes are present only in one step and are masked in subsequent ones.
Additionally, we can see how  
since the \emph{terrain} in the first row and the \emph{train} in the second row are missing in the IDD dataset,
we masked them to \emph{void} from the very first step when the IDD split is used, as can be verified by third and fourth row. On Cityscapes, instead, these classes are present (first two rows).
This results in a different pixel-level class distribution between the two benchmarks.

\begin{figure}[ht]
\centering
\begin{subfigure}{\textwidth}
\centering
\includegraphics[width=\textwidth]{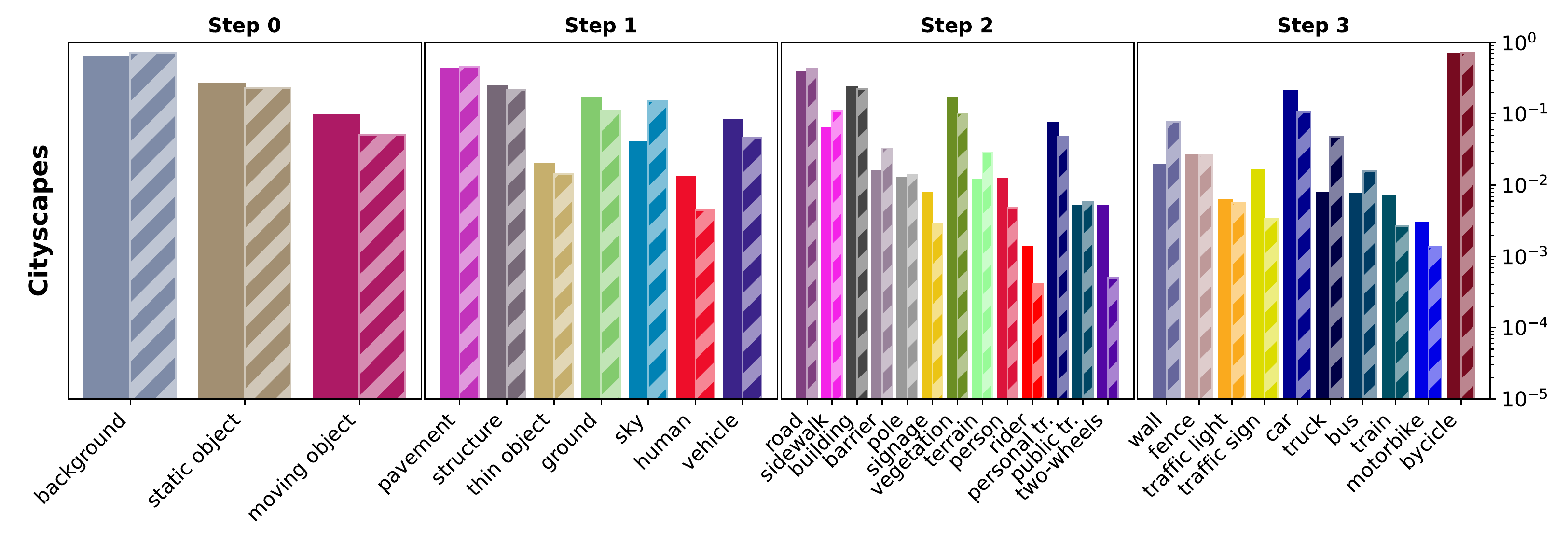}
\end{subfigure}\\
\begin{subfigure}{\textwidth}
\centering
\includegraphics[width=\textwidth]{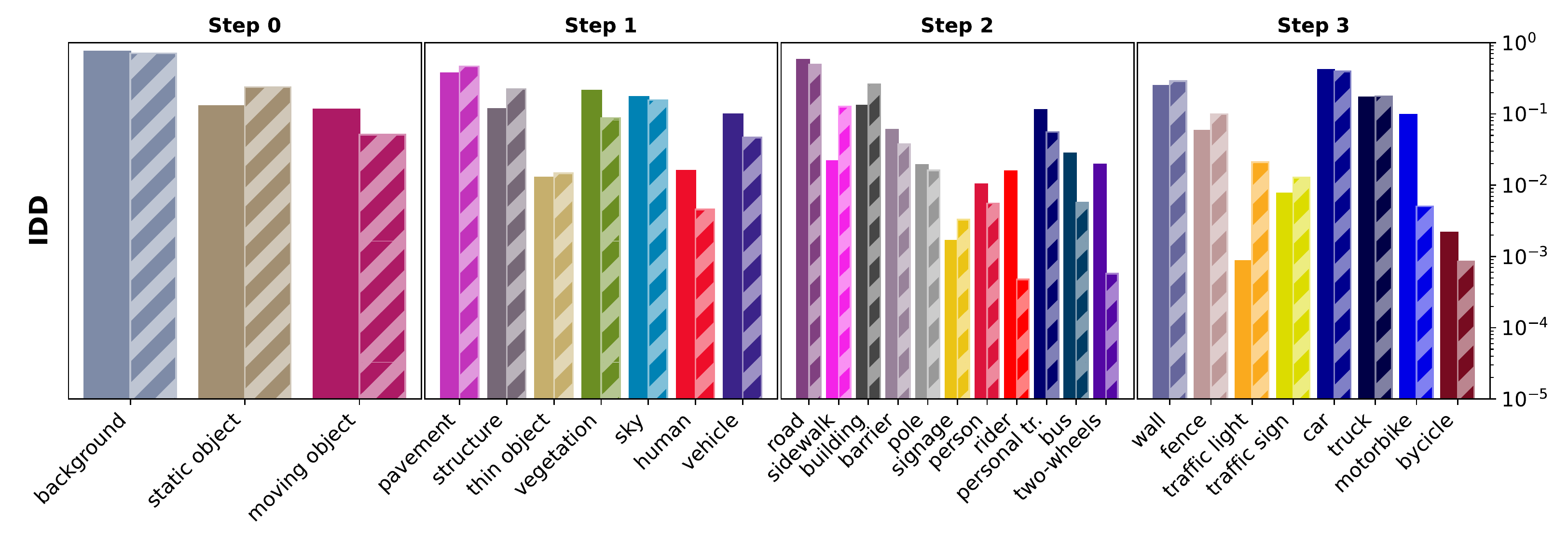}
\end{subfigure}
\caption{Relative frequency of pixels of given classes in each incremental step for both the considered target datasets. The classes at the incremental steps are refined according to the label splits of Figure~\ref{fig:label_splits}}
\label{fig:train_classes}
\end{figure}

Performing the training in an incremental manner dramatically changes the pixel-class distributions seen by the architecture during training and significantly bridges the domain gap. This is in part due to the effect of masking common old fine classes (\ie, with many pixels), which allows to remove their contribution and leads to a much more uniform distribution, and in part due to the semantic grouping we propose, which is designed to reduce the distribution shift between source and target domain in each of the incremental steps.
In Figure~\ref{fig:train_classes} we report the relative class frequency of each incremental step comparing the source and target distributions. From visual inspection it is clear that the distributions are much more aligned (both in the \textit{GTA5}$\rightarrow$\textit{Cityscapes} and in the \textit{GTA5}$\rightarrow$\textit{IDD} setups) than the complete distributions shown in Figure~\ref{fig:gen_dset_freqs}. This empirical finding is quantitatively confirmed when computing the KL-Divergence on the distributions. The original distribution scores $0.1361$ and $0.2415$ on GTA5-Cityscapes and GTA5-IDD, respectively. 
On the other hand, the incremental steps score {$\{0.0216, 0.0925, 0.0645, 0.1138\}$} (average $0.0731$) and {$\{0.0604, 0.1374, 0.2556, 0.2286\}$} (average $0.1705$), respectively. Such scores imply an average reduction in the distribution shift of $46.3\%$ when using Cityscapes as target and of $29.4\%$ when using IDD.

\section{Proposed Approach}
\label{sec:method}
\begin{figure}[t]
    \centering
    \includegraphics[width=\textwidth]{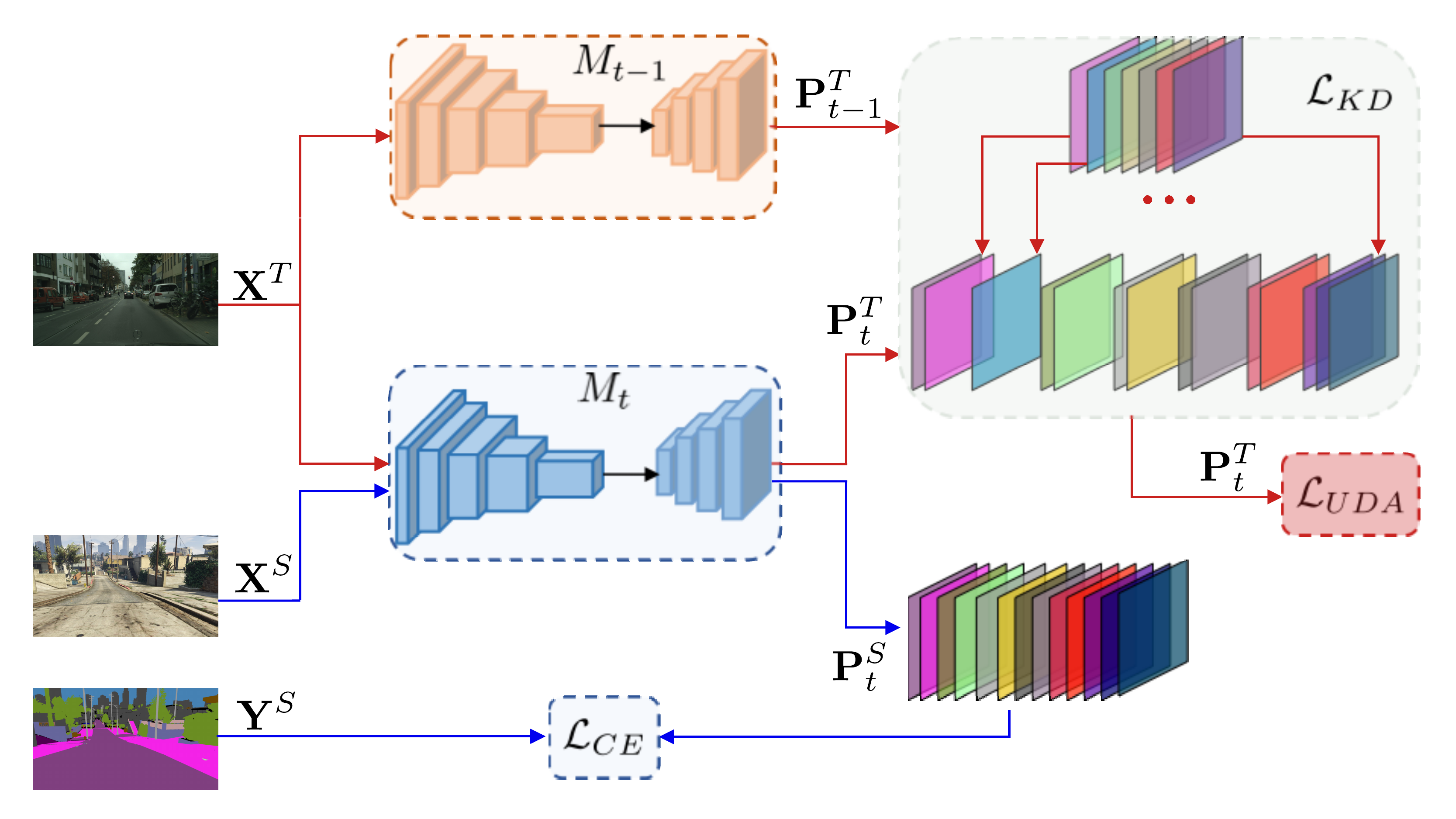}
    \caption{Overview of the proposed approach. The new refined classes are learned on the supervised source domain through $\mathcal{L}_{CE}$. At the same time, the predictions at current step are adapted to the unsupervised target domain with $\mathcal{L}_{UDA}$. Finally, the coarse-to-fine distillation module $\mathcal{L}_{KD}$ is used to semantically adapt the knowledge from coarse to finer classes, while preserving the knowledge of the previously learned ones.}
    \label{fig:proposed_approach}
\end{figure}

In this section, we provide a detailed description of our proposed method: CCDA, Continual Coarse-to-fine Domain Adaptation. 
Our approach combines continual learning techniques with domain adaptation ones allowing incremental learning of semantically refined (finer) classes, while preserving the knowledge of older (coarse) classes and the semantic relation between any coarse class and its fine child classes.
Moreover, to reduce the distribution gap between source and target domains, we include  ad-hoc unsupervised domain adaptation techniques. Such strategies are employed jointly with the continual learning ones, allowing to incrementally adapt the architecture from simpler to harder settings (the number of classes increases throughout training) while at the same time improving the adaptation performance.
A graphical representation of the proposed approach is shown in Figure~\ref{fig:proposed_approach}, highlighting the three modules that make up our strategy. Here we see how the cross-entropy, knowledge distillation and domain adaptation objectives are used in conjunction with the previous step's model and with source and target samples to optimize and adapt the current step's architecture.
The contribution of the various modules is summarized in the loss function:
\begin{equation}
    \mathcal{L} = \mathcal{L}_{CE} + \lambda_{UDA}\mathcal{L}_{UDA} + \lambda_{KD}^{c}\mathcal{L}_{KD}^{c} + \lambda_{KD}^{f}\mathcal{L}_{KD}^{f} ,
\end{equation}
where $\mathcal{L}_{CE}$ is the cross entropy loss, used to learn new classes in the supervised source domain.
The other loss terms will be described in detail in the following subsections: 
the domain adaptation ($\mathcal{L}_{UDA}$)  loss based on maximum squares minimization is discussed in Section~\ref{subsec:uda_method}, while the coarse-to-fine knowledge distillation loss ($\mathcal{L}_{KD}^{c}$ and $\mathcal{L}_{KD}^{f}$) is introduced in Section~\ref{subsec:c2fkd}.
The hyper-parameters $\lambda_{UDA}, \lambda_{KD}^{c}$ and $\lambda_{KD}^{f}$ control the relative contribution of each component, and have been optimized on a validation set extracted from the training data of the target domains (as detailed in Section~\ref{sec:impl}).
Finally, the coarse-to-fine weight initialization rule is discussed in Section~\ref{subsec:c2fuwi}.

\subsection{Unsupervised Domain Adaptation}
\label{subsec:uda_method}
In this section we overview the UDA module of our method, presenting the problems that sparked its design.
In a typical (closed-set) UDA setting we are interested in performing adaptation between two domains sharing a fixed set of classes in a single step. However, when the set of classes is large, when the target dataset does not provide enough distinctive cues to differentiate between classes or when such cues are too different from those learned on the source domain, the adaptation process has sub-optimal performances.
A further cause of performance degradation is the inherent penalization of less common classes, which are learned through few very specific features (given the small number of samples and, therefore, the limited variability available) which can easily be missing or different in the target domain.

To tackle these issues, a good choice is to perform UDA in a curriculum learning fashion, starting from the adaptation of high level semantic concepts and transferring, in a multitude of steps, the finer and finer representations of the classes in a format compatible with the target dataset.
The result of this approach is a more gradual and resilient adaptation, since the number of the new classes to be adapted is always smaller than in the joint case. This reduced number of classes can lead to significant improvements in the adaptation procedure, particularly when self-supervised techniques are employed. This is due to the reduced entropy of the prediction system, which having a restricted number of possible outputs has also a much smaller number of possible false predictions.
A possible solution to these problems is the the maximum squares minimization loss \cite{Chen2019}, which we reframe for use in the proposed novel coarse-to-fine task.
Formally, at each training step $t$, the objective to be optimized is:
\begin{align}
        \label{eq:iw_uda}
        \mathcal{L}_{UDA} &= - \sum_{\mathbf{X} \in \mathcal{T}_{t}^T}  \sum_{c\in \mathcal{C}_{t}} \frac{1}{2|\mathcal{C}_{t}|^\alpha (H W)^{1-\alpha} }  \left(\mathbf{P}^T_{t}[c]\right)^2,
\end{align}
where $|\mathcal{C}_{t}|$ represents the number of classes at each training step $t$, $H$ and $W$ are the  image dimensions (\ie, $HW$ is the number of pixels), and $\alpha$ is an hyper-parameter.
Recall, also, that $\mathbf{P}_t = M_t(\mathbf{X})$ for any given input image $\mathbf{X}$.
At the end of each incremental step the model will be able to produce increasingly accurate predictions on the target domain, which are crucial for the successful domain-bridging effect of our knowledge distillation module (see Section~\ref{subsec:c2fkd}).

\subsection{Coarse-to-Fine Learning}
\label{subsec:c2f_method}
In this section we report in detail the two continual learning modules employed by our architecture: in Section \ref{subsec:c2fkd} we present our novel cross-domain coarse-to-fine knowledge distillation strategy, while in Section \ref{subsec:c2fuwi} we show our unbiased coarse-to-fine weight initialization strategy.

\subsubsection{Coarse-to-Fine Knowledge Distillation}
\label{subsec:c2fkd}
Knowledge distillation is an established technique typically used in continual learning settings to preserve previous learned knowledge.
Given its effectiveness, we extended the standard definition for use in our coarse-to-fine task.
This technique assumes that at each training step the previous learned model is available, which will be used to guide the preservation of the learned knowledge on the previous classes. Storing the
previous model, does not compromise the privacy of the training procedure, because we do not retain any previous image nor label used during training, and the model will be used just to perform inference on new training images (as it would do in any deployment setting).

Our key insight in improving the knowledge transfer across coarse and fine class sets, was to realize the importance of constraining the prediction of the fine classes into resembling their parent coarse class (\ie, the one from which they have been derived). With this approach we benefit both in terms of accuracy in the recognition of new classes, and in the reduction of misclassification occurrences (particularly between semantically unrelated classes, \ie, those that do not share a common ancestor in the hierarchical split, see Fig. \ref{fig:label_splits}). 

Hence, at each training step $t$, we force the output of the previous model $M_{t-1}$ at channel (coarse class) $c$, to match to the sum of the output channels $f \in S_t(c)$ (\ie, the set of $c$'s child classes) of the current model $M_t$. The matching is implemented with a cross entropy loss function, resorting to pseudo-labels computed on target domain samples, effectively enforcing an additional domain-bridging objective.
This constraint results to be the natural coarse-to-fine extension of the simple approach proposed for standard continual learning in \cite{cermelli2020modeling}, where all fine classes are derived from a unique \emph{background} macro class.
More formally, the knowledge distillation $\mathcal{L}^{c}_{KD}$ for the classes $\mathcal{C}^c_{t-1}$ that are further split can be expressed as:
\begin{align}
    \label{eq:kd_macro}
    \mathcal{L}^{c}_{KD} = \frac{1}{|\mathcal{T}_{t}^T|} \sum_{\mathbf{X} \in \mathcal{T}_{t}^T}  \sum_{c\in \mathcal{C}^c_{t-1}} \mathbf{P}^T_{t-1}[c]  \sum_{f\in S_{t}(c)} \log{\mathbf{P}^T_t[f]}.
\end{align}

When considering classes that have reached the maximum level of semantic refinement (\ie, that are not going to be further split), the expression collapses into the standard cross-entropy based knowledge distillation (since the sets $\mathcal{S}_t(c)$ contain only one class, $c$ itself, the sum of logs disappears), which forces the prediction of each channel $c$ in the previous step $t-1$, to approximate the same class $c$ in the next step $t$.
Mathematically:
\begin{align}
    \label{eq:kd_fine}
    \mathcal{L}_{KD}^{f} = \frac{1}{|\mathcal{T}_{t}^T|} \sum_{\mathbf{X} \in \mathcal{T}_{t}^T}  \sum_{c\in \mathcal{C}^f_{t-1}} \mathbf{P}^T_{t-1}[c] \log{ \mathbf{P}^T_t[c]}.
\end{align}

In our approach (CCDA), at each training step $t$, we can rely upon the knowledge that the previous model $M_{t-1}$ has been adapted to segment images coming from the target domain. 
For this reason, the knowledge distillation is performed exploiting target domain samples, leading to an additional domain bridging effect. 
As an additional study, we analyzed a variant of this approach SKDC (Source domain Knowledge-Distillation for Coarse-to-fine) which implements the same knowledge distillation module, but the pseudo-labels are computed from the source domain, removing the domain bridging component.

\subsubsection{Coarse-to-Fine Weight Initialization}
\label{subsec:c2fuwi}
When training on a new model, the weights of the classifier are usually initialized in a random fashion, since there is no \textit{a priori} information to be exploited to identify a better value.
In a continual learning setup, though, this is true only in the first incremental step ($t=0$). In the subsequent steps (\ie, $t>0$) a better strategy for the initialization of the weights of the decoder $D_t$ is to exploit the knowledge provided by the decoder of the previous step, $D_{t-1}$ (recall that $M_t = D_t \circ E_t$), which can be used to obtain weights relative to already-seen classes (reducing the number of random values to be used). Furthermore, failing to recognize this detail can lead to significant performance losses during the network optimization, as a total random initialization may compromise the effectiveness of the knowledge distillation objective. 
Our setup, differently from the incremental learning case (where classes are added without any semantic relationship between each other), provides additional \textit{priors} which can be exploited to significantly improve the weights initialization: the hierarchical coarse-to-fine labels relationship. Such information allows to remove completely the need for random initialization, since all the finer classes predicted by $D_t$ have a macro class in $D_{t-1}$.
Our strategy, therefore, tracks and exploits this relationship to initialize weights for the new fine classes: the weights of each new child fine class are initialized with the value of its coarse parent class. Naturally, when a class reached the maximum level of refinement in the subsequent steps, its weights will be initialized with the same values of the previous step.
Formally, the operations can be defined as:
\begin{align}
    \{\omega^{f}_{t} \} &=  \{\omega^c_{t-1} \} \quad \forall c \in C_{t-1}^c, \forall f \in S_{t}(c) \label{eq:weight:coarse} \\
    \{\omega^{f}_{t} \} &=  \{\omega^f_{t-1} \} \quad \forall f \in C_{t-1}^f. \label{eq:weight:fine}
\end{align}
For what concerns the biases initialization, we adopted the idea introduced in\cite{cermelli2020modeling} for class incremental learning, again extending it to the more general case of coarse-to-fine learning.
The idea consists in spreading the probability of each coarse class $c \in C_{t-1}^c$ towards its respective set of fine classes $S_{t}(c)$.
In formal terms, the bias update expression becomes:  
\begin{align}
    \{\beta^{f}_{t}\} &= \{\beta^{c}_{t} - \log(|S_{t}(c)|) \}  \quad \forall c \in C_{t-1}^c, \forall f \in S_{t}(c) \label{eq:bias:coarse}\\  
    \{\beta^{f}_{t}\} &= \{\beta^{f}_{t} \} \quad \forall f \in C_{t-1}^f. \label{eq:bias:fine}
\end{align}

After applying a softmax layer to compute the output probabilities, this updated rule leads all the fine classes to share an equal fraction of the original coarse class prediction probability. 

For all the competing approaches, the results are reported initializing new weights according to Eq.~\eqref{eq:weight:coarse} and ~\eqref{eq:weight:fine} for fair comparison. Instead, bias initialization rule (Eq.~\eqref{eq:bias:coarse} and ~\eqref{eq:bias:fine}) is only employed in our approach.

In Table~\ref{tab:weight_ablation} of the ablation studies (Section~\ref{subsec:ablation}) we analyzed the effects of this refined bias initialization (referred to as \textit{Unbiased} weight initialization). Particularly, we observe that disabling it (\ie, not subtracting the $\log$ in Eq.~\eqref{eq:bias:coarse}) leads to a serious degradation of the performance. Our results confirm and extend the findings of \cite{cermelli2020modeling} even in the more general case of coarse-to-fine learning, showing how an unbiased strategy can handle difficult class relationships.

\section{Implementation Details}
\label{sec:impl}
The proposed framework is implemented in PyTorch. The semantic segmentation network is a DeepLab-V3 with ResNet-101 as the backbone (pre-trained weights are available at {\url{https://pytorch.org/vision/stable/models.html}}).
In order to perform hyper-parameters selection, we identified a validation set of respectively $496$ and $50$ samples within the original training set of Cityscapes and IDD.
We recall that, at the beginning of each incremental step, the model is initialized with a new classifier, depending on the number of classes to be recognized.
Following \cite{tsai2018,chen2017rethinking} the model is trained using the Stochastic Gradient Descent (SGD) optimizer with momentum equal to $0.9$ and weight decay of $5\times10^{-4}$.
The learning rate is polynomially decreased with a decay power of $0.9$.
The initial value of learning rate for the step $0$ is set to $2.5\times 10^{-4}$, while in the incremental steps it is initialized to $10^{-4}$.
In the \textit{GTA5}$\rightarrow$\textit{Cityscapes} setup, the batch size per domain is set to $2$ (total of $4$ samples), images are resized to $1280 \times 720$ and $1280 \times 640$, respectively for the source and target domains.
Instead, in the \textit{GTA5}$\rightarrow$\textit{IDD} setup, the batch size per domain is set to $1$ (total of $2$ samples) and images are resized to $1280 \times 720$ for both domains.
In both cases, we trained our models on a single NVIDIA RTX 3090, performing $50000$ iterations in the first step and $25000$ iterations for each incremental step.
Following \cite{Chen2019,zhao2017}, we perform data augmentation through random left-right flipping and Gaussian blur to improve generalization properties.
The hyper-parameters, determined on the validation, are the same in both setups: $\lambda_{UDA}=0.1$, $\lambda_{KD}^{f}=10$ except for $\lambda_{KD}^{c}$ which is equal to $10$ for 
\textit{GTA5}$\rightarrow$\textit{Cityscapes} and equal to $1$ for \textit{GTA5}$\rightarrow$\textit{IDD}. 
Finally, when a model makes use of maximum squares loss, we include a warm-up stage of $100$ iterations, and we activate the maximum squares constraint thereafter.


\section{Experimental Results}
\label{sec:results}
In this section we report the experimental results on two synthetic-to-real benchmarks: {\textit{GTA5}$\rightarrow$\textit{Cityscapes}} and {\textit{GTA5}$\rightarrow$\textit{IDD}}, comparing our approach with several methods proposed to solve the two tasks separately. Then, we report some ablation studies to carefully validate the effects of the various components of our approach.

To evaluate our approach, we compare our strategy with several baselines. 
As a global upper bound, we report the results of the offline (single-shot) training on the target dataset, \ie, the Joint Target Only, JTO.
As an upper bound for all the steps of incremental learning approaches, we report the coarse-to-fine training on the target dataset, that we call Target Only Non-Continual, TNC; here, at each step the algorithm sees all the labels including the ones of the classes not refined in the current step.
As a global lower bound for both incremental and UDA approaches we report the na\"ive fine-tuning optimization of the network on the source dataset (Source Only).
Finally, we compare our approach with two competitors, namely: MaxSquareIW \cite{Chen2019} (MSIW) as state-of-the-art UDA approach, which does not employ any incremental technique, and MiB \cite{cermelli2020modeling} as state-of-the-art continual learning technique, without domain adaptation.

Our contributions are named as SKDC (Source Knowledge Distillation for Coarse-to-Fine) and CCDA (Continual Coarse-to-fine Domain Adaptation).\\
SKDC is the continual counterpart of TNC trained on the source dataset, where our Coarse-to-fine Knowledge Distillation strategy is employed as a mean to preserve previously acquired knowledge. CCDA is our full approach and it combines our domain-adaptive approach and SKDC: here, both domain adaptation and knowledge distillation losses are employed in all steps. Differently from SKDC, though, the preserved knowledge comes from the target dataset, instead of the source one (\ie, the pseudo-labels used in the knowledge distillation loss are generated via inference on the target dataset sample). This allows the knowledge distillation loss to perform additional entropy minimization and domain bridging tasks.

In Table~\ref{tab:main} we report the mean quantitative results of all the approaches in both considered settings. In Tables~\ref{tab:gta_city_quantitative_per_class} and~\ref{tab:gta_IDD_quantitative_per_class} we report the detailed per-step and per-class IoU scores attained by our full approach (CCDA) in the benchmarks. 
All tables show the score on the target dataset. 

\begin{table}[ht]
\centering
\setlength\tabcolsep{1.1em}
\renewcommand{\arraystretch}{.8}
\begin{tabularx}{\textwidth}{cXcccc}
\toprule
 & Model & mIoU\textsubscript{0} & mIoU\textsubscript{1} & mIoU\textsubscript{2} & mIoU\textsubscript{3} \\
\toprule
 \multirow{7}{*}{\rotatebox{90}{GTA5$\rightarrow$Cityscapes}} & JTO & - & - & - & 68.6 \\
 \cdashline{2-6}
 & TNC & 92.1 & 83.7 & 72.6 & 66.5 \\
 &  Source Only & 65.0 & 56.7 & 25.1 & 4.5 \\ 
 \cdashline{2-6}
 & MiB \cite{cermelli2020modeling} & 65.0 & 56.1 & 27.8 & 26.5 \\ 
 & MSIW \cite{Chen2019} & \textbf{85.4} & 62.2 & \textbf{38.5} & 6.9 \\
 \cdashline{2-6}
 & SKDC (ours)  & 65.0 & 65.4 & 34.4 & 30.4 \\
 & CCDA (ours)  & \textbf{85.4} & \textbf{67.9} & 37.2 & \textbf{33.1} \\
\midrule
 \multirow{7}{*}{\rotatebox{90}{GTA5$\rightarrow$IDD}} & JTO & - & - & - & 65.9 \\
 \cdashline{2-6}
 & TNC & 87.0 & 78.8 & 73.2 & 64.2 \\
& Source only & 72.9 & 60.8 & 30.0 & 6.1 \\ 
 \cdashline{2-6}
& MiB \cite{cermelli2020modeling} & 72.9 & 61.8 & \textbf{44.7} & 26.8 \\
& MSIW \cite{Chen2019} & \textbf{75.9} & 62.7 & 30.2 & 8.9 \\ 
 \cdashline{2-6}
 & SKDC (ours)  & 72.9 & 60.9 & 38.7 & 32.4 \\
 & CCDA (ours)  & \textbf{75.9} & \textbf{63.5} & 40.3 & \textbf{33.0} \\
\bottomrule
\end{tabularx}
\caption{Quantitative results in terms of mIoU computed on the validation sets of Cityscapes and IDD, when employing source knowledge from the GTA5 dataset.}
\label{tab:main}
\end{table}

\subsection{GTA5\texorpdfstring{$\rightarrow$}{->}Cityscapes adaptation}

\definecolor{road}{rgb}{.502,.251,.502}
\definecolor{sidewalk}{rgb}{.957,.137,.910}
\definecolor{building}{rgb}{.275,.275,.275}
\definecolor{wall}{rgb}{.4,.4,.612}
\definecolor{fence}{rgb}{.745,.6,.6}
\definecolor{pole}{rgb}{.6,.6,.6}
\definecolor{tlight}{rgb}{.980,.667,.118}
\definecolor{tsign}{rgb}{.863,.863,0}
\definecolor{vegetation}{rgb}{.420,.557,.137}
\definecolor{terrain}{rgb}{.596,.984,.596}
\definecolor{sky}{rgb}{0,.510,.706}
\definecolor{person}{rgb}{.863,.078,.235}
\definecolor{rider}{rgb}{1,0,0}
\definecolor{car}{rgb}{0,0,.557}
\definecolor{truck}{rgb}{0,0,.275}
\definecolor{bus}{rgb}{0,.235,.392}
\definecolor{train}{rgb}{0,.314,.392}
\definecolor{motorbike}{rgb}{0,0,.902}
\definecolor{bicycle}{rgb}{.467,.043,.125}
\definecolor{unlabelled}{rgb}{0,0,0}
\definecolor{backgroud}{rgb}{0.494,0.545,0.655}
\definecolor{static_object}{rgb}{0.635,0.561,0.447}
\definecolor{moving_object}{rgb}{0.678,0.102,0.396}
\definecolor{signage}{rgb}{0.922,0.769,0.082}
\definecolor{barrier}{rgb}{0.596,0.510,0.604}
\definecolor{twowheels}{rgb}{0.329,0.027,0.643}
\definecolor{personaltransport}{rgb}{0.000,0.000,0.439}
\definecolor{publictransport}{rgb}{0.000,0.275,0.392}
\definecolor{pavement}{rgb}{0.761,0.200,0.733}
\definecolor{ground_m}{rgb}{0.514,0.796,0.431}
\definecolor{thinobject}{rgb}{0.776,0.686,0.427}
\definecolor{structure}{rgb}{0.463,0.408,0.467}
\definecolor{human}{rgb}{0.933,0.055,0.165}
\definecolor{vehicle}{rgb}{0.231,0.137,0.537}

\renewcommand{\imgWidth}{0.14\textwidth}
\begin{figure}[t]
\newcolumntype{Y}{>{\centering\arraybackslash}X}
\centering
\begin{subfigure}{.6em}
\scriptsize\rotatebox{90}{Step 0~~~~~~}
\end{subfigure}%
\begin{subfigure}{\textwidth}
\hspace*{.1em}%
\begin{subfigure}{\imgWidth}
    \caption{RGB}
\includegraphics[width=\textwidth]{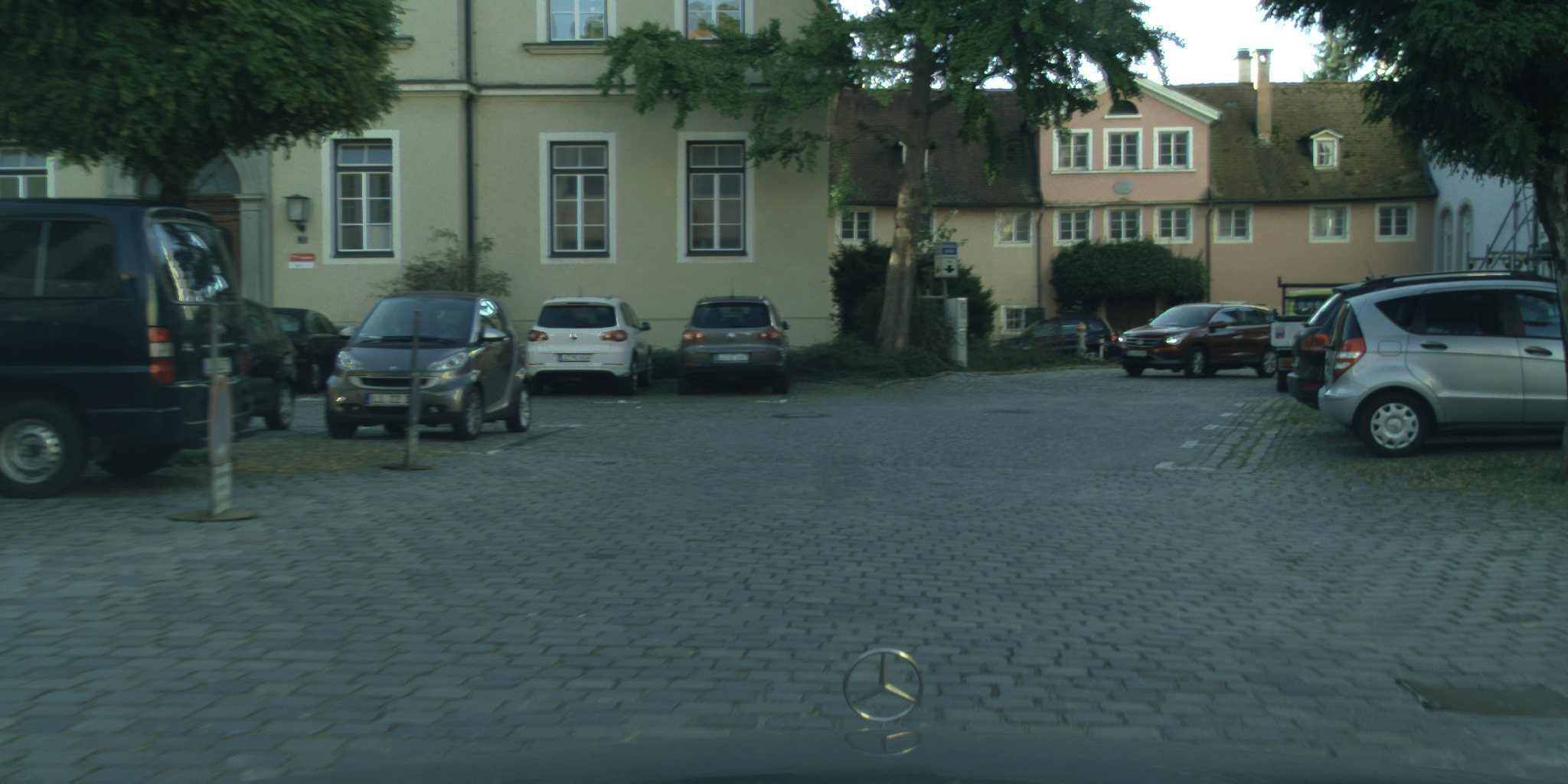}
\end{subfigure}%
\begin{subfigure}{\imgWidth}
    \caption{GT}
    \includegraphics[width=\textwidth]{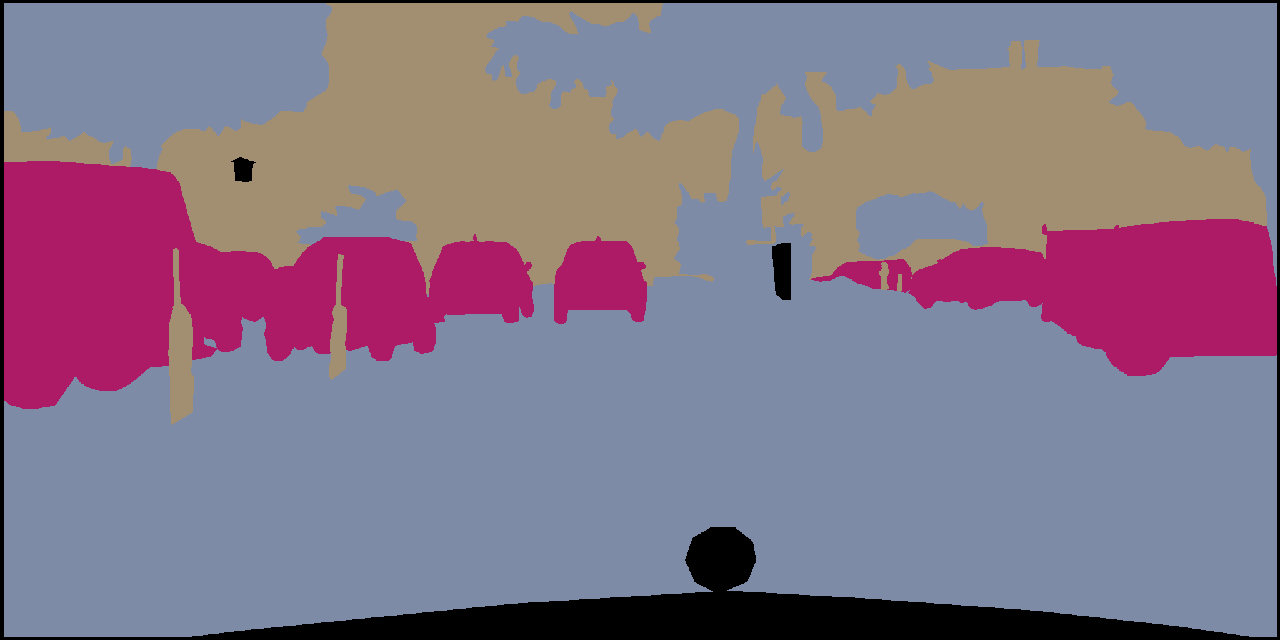}
\end{subfigure}%
\begin{subfigure}{\imgWidth}
    \caption{S.O.}
\includegraphics[width=\textwidth]{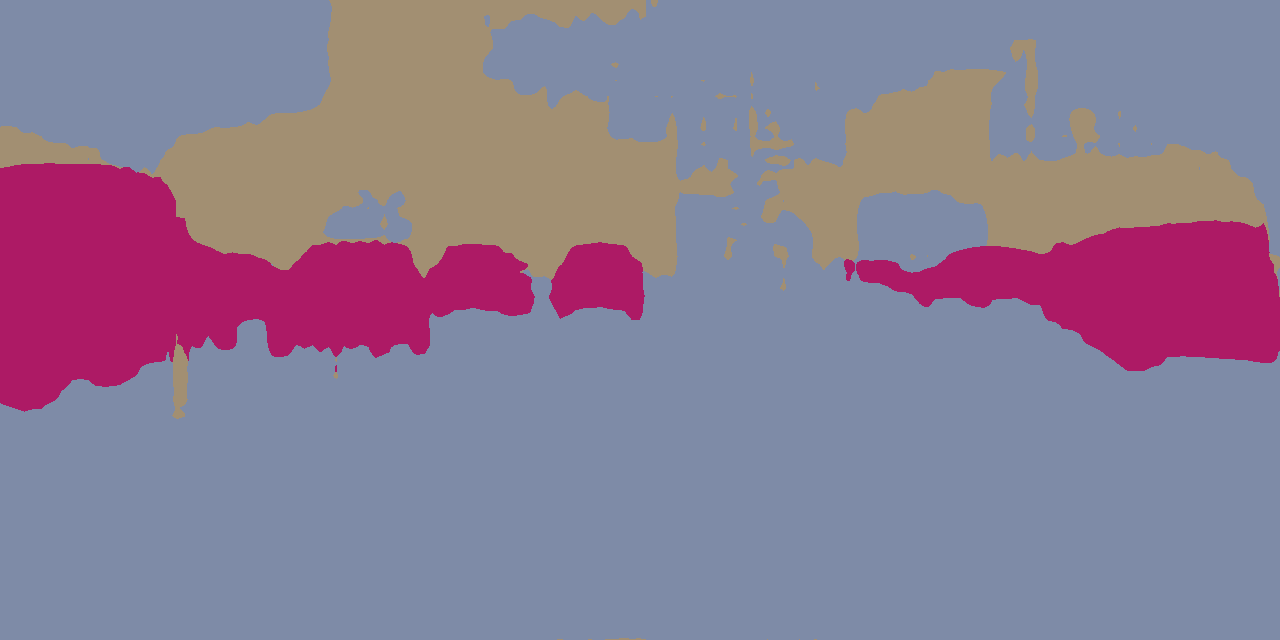}
\end{subfigure}%
\begin{subfigure}{\imgWidth}
    \caption{\footnotesize{MiB~\cite{cermelli2020modeling}}\vphantom{I}}
\includegraphics[width=\textwidth]{figures/qualitative_gta_city/step0_skdc.png}
\end{subfigure}%
\begin{subfigure}{\imgWidth}
    \caption{SKDC}
\includegraphics[width=\textwidth]{figures/qualitative_gta_city/step0_skdc.png}
\end{subfigure}%
\begin{subfigure}{\imgWidth}
    \caption{\footnotesize{MSIW~\cite{Chen2019}}\vphantom{I}}
    \includegraphics[width=\textwidth]{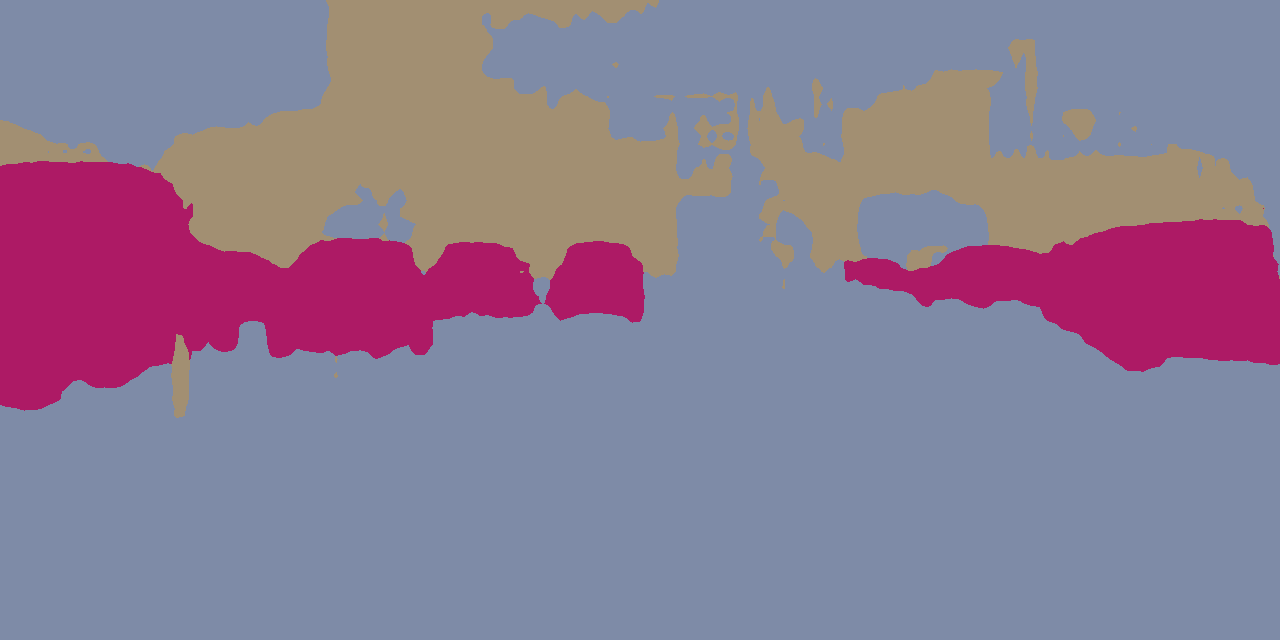}
\end{subfigure}%
\begin{subfigure}{\imgWidth}
    \caption{CCDA}
\includegraphics[width=\textwidth]{figures/qualitative_gta_city/step0_msiw.png}
\end{subfigure}%
\end{subfigure}

\begin{subfigure}{\textwidth}
    \tiny
    \begin{tabularx}{\textwidth}{YYYY}
    \cellcolor{unlabelled} \textcolor{white}{unlabelled} & \cellcolor{backgroud} \textcolor{white}{backgroud} & \cellcolor{static_object} \textcolor{white}{static object} & \cellcolor{moving_object} \textcolor{white}{moving object}
    \end{tabularx}
\end{subfigure}

\begin{subfigure}{.6em}
\scriptsize\rotatebox{90}{Step 1}
\end{subfigure}%
\begin{subfigure}{\textwidth}
\begin{subfigure}{\imgWidth}
\includegraphics[width=\textwidth]{figures/qualitative_gta_city/rgb.jpeg}
\end{subfigure}%
\begin{subfigure}{\imgWidth}
\includegraphics[width=\textwidth]{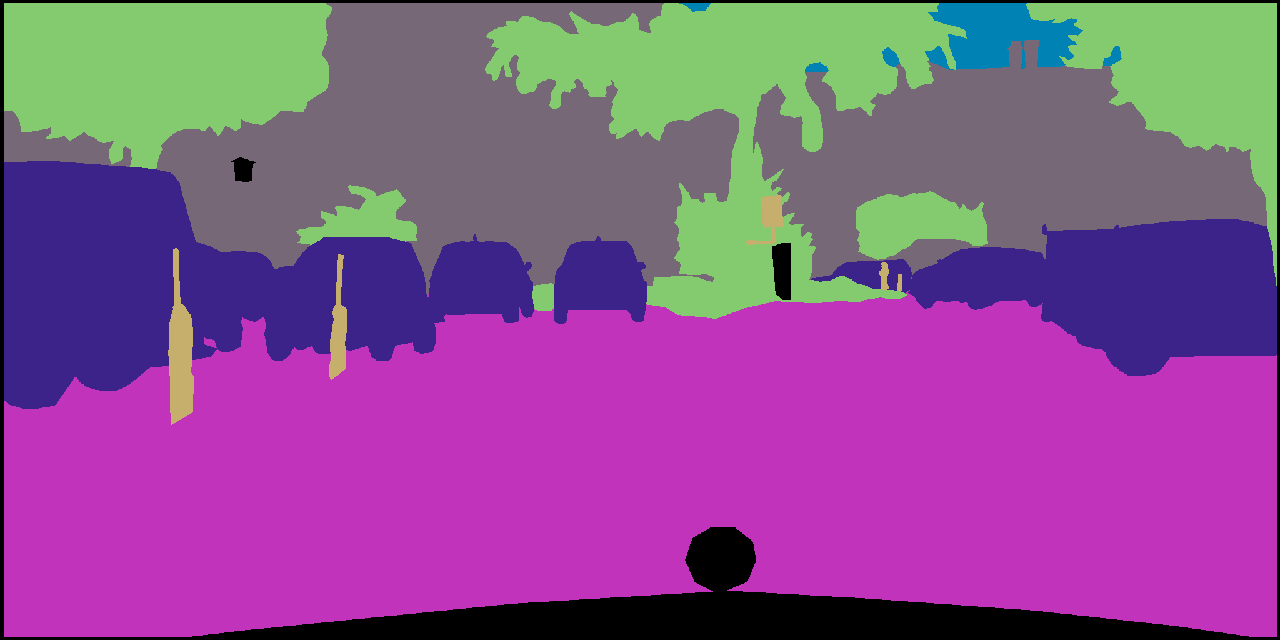}
\end{subfigure}%
\begin{subfigure}{\imgWidth}
\includegraphics[width=\textwidth]{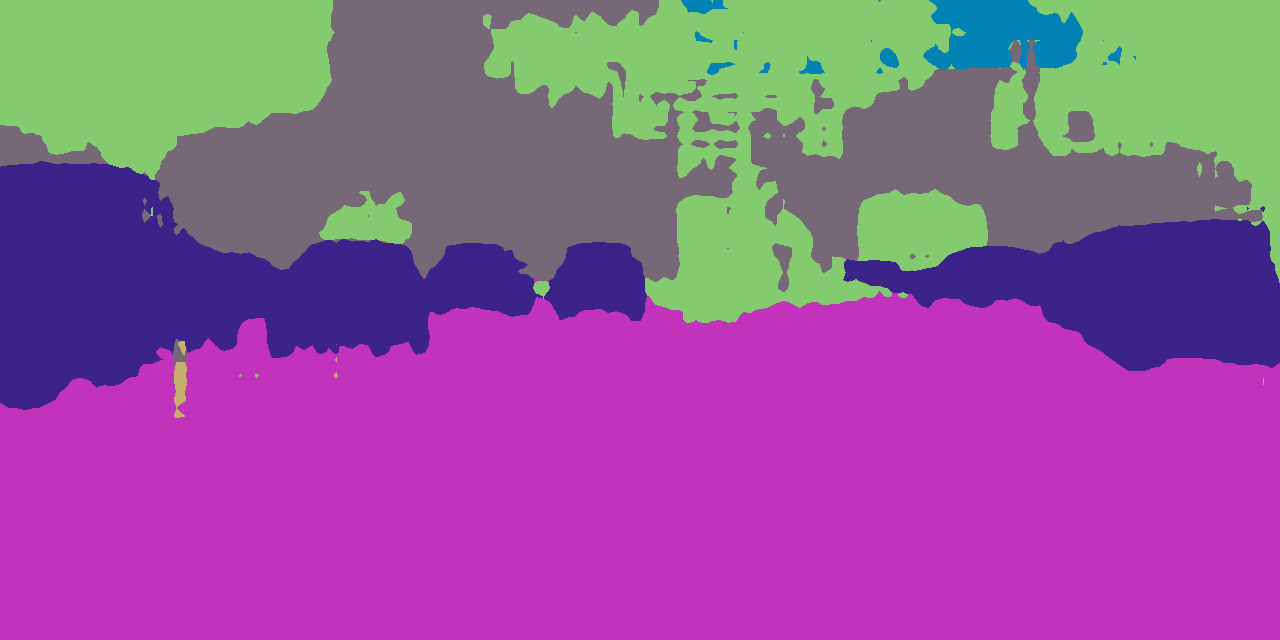}
\end{subfigure}%
\begin{subfigure}{\imgWidth}
\includegraphics[width=\textwidth]{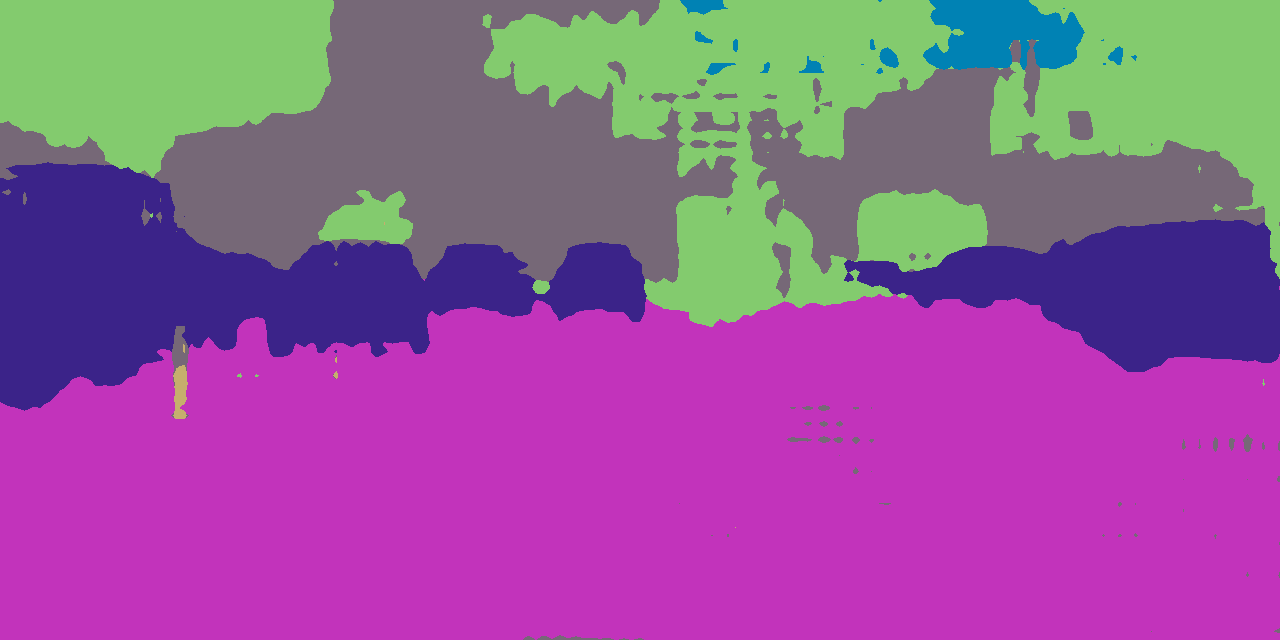}
\end{subfigure}%
\begin{subfigure}{\imgWidth}
\includegraphics[width=\textwidth]{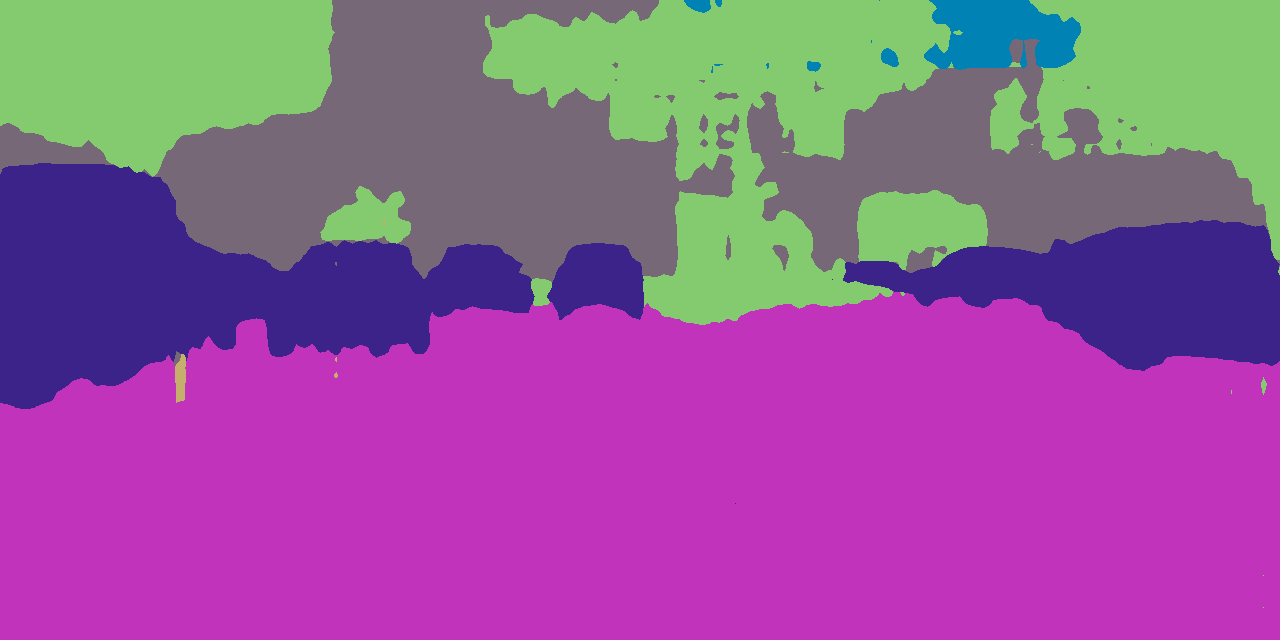}
\end{subfigure}%
\begin{subfigure}{\imgWidth}
\includegraphics[width=\textwidth]{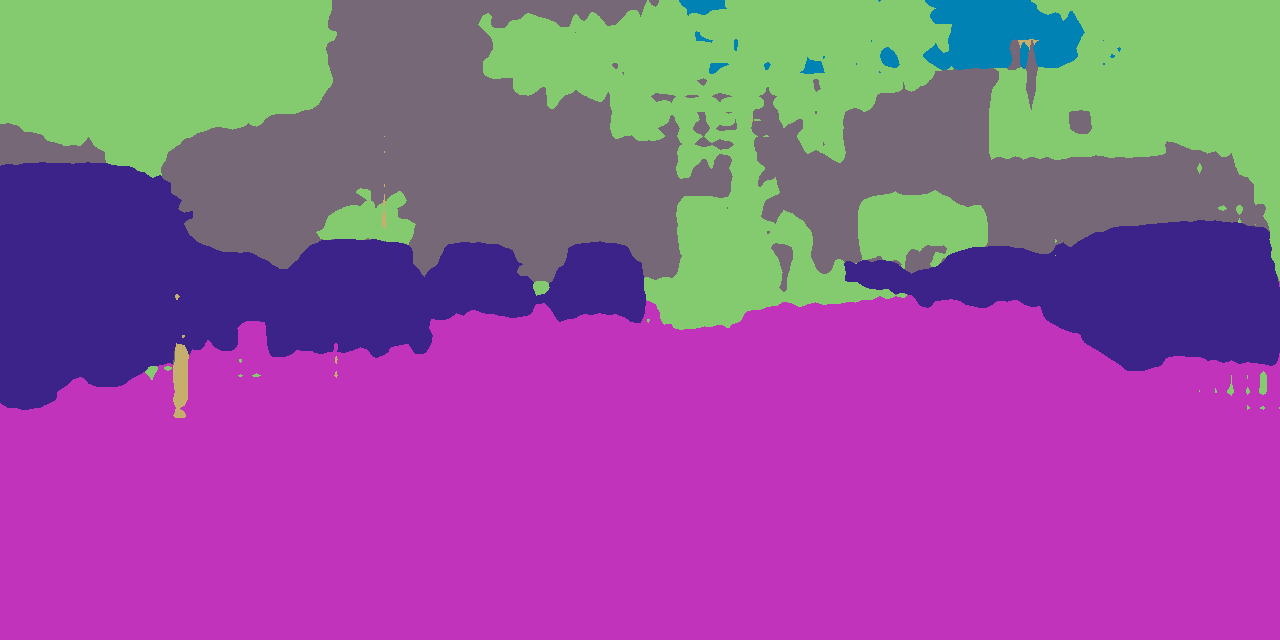}
\end{subfigure}%
\begin{subfigure}{\imgWidth}
\includegraphics[width=\textwidth]{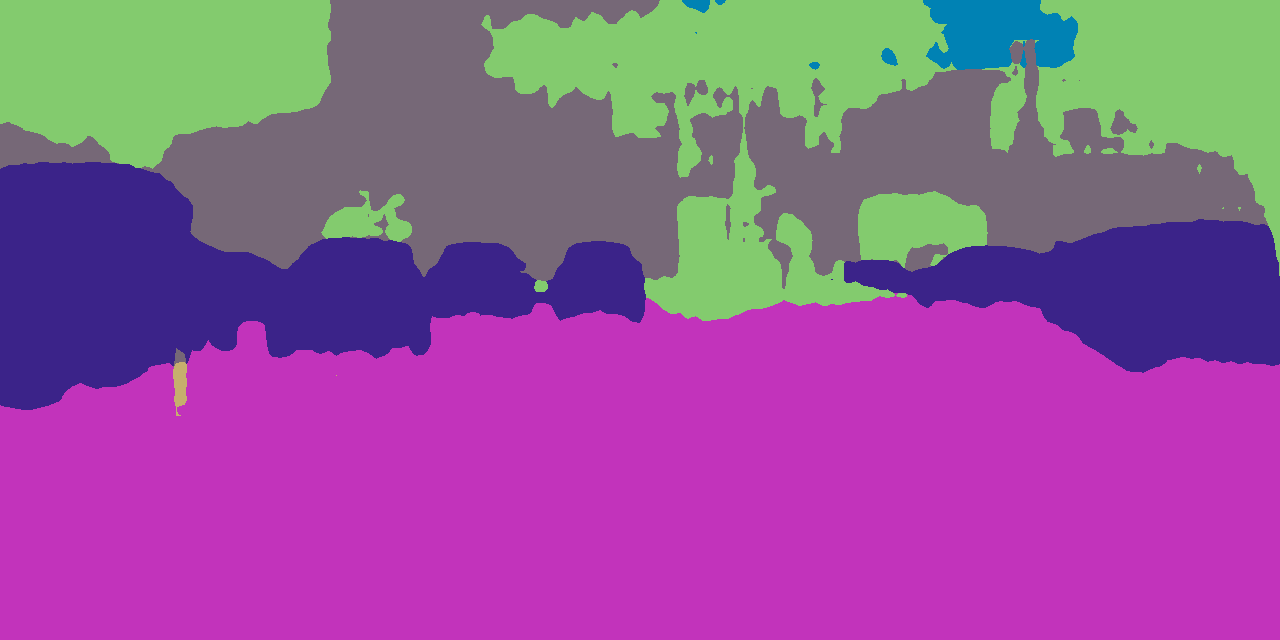}
\end{subfigure}%
\end{subfigure}

\begin{subfigure}{\textwidth}
    \tiny
    \begin{tabularx}{\textwidth}{YYYYYYY}
    \cellcolor{pavement} \textcolor{white}{pavement} & \cellcolor{sky} \textcolor{white}{sky} & \cellcolor{ground_m} \textcolor{black}{ground} & \cellcolor{thinobject} \textcolor{black}{thin ob.} & \cellcolor{structure} \textcolor{white}{structure} & \cellcolor{human} \textcolor{white}{human} & \cellcolor{vehicle} \textcolor{white}{vehicle}
    \end{tabularx}
\end{subfigure}

\begin{subfigure}{.6em}
\scriptsize\rotatebox{90}{Step 2}
\end{subfigure}%
\begin{subfigure}{\textwidth}
\hspace*{.2em}%
\begin{subfigure}{\imgWidth}
\includegraphics[width=\textwidth]{figures/qualitative_gta_city/rgb.jpeg}
\end{subfigure}%
\begin{subfigure}{\imgWidth}
\includegraphics[width=\textwidth]{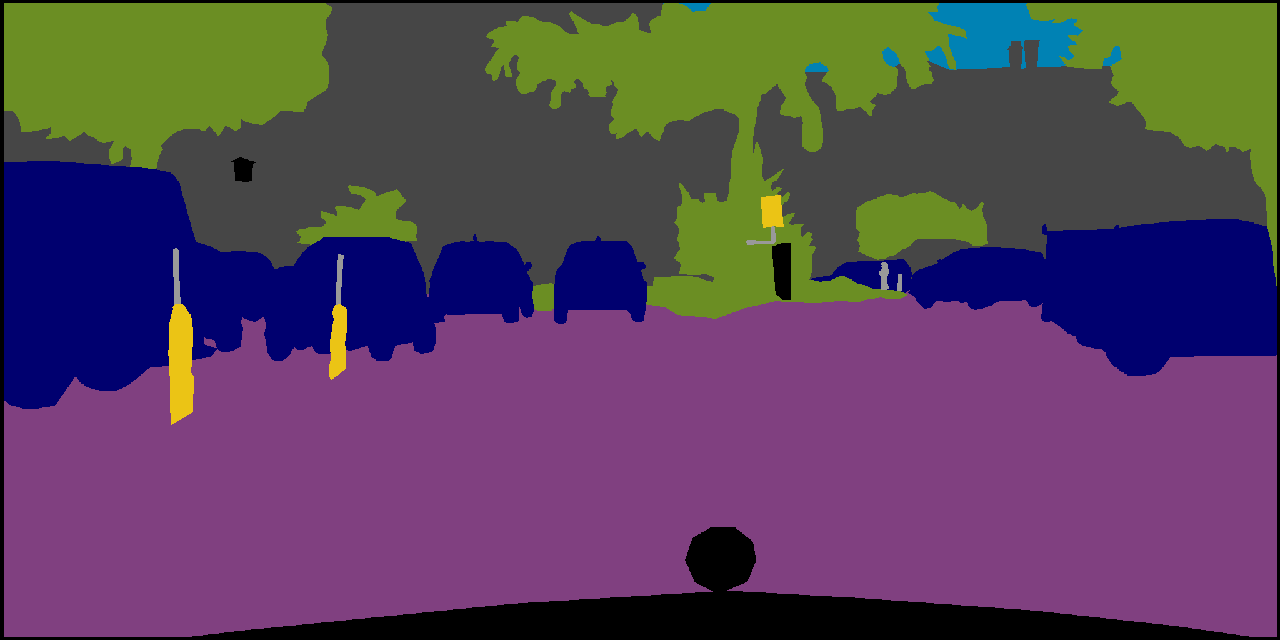}
\end{subfigure}%
\begin{subfigure}{\imgWidth}
\includegraphics[width=\textwidth]{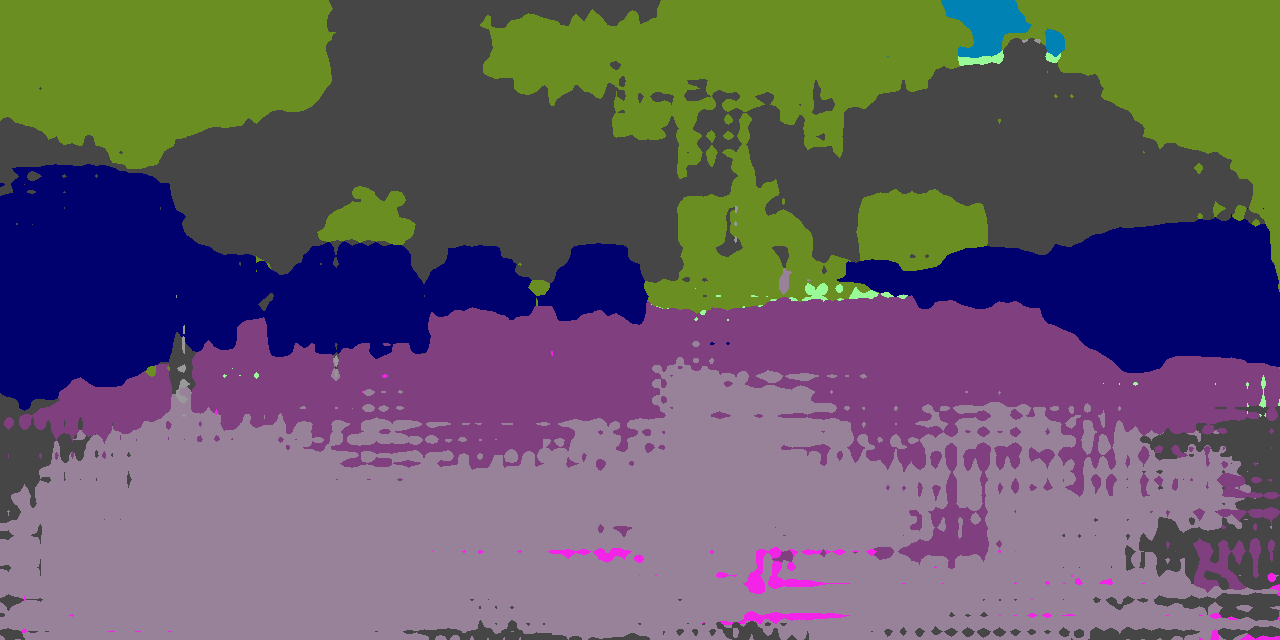}
\end{subfigure}%
\begin{subfigure}{\imgWidth}
\includegraphics[width=\textwidth]{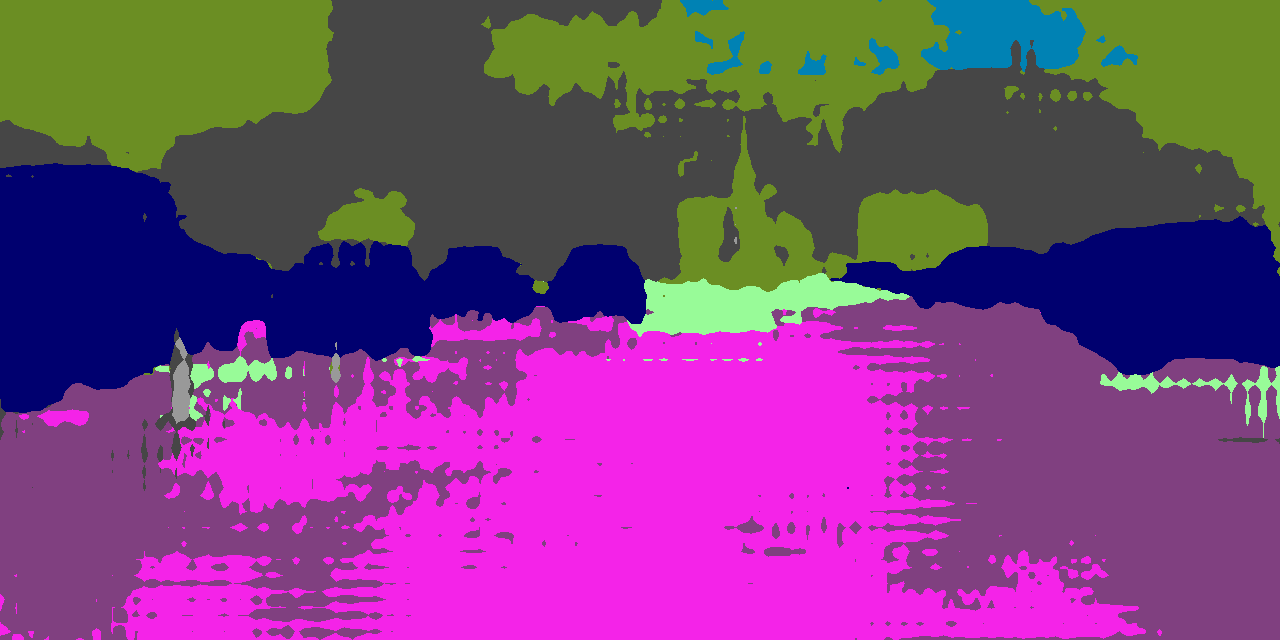}
\end{subfigure}%
\begin{subfigure}{\imgWidth}
\includegraphics[width=\textwidth]{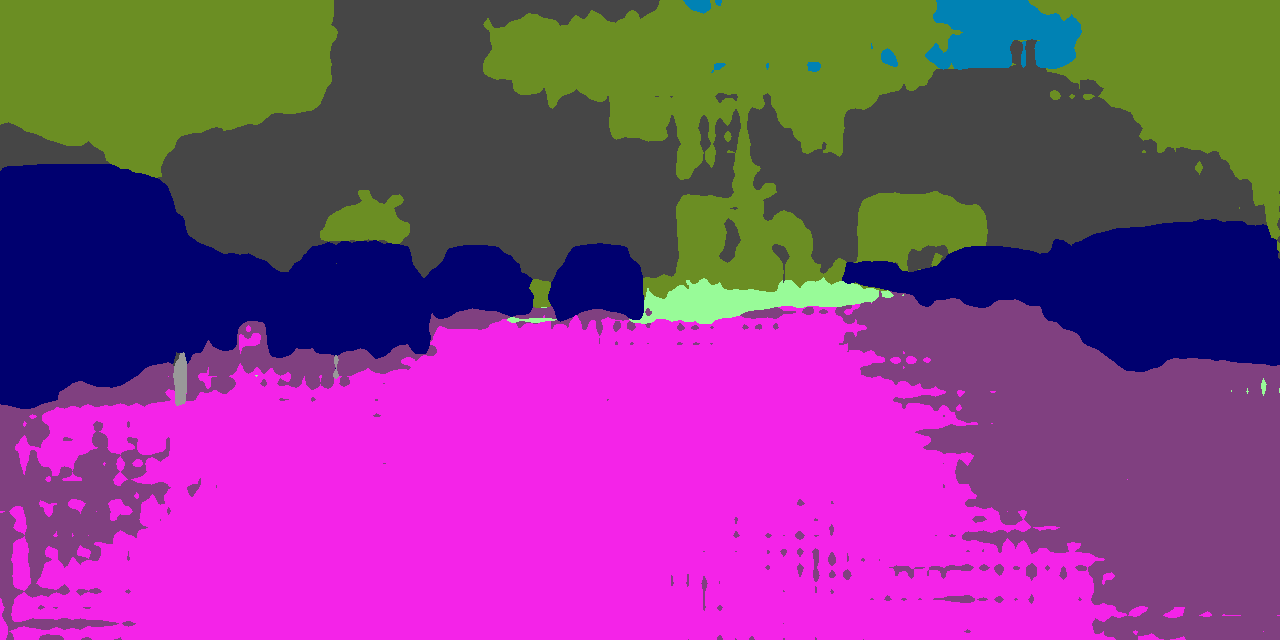}
\end{subfigure}%
\begin{subfigure}{\imgWidth}
\includegraphics[width=\textwidth]{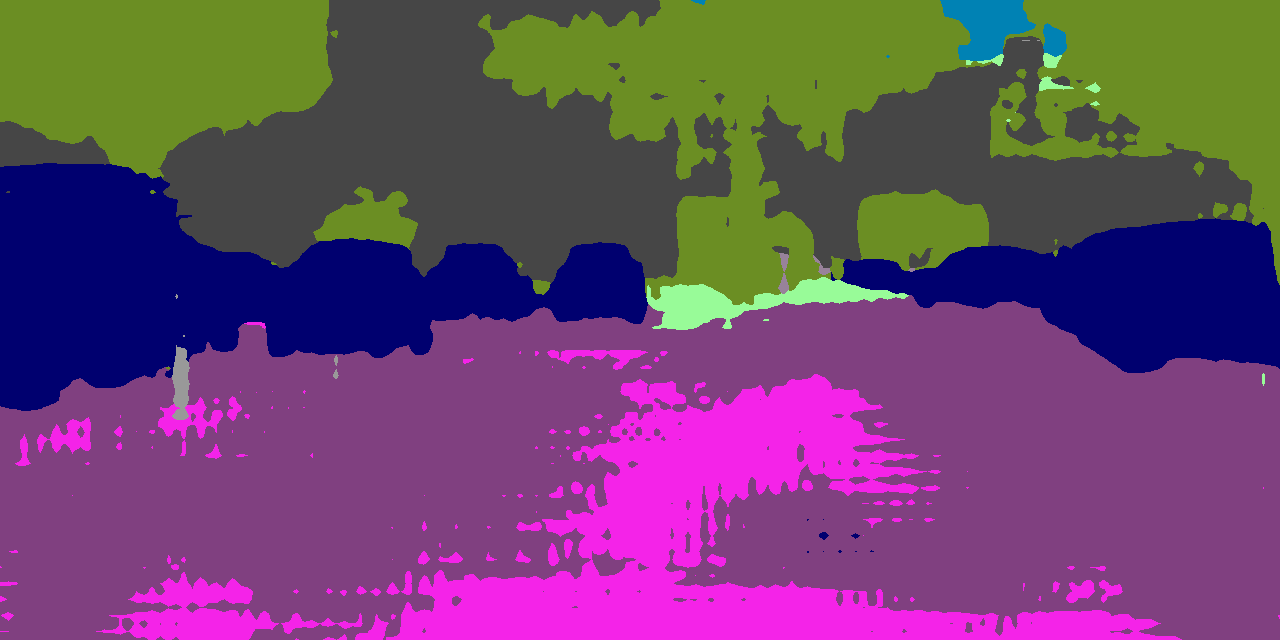}
\end{subfigure}%
\begin{subfigure}{\imgWidth}
\includegraphics[width=\textwidth]{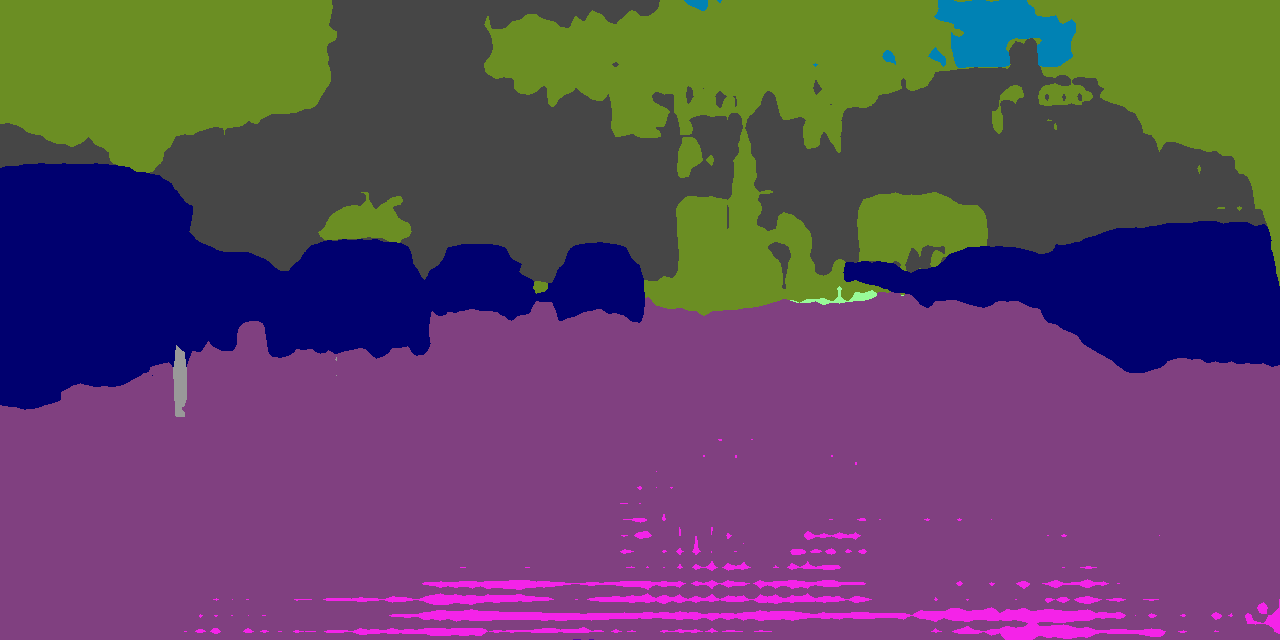}
\end{subfigure}%
\end{subfigure}

\begin{subfigure}{\textwidth}
    \tiny
    \begin{tabularx}{\textwidth}{YYYYYYYYY}
    \cellcolor{road} \textcolor{white}{road} & \cellcolor{sidewalk} \textcolor{white}{sidewalk} & \cellcolor{vegetation} \textcolor{white}{vegetation} & \cellcolor{terrain} \textcolor{black}{terrain} & 
    \cellcolor{pole} \textcolor{white}{pole} & \cellcolor{signage} \textcolor{black}{signage} & \cellcolor{building} \textcolor{white}{building} 
\end{tabularx}
\begin{tabularx}{\textwidth}{YYYYYYY}
        \cellcolor{barrier} \textcolor{white}{barrier} & \cellcolor{person} \textcolor{white}{person} & \cellcolor{rider} \textcolor{white}{rider} &
        \cellcolor{twowheels} \textcolor{white}{two wheels} & \cellcolor{personaltransport} \textcolor{white}{per. tr.} & \cellcolor{publictransport} \textcolor{white}{pub. tr.} 
\end{tabularx}
\end{subfigure}

\begin{subfigure}{.6em}
\scriptsize\rotatebox{90}{Step 3}
\end{subfigure}%
\begin{subfigure}{\textwidth}
\hspace*{.2em}%
\begin{subfigure}{\imgWidth}
\includegraphics[width=\textwidth]{figures/qualitative_gta_city/rgb.jpeg}
\end{subfigure}%
\begin{subfigure}{\imgWidth}
\includegraphics[width=\textwidth]{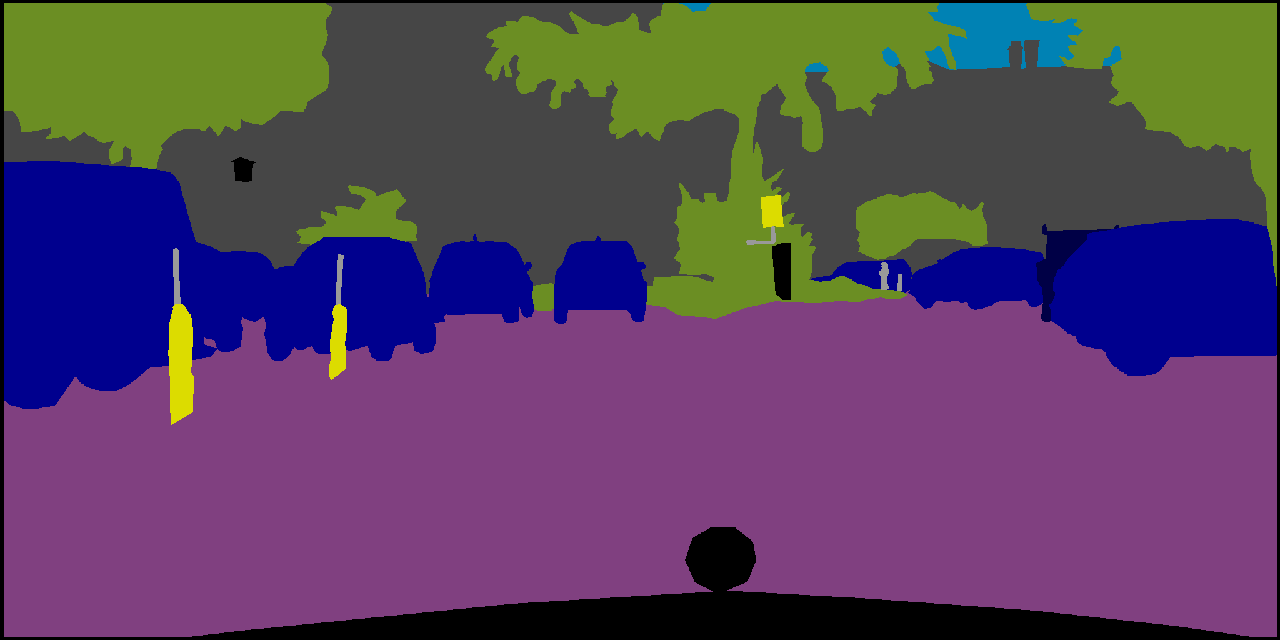}
\end{subfigure}%
\begin{subfigure}{\imgWidth}
\includegraphics[width=\textwidth]{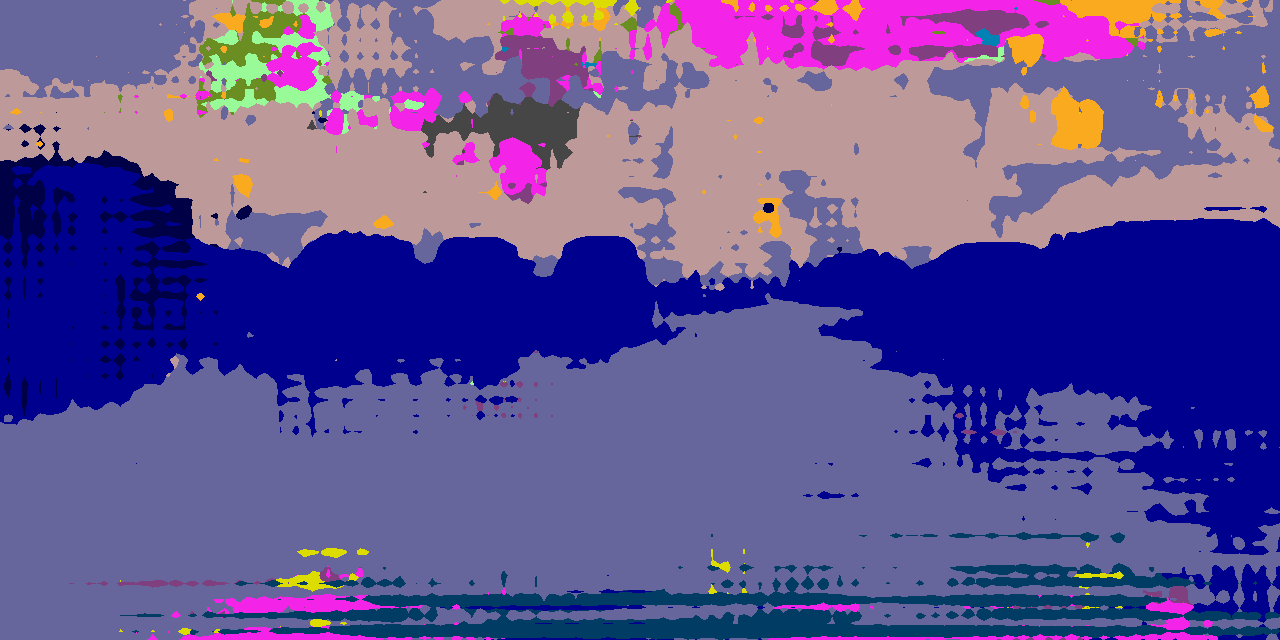}
\end{subfigure}%
\begin{subfigure}{\imgWidth}
\includegraphics[width=\textwidth]{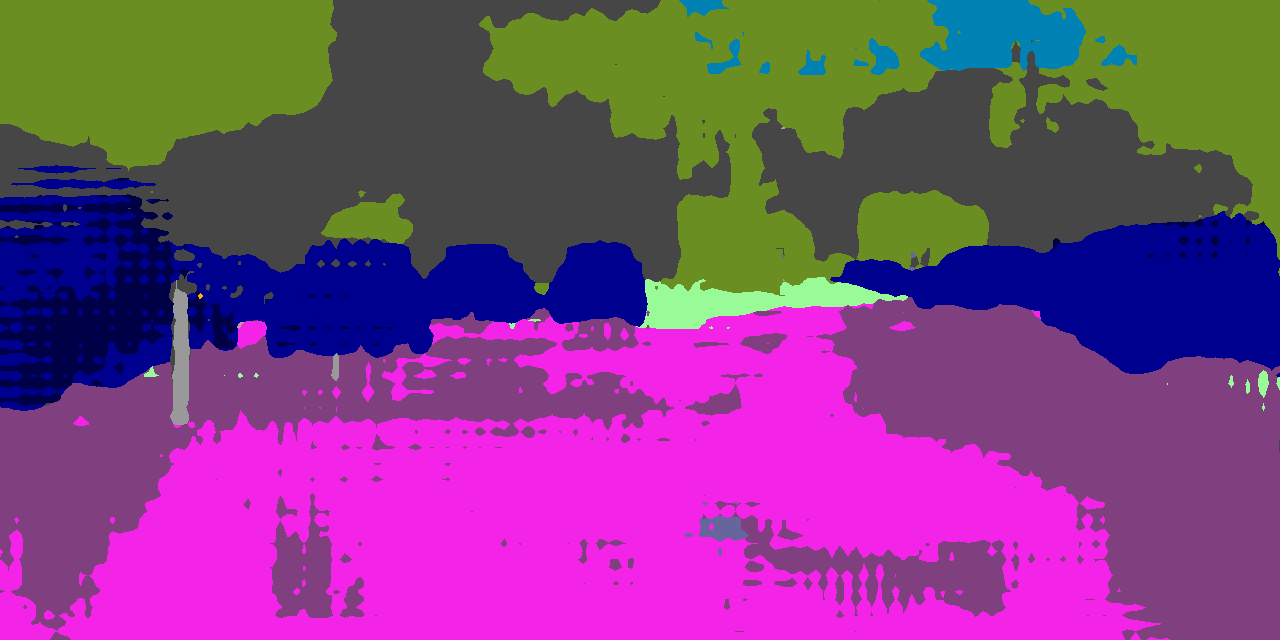}
\end{subfigure}%
\begin{subfigure}{\imgWidth}
\includegraphics[width=\textwidth]{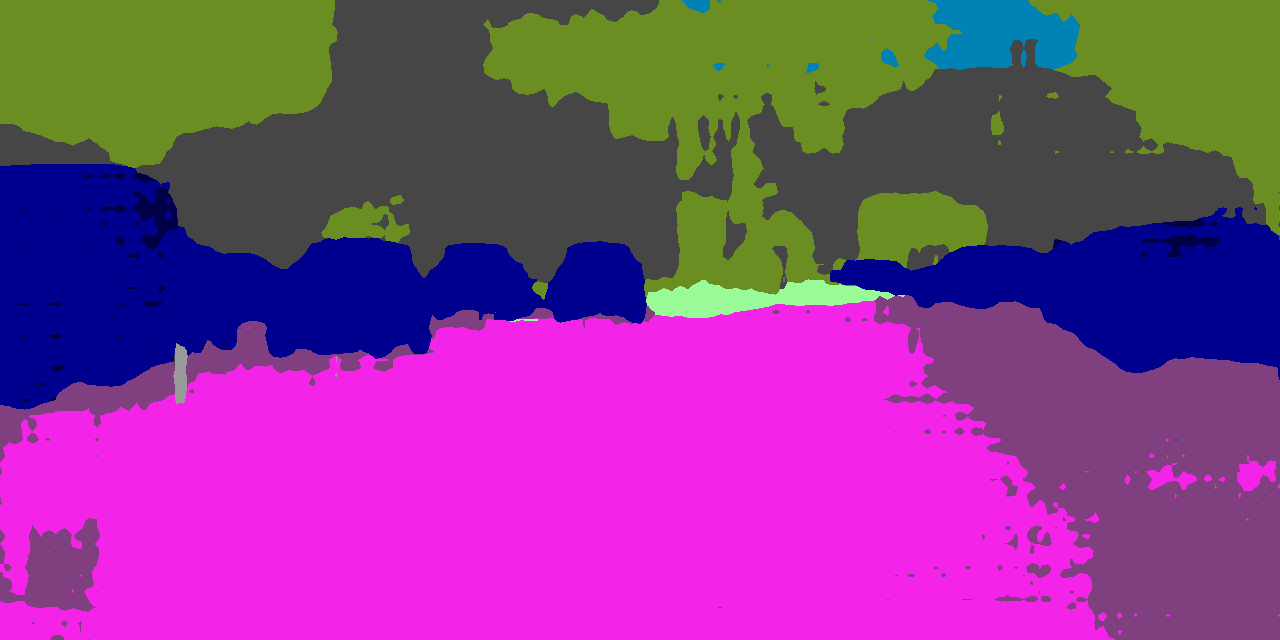}
\end{subfigure}%
\begin{subfigure}{\imgWidth}
\includegraphics[width=\textwidth]{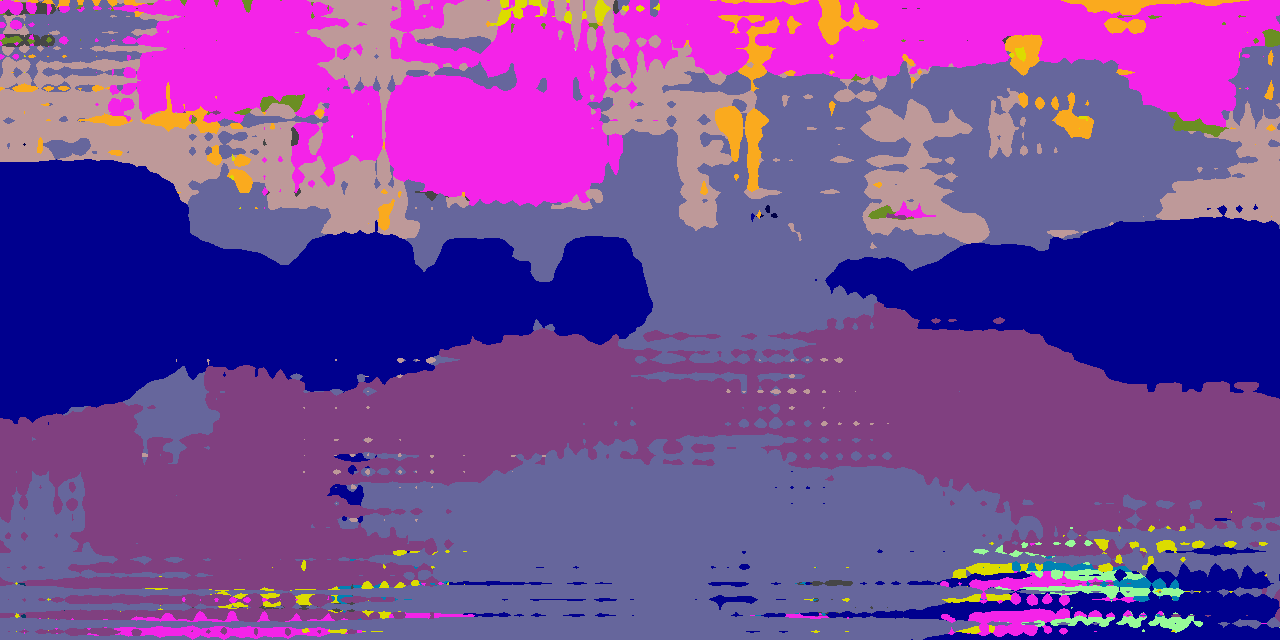}
\end{subfigure}%
\begin{subfigure}{\imgWidth}
\includegraphics[width=\textwidth]{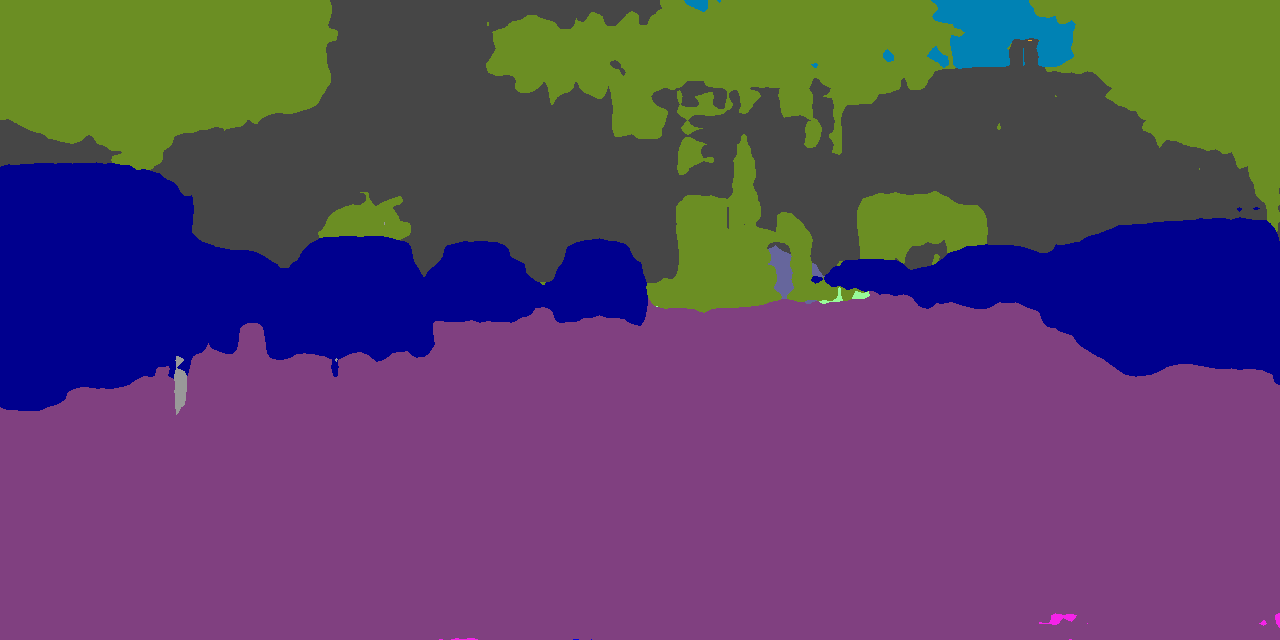}
\end{subfigure}%
\end{subfigure}
\begin{subfigure}{\textwidth}
    \tiny
    \begin{tabularx}{\textwidth}{YYYYYYYYYY}
    \cellcolor{tlight} \textcolor{black}{t. light} & \cellcolor{tsign} \textcolor{black}{t. sign} & \cellcolor{fence} \textcolor{black}{fence} & 
    \cellcolor{wall} \textcolor{white}{wall} & \cellcolor{bicycle} \textcolor{white}{bicycle} & 
    \cellcolor{motorbike} \textcolor{white}{mbike} & \cellcolor{car} \textcolor{white}{car} & \cellcolor{truck} \textcolor{white}{truck} &
    \cellcolor{bus} \textcolor{white}{bus} & \cellcolor{train} \textcolor{white}{train}
    \end{tabularx}
\end{subfigure}
\caption{GTA5$\rightarrow$Cityscapes qualitative results. GT refers to the complete ground truth for the given step.}
\label{fig:quali_city}
\end{figure}

\begin{table}[h!]
\centering
\resizebox{\textwidth}{!}{%
\setlength{\tabcolsep}{3pt}
\begin{tabular}{c||ccccc|cccccc|cccccccc|c}
\rotatebox{45}{step} & \rotatebox{90}{road} & \rotatebox{90}{sidewalk} & \rotatebox{90}{sky} & \rotatebox{90}{vegetation} & \rotatebox{90}{terrain} & \rotatebox{90}{pole} & \rotatebox{90}{traffic light} & \rotatebox{90}{traffic sign} & \rotatebox{90}{building} & \rotatebox{90}{fence} & \rotatebox{90}{wall} & \rotatebox{90}{person} & \rotatebox{90}{rider} & \rotatebox{90}{bicycle} & \rotatebox{90}{motorbike} & \rotatebox{90}{car} & \rotatebox{90}{truck} & \rotatebox{90}{bus} & \rotatebox{90}{train} & mIoU \\
\toprule
\toprule
0 & \multicolumn{5}{c}{92.9} & \multicolumn{6}{|c}{84.2} & \multicolumn{8}{|c|}{79.2} & 85.4 \\
\midrule
1 & \multicolumn{2}{c}{95.6} & 74.9 & \multicolumn{2}{c}{80.8} & \multicolumn{3}{|c}{9.7}            & \multicolumn{3}{c}{78.0} & \multicolumn{2}{|c}{55.1} & \multicolumn{6}{c|}{81.3} & 67.9  \\
\midrule
2 & 53.2 & 15.9 & 75.0 & 80.6 & 35.9 & 14.3 & \multicolumn{2}{c}{8.4} & 73.9 &  \multicolumn{2}{c|}{10.5} & 49.9 & 0.0 & \multicolumn{2}{c}{3.1} & \multicolumn{2}{c}{79.2} & \multicolumn{2}{c|}{21.0} & 37.2  \\
\midrule
3 & 50.8 & 14.1 & 74.7 & 77.1 & 29.8 & 12.9 & 27.1 & 16.4 & 77.9 & 12.6 & 21.7 & 50.2 & 0.0 & 6.7 & 12.5 & 74.4 & 25.4 & 27.2 & 17.7  & 33.1 \\
\toprule
\end{tabular}%
\setlength{\tabcolsep}{6pt}
}
\caption{\textit{GTA5}$\rightarrow$\textit{Cityscapes}, CCDA per-class per-step IoU results on the Cityscapes test set.}  
\label{tab:gta_city_quantitative_per_class}
\end{table}

On the \textit{Cityscapes} dataset the upper bound on the  mIoU score given by the target offline training (JTO) is $68.6\%$, while the coarse-to-fine training on 
the target dataset (TNC)  leads to a reduction of $2.1\%$ with a score of $66.5\%$.

When using only supervised source data (\textit{GTA5$\rightarrow$Cityscapes} setup) we can see that 
the Source Only method suffers greatly from both catastrophic forgetting and domain shift, obtaining very low score of $4.5\%$. 
The domain adaptation strategy of MSIW  \cite{Chen2019} is able to only marginally improve performances, reaching $6.9\%$ with an improvement of $2.4\%$. 
 Such results, shows how a  continual learning approach is needed to properly tackle the task: indeed, MiB \cite{cermelli2020modeling} reaches $26.5\%$, with an improvement of $22\%$ points over the baseline.
Nevertheless, both our strategies are able to surpass such performance: SKDC reaches a modest $30.4\%$, while our full approach (CCDA) further surpasses it with a mIoU score of $33.1\%$. This corresponds to an improvement of $6.6\%$ with respect to the best competitor (MiB). We further remark how the difference in score between CCDA and SKDC ($2.7\%$) is higher than the one between MSIW and Source Only, meaning that our target self-supervised continual approach was able to improve the entropy minimization effect of MSIW, allowing to achieve better domain transfer capabilities. 

In Table~\ref{tab:main} we can also see that MSIW surpasses our strategy at step $2$, this is due to the high number of new classes present in such step, which are generated by splitting of most existing classes in the previous step (there are $13$ new classes, see Figure~\ref{fig:train_classes}). This allows the improvement in domain shift to surpass the performance loss caused by the catastrophic forgetting. However, at the next step the capability of preserving old knowledge is required and the superior performance of our approach are clearly visible.
In the detailed results reported in Table~\ref{tab:gta_city_quantitative_per_class} we can see that an important component of the performance degradation derives from the \textit{thin object} class, being it very difficult. Another cause of degradation is the \textit{rider} class, which is confused for \textit{person} leading to a drop in performance from $55.1\%$ in class \textit{human} at  step $1$ to $0\%$ when the class is split in step $2$. 

From the qualitative results reported in Figure~\ref{fig:quali_city} we can appreciate how our CCDA strategy is consistently better than the competitors, in all the incremental steps. In particular, we can notice how in the third and last step CCDA is the only method that can correctly classify the \textit{road} in the bottom half of the image, which the other methods confuse with \textit{sidewalk}, \textit{wall} or \textit{car}. At the same time, we can notice how the shape of the parked cars on the sides of the image are much better defined in our method.
Furthermore, also the visual results show a dramatic decrease in performance due to the catastrophic forgetting for Source Only and MSIW methods, and how such effect is strongly reduced in  methods that exploit continual learning strategies (MiB, SKDC, CCDA). 

\subsection{GTA5\texorpdfstring{$\rightarrow$}{->}IDD adaptation}

\definecolor{road}{rgb}{.502,.251,.502}
\definecolor{sidewalk}{rgb}{.957,.137,.910}
\definecolor{building}{rgb}{.275,.275,.275}
\definecolor{wall}{rgb}{.4,.4,.612}
\definecolor{fence}{rgb}{.745,.6,.6}
\definecolor{pole}{rgb}{.6,.6,.6}
\definecolor{tlight}{rgb}{.980,.667,.118}
\definecolor{tsign}{rgb}{.863,.863,0}
\definecolor{vegetation}{rgb}{.420,.557,.137}
\definecolor{terrain}{rgb}{.596,.984,.596}
\definecolor{sky}{rgb}{0,.510,.706}
\definecolor{person}{rgb}{.863,.078,.235}
\definecolor{rider}{rgb}{1,0,0}
\definecolor{car}{rgb}{0,0,.557}
\definecolor{truck}{rgb}{0,0,.275}
\definecolor{bus}{rgb}{0,.235,.392}
\definecolor{train}{rgb}{0,.314,.392}
\definecolor{motorbike}{rgb}{0,0,.902}
\definecolor{bicycle}{rgb}{.467,.043,.125}
\definecolor{unlabelled}{rgb}{0,0,0}
\definecolor{backgroud}{rgb}{0.494,0.545,0.655}
\definecolor{static_object}{rgb}{0.635,0.561,0.447}
\definecolor{moving_object}{rgb}{0.678,0.102,0.396}
\definecolor{signage}{rgb}{0.922,0.769,0.082}
\definecolor{barrier}{rgb}{0.596,0.510,0.604}
\definecolor{twowheels}{rgb}{0.329,0.027,0.643}
\definecolor{personaltransport}{rgb}{0.000,0.000,0.439}
\definecolor{publictransport}{rgb}{0.000,0.275,0.392}
\definecolor{pavement}{rgb}{0.761,0.200,0.733}
\definecolor{ground_m}{rgb}{0.514,0.796,0.431}
\definecolor{thinobject}{rgb}{0.776,0.686,0.427}
\definecolor{structure}{rgb}{0.463,0.408,0.467}
\definecolor{human}{rgb}{0.933,0.055,0.165}
\definecolor{vehicle}{rgb}{0.231,0.137,0.537}

\begin{figure}[t]
\newcolumntype{Y}{>{\centering\arraybackslash}X}
\centering
\begin{subfigure}{.6em}
\scriptsize\rotatebox{90}{Step 0~~~~~~}
\end{subfigure}%
\begin{subfigure}{\textwidth}
\hspace*{.1em}%
\begin{subfigure}{\imgWidth}
    \caption{RGB}
\includegraphics[width=\textwidth]{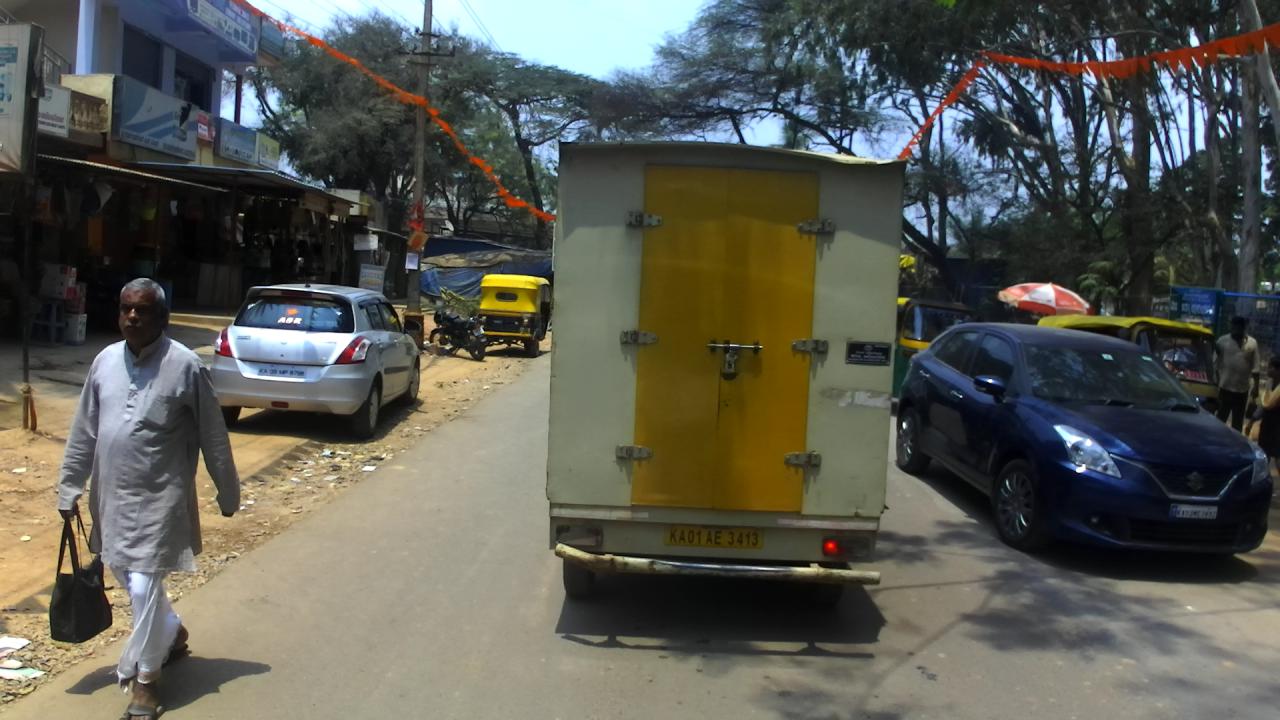}
\end{subfigure}%
\begin{subfigure}{\imgWidth}
    \caption{GT}
    \includegraphics[width=\textwidth]{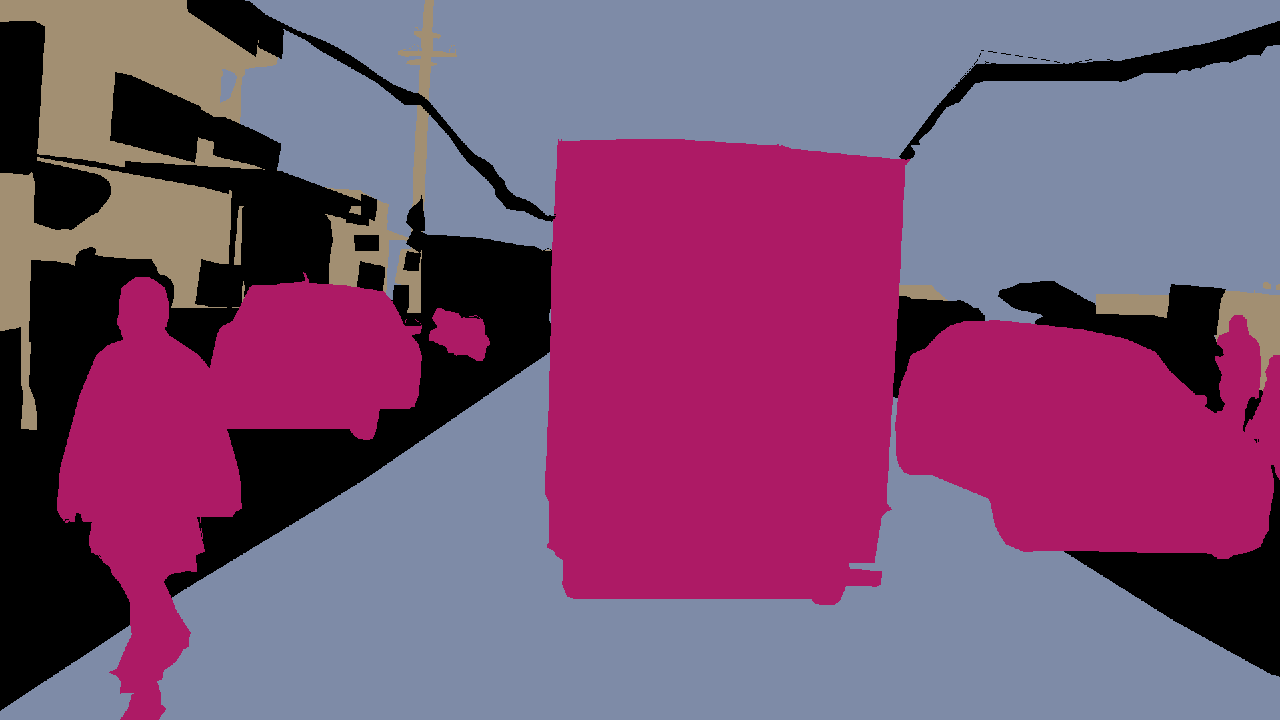}
\end{subfigure}%
\begin{subfigure}{\imgWidth}
    \caption{S.O.}
\includegraphics[width=\textwidth]{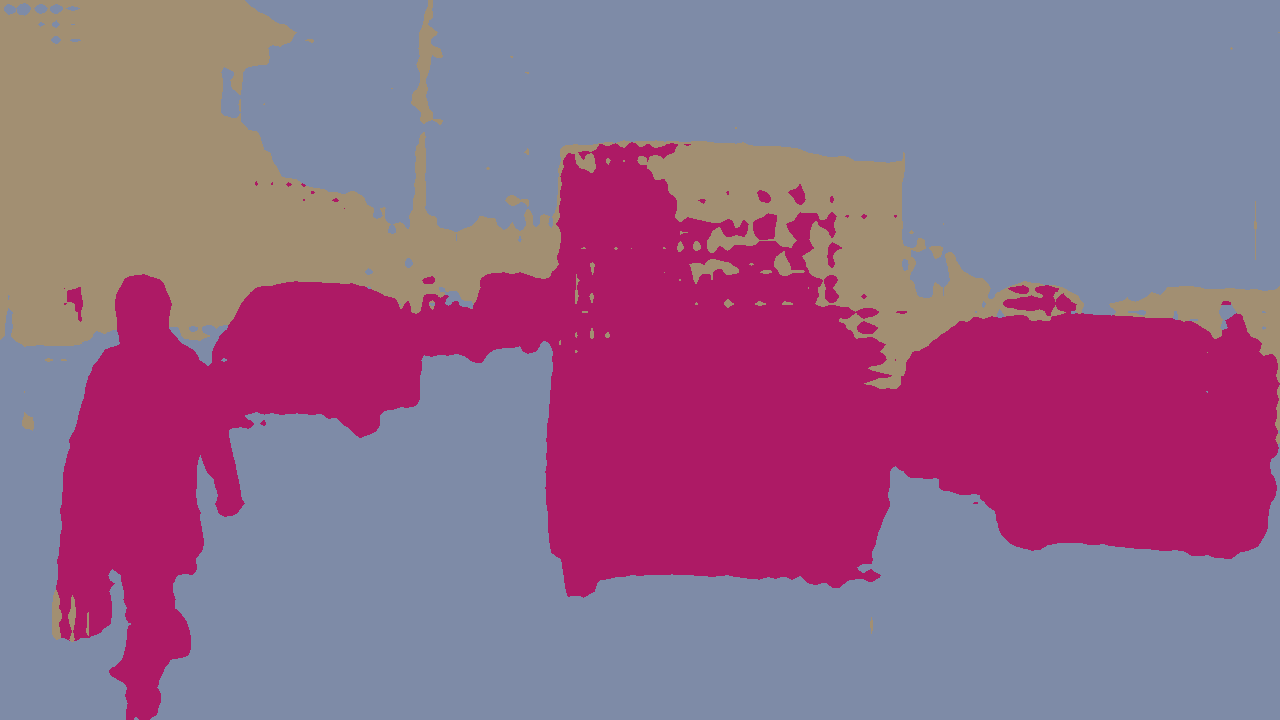}
\end{subfigure}%
\begin{subfigure}{\imgWidth}
    \caption{\footnotesize{MiB~\cite{cermelli2020modeling}}}
\includegraphics[width=\textwidth]{figures/qualitative_gta_idd/step0_skdc.png}
\end{subfigure}%
\begin{subfigure}{\imgWidth}
    \caption{SKDC}
\includegraphics[width=\textwidth]{figures/qualitative_gta_idd/step0_skdc.png}
\end{subfigure}%
\begin{subfigure}{\imgWidth}
    \caption{\footnotesize{MSIW~\cite{Chen2019}}}
    \includegraphics[width=\textwidth]{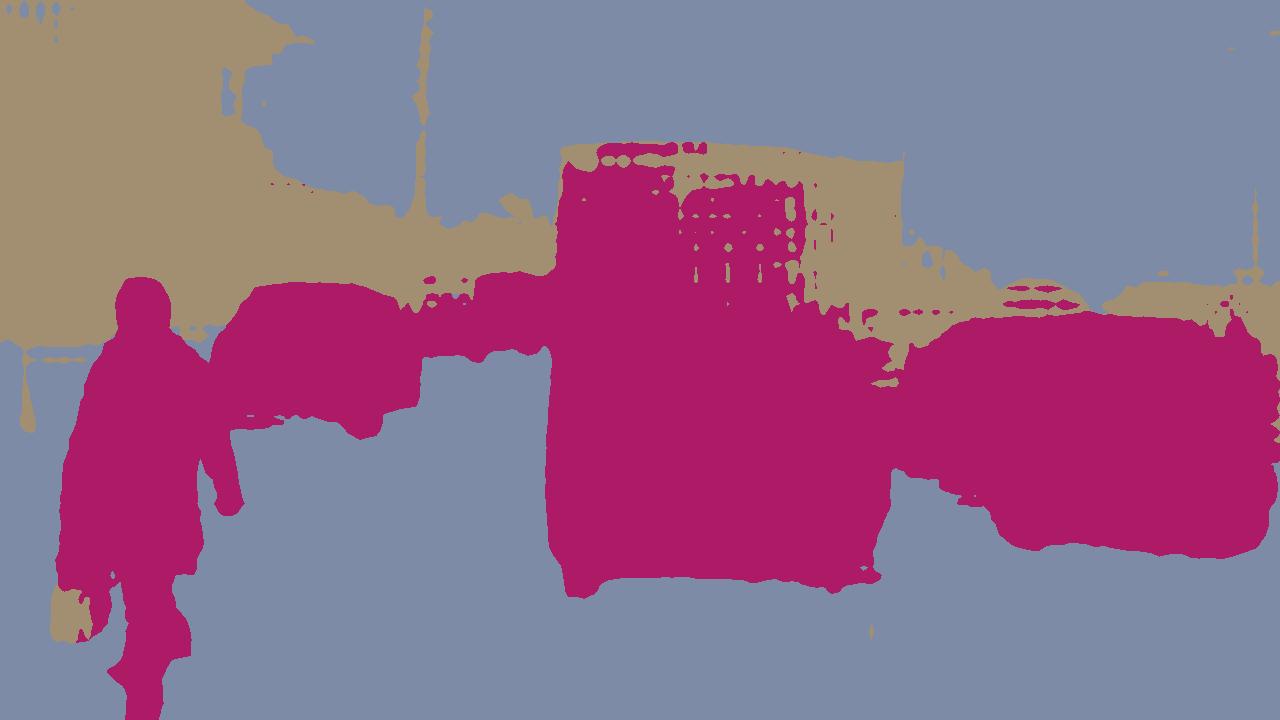}
\end{subfigure}%
\begin{subfigure}{\imgWidth}
    \caption{CCDA}
\includegraphics[width=\textwidth]{figures/qualitative_gta_idd/step0_ccda.png} 
\end{subfigure}%
\end{subfigure}

\begin{subfigure}{\textwidth}
    \tiny
    \begin{tabularx}{\textwidth}{YYYY}
    \cellcolor{unlabelled} \textcolor{white}{unlabelled} & \cellcolor{backgroud} \textcolor{white}{backgroud} & \cellcolor{static_object} \textcolor{white}{static object} & \cellcolor{moving_object} \textcolor{white}{moving object}
    \end{tabularx}
\end{subfigure}

\begin{subfigure}{.6em}
\scriptsize\rotatebox{90}{Step 1}
\end{subfigure}%
\begin{subfigure}{\textwidth}
\begin{subfigure}{\imgWidth}
\includegraphics[width=\textwidth]{figures/qualitative_gta_idd/rgb.jpeg}
\end{subfigure}%
\begin{subfigure}{\imgWidth}
\includegraphics[width=\textwidth]{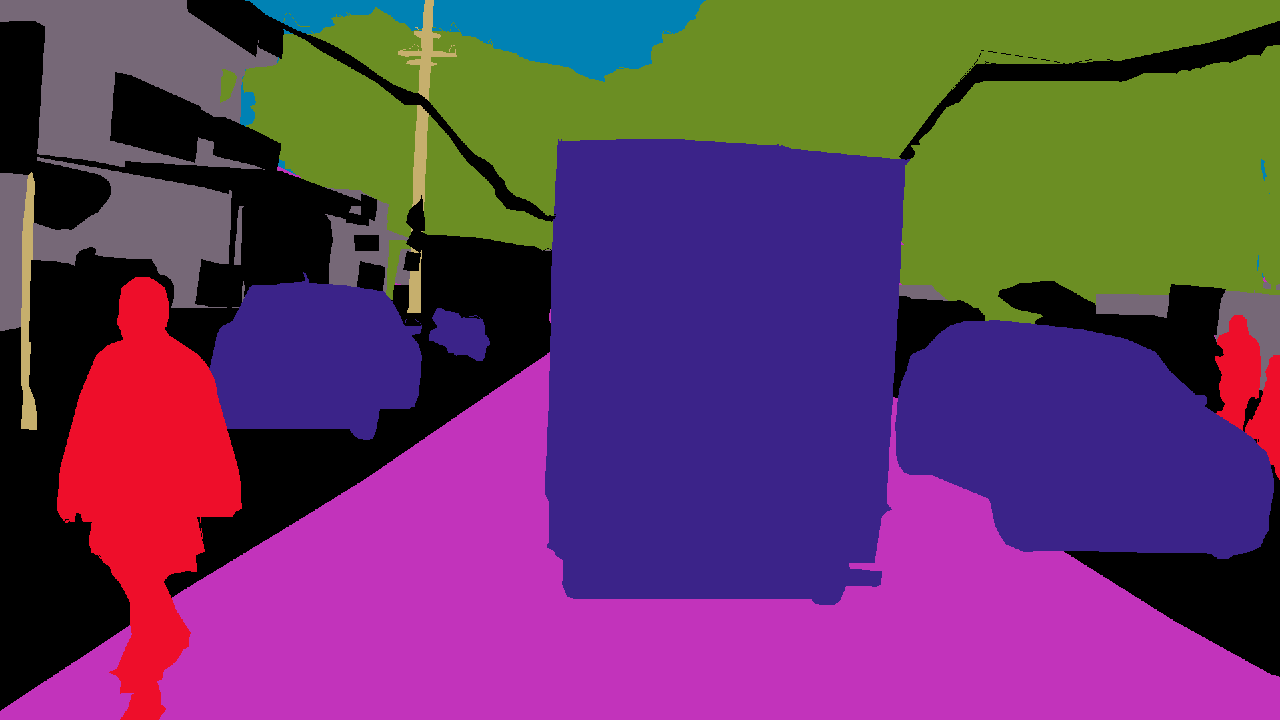}
\end{subfigure}%
\begin{subfigure}{\imgWidth}
\includegraphics[width=\textwidth]{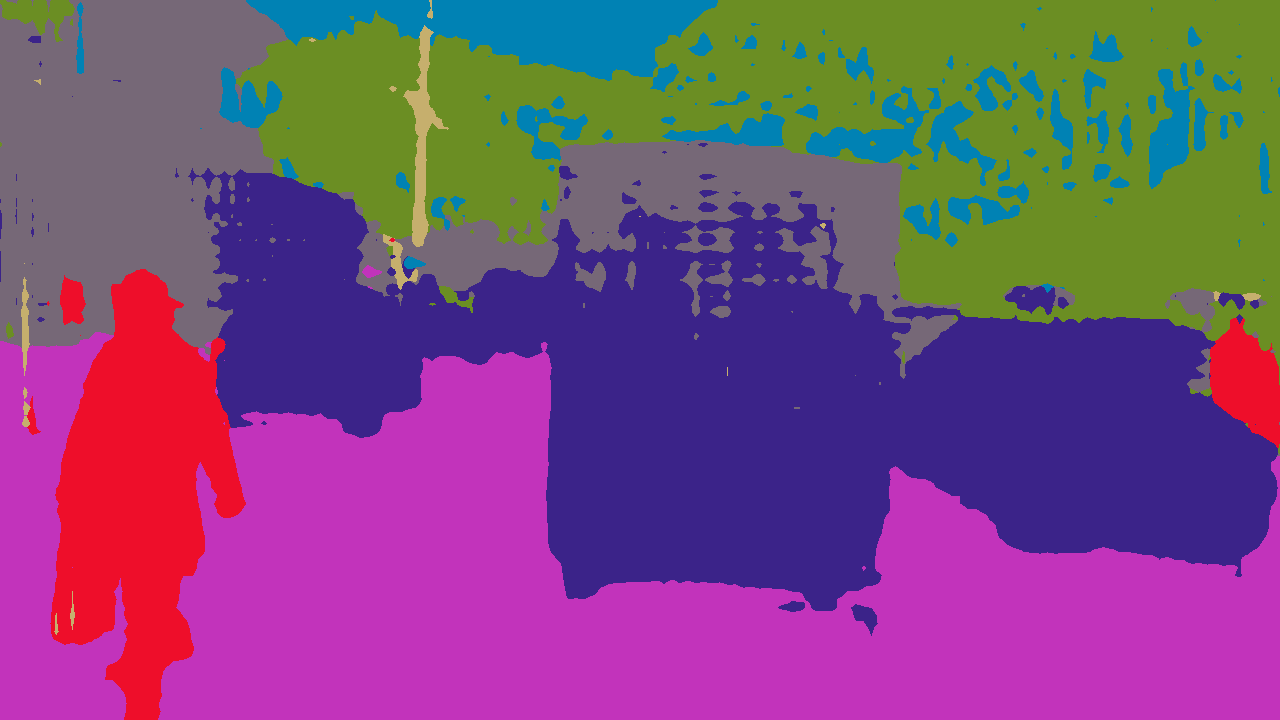}
\end{subfigure}%
\begin{subfigure}{\imgWidth}
\includegraphics[width=\textwidth]{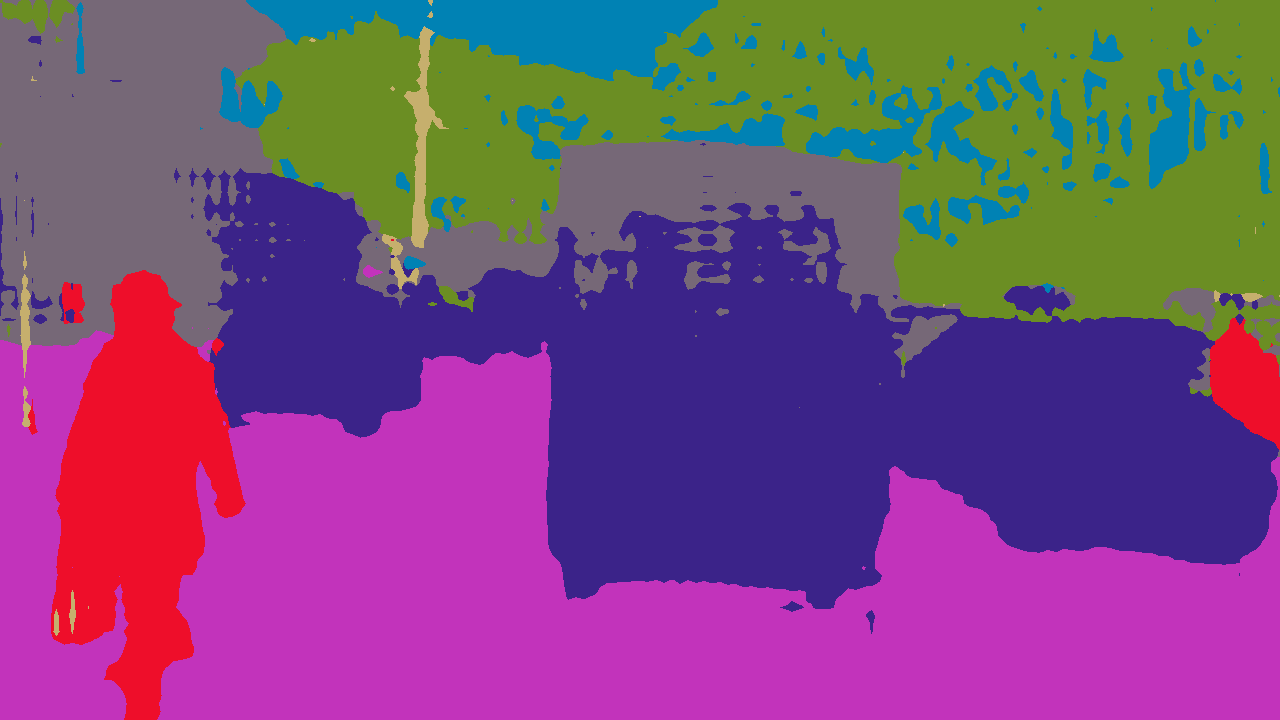}
\end{subfigure}%
\begin{subfigure}{\imgWidth}
\includegraphics[width=\textwidth]{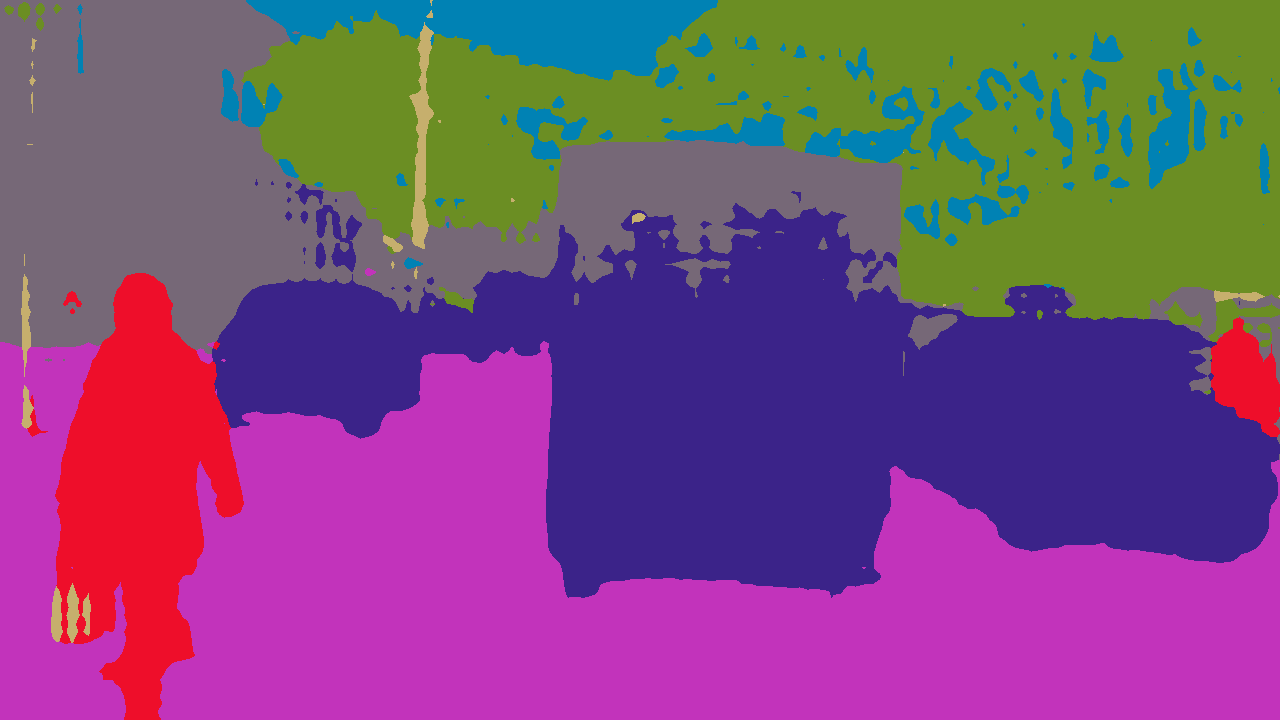}
\end{subfigure}%
\begin{subfigure}{\imgWidth}
\includegraphics[width=\textwidth]{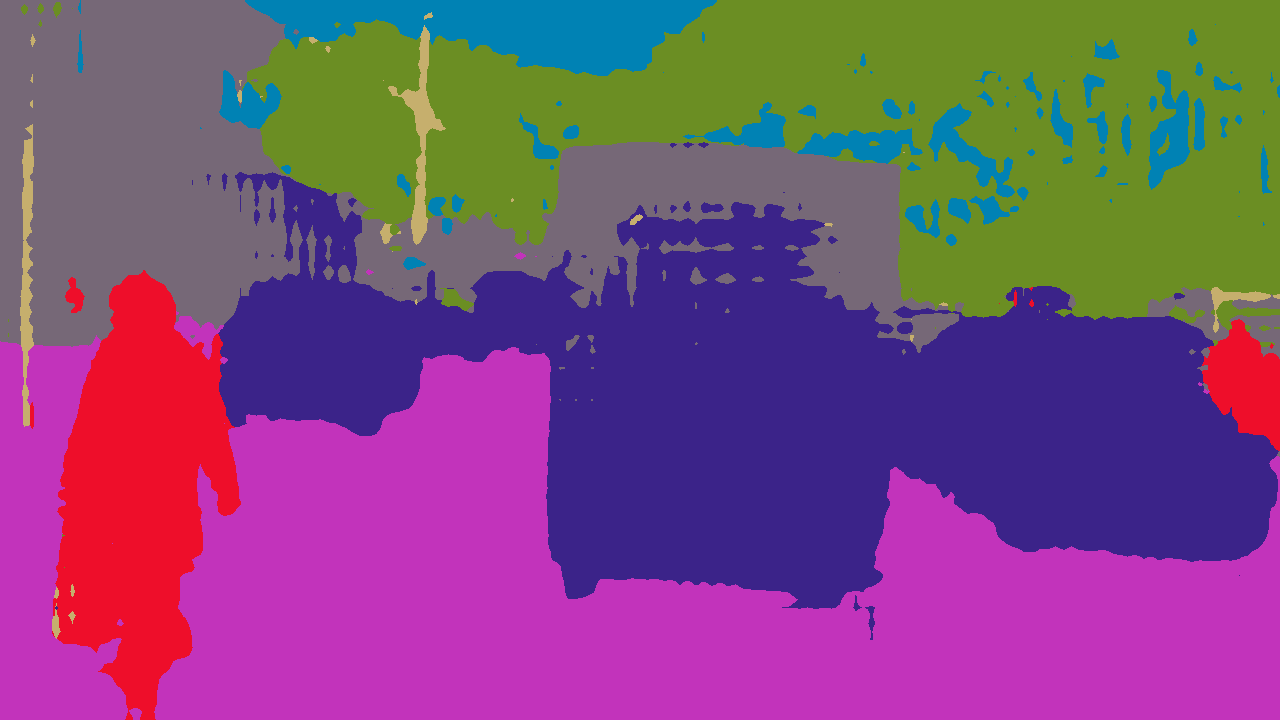}
\end{subfigure}%
\begin{subfigure}{\imgWidth}
\includegraphics[width=\textwidth]{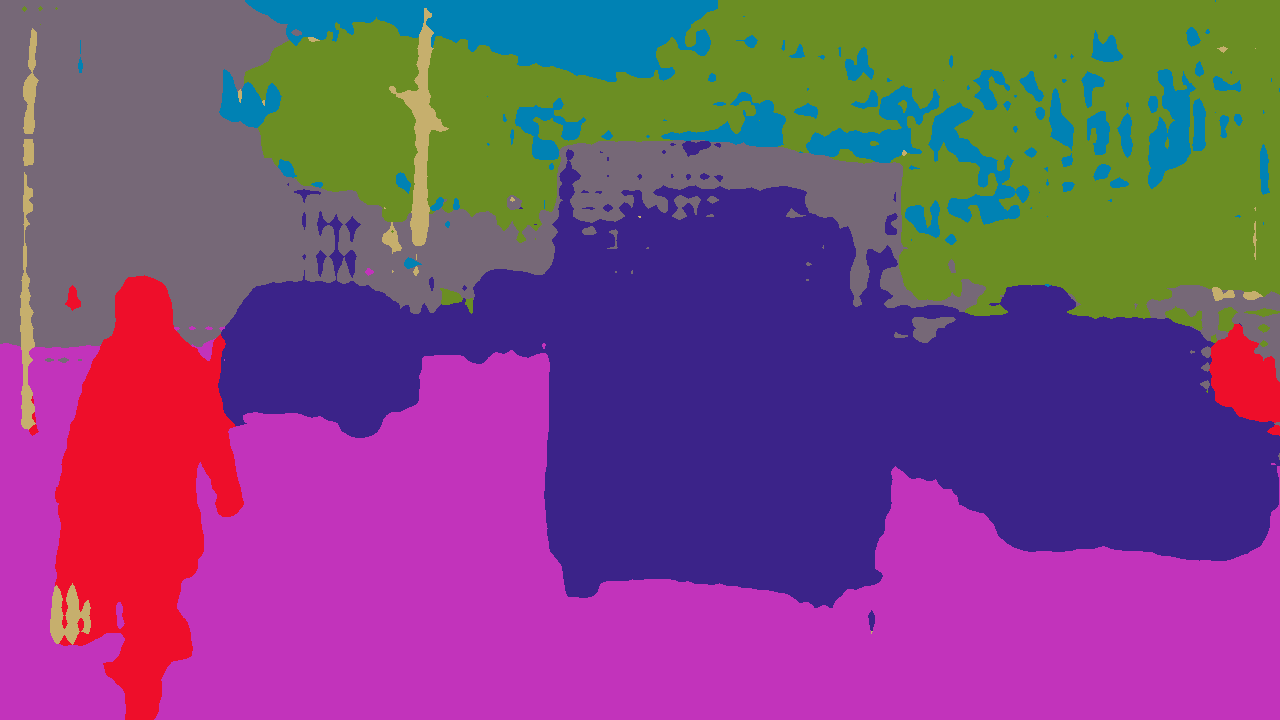}
\end{subfigure}%
\end{subfigure}

\begin{subfigure}{\textwidth}
    \tiny
    \begin{tabularx}{\textwidth}{YYYYYYY}
    \cellcolor{pavement} \textcolor{white}{pavement} & \cellcolor{sky} \textcolor{white}{sky} & \cellcolor{ground_m} \textcolor{black}{ground} & \cellcolor{thinobject} \textcolor{black}{thin ob.} & \cellcolor{structure} \textcolor{white}{structure} & \cellcolor{human} \textcolor{white}{human} & \cellcolor{vehicle} \textcolor{white}{vehicle}
    \end{tabularx}
\end{subfigure}

\begin{subfigure}{.6em}
\scriptsize\rotatebox{90}{Step 2}
\end{subfigure}%
\begin{subfigure}{\textwidth}
\hspace*{.2em}%
\begin{subfigure}{\imgWidth}
\includegraphics[width=\textwidth]{figures/qualitative_gta_idd/rgb.jpeg}
\end{subfigure}%
\begin{subfigure}{\imgWidth}
\includegraphics[width=\textwidth]{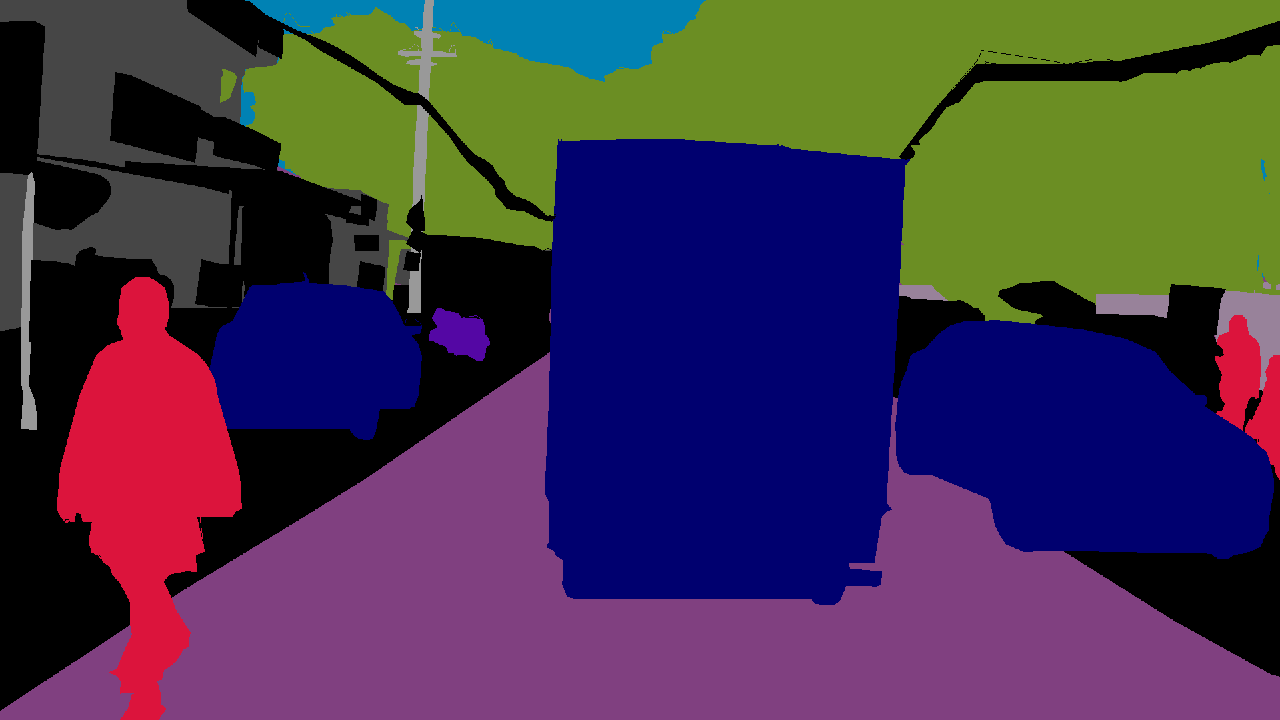}
\end{subfigure}%
\begin{subfigure}{\imgWidth}
\includegraphics[width=\textwidth]{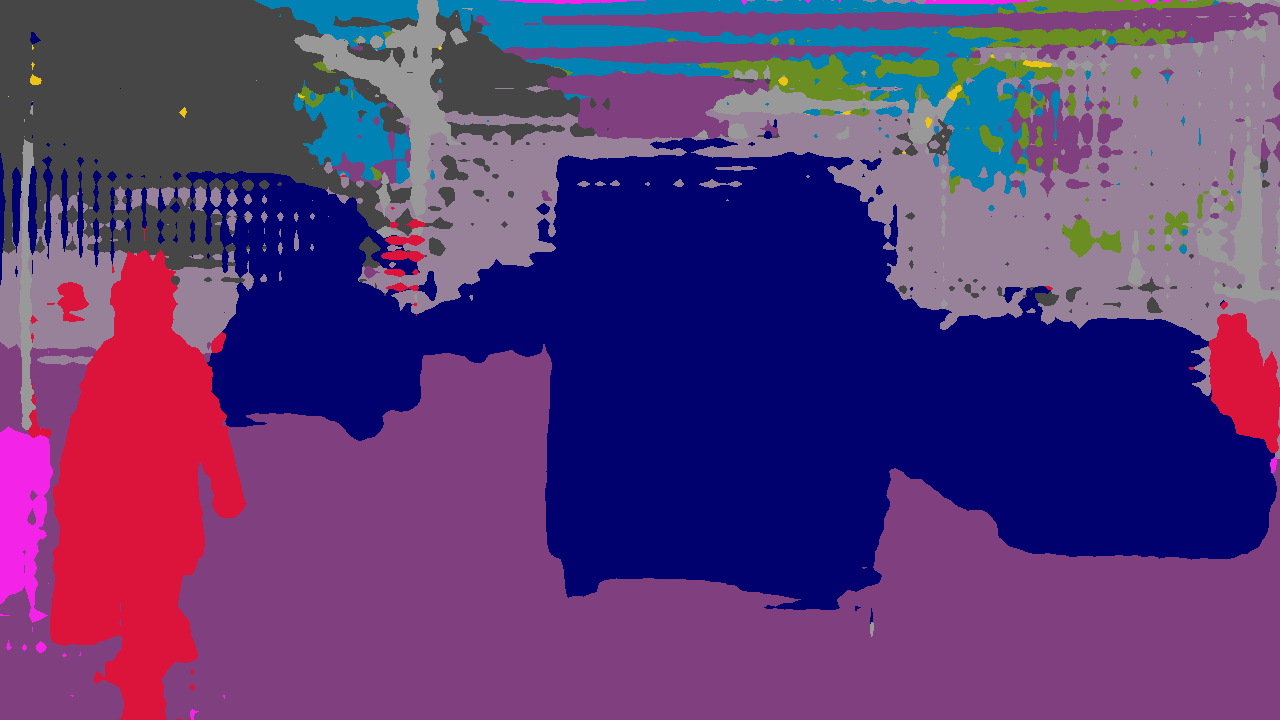}
\end{subfigure}%
\begin{subfigure}{\imgWidth}
\includegraphics[width=\textwidth]{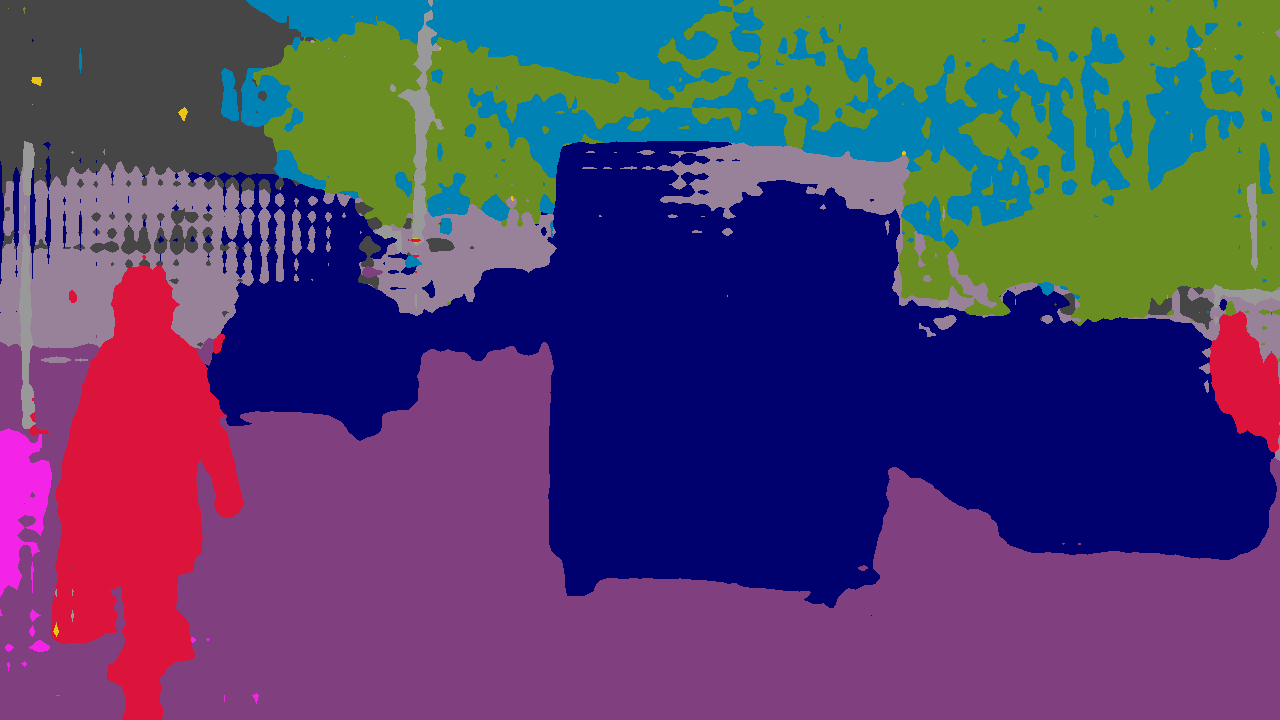}
\end{subfigure}%
\begin{subfigure}{\imgWidth}
\includegraphics[width=\textwidth]{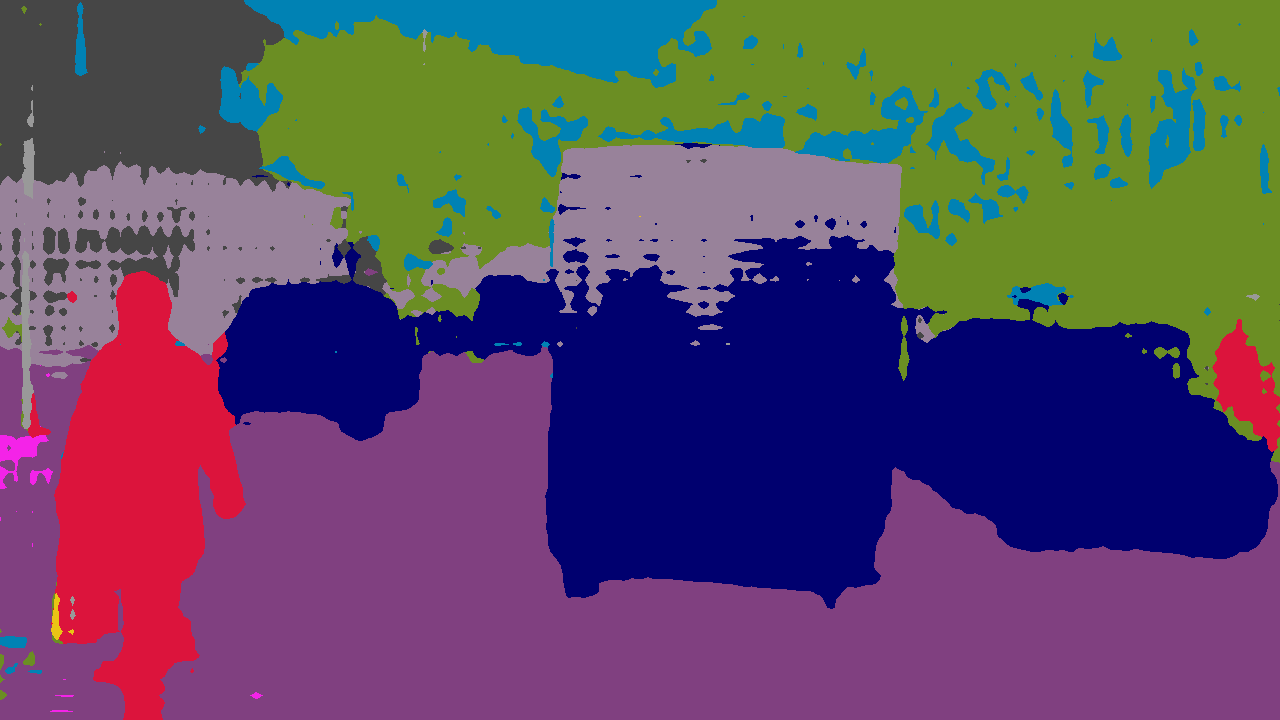}
\end{subfigure}%
\begin{subfigure}{\imgWidth}
\includegraphics[width=\textwidth]{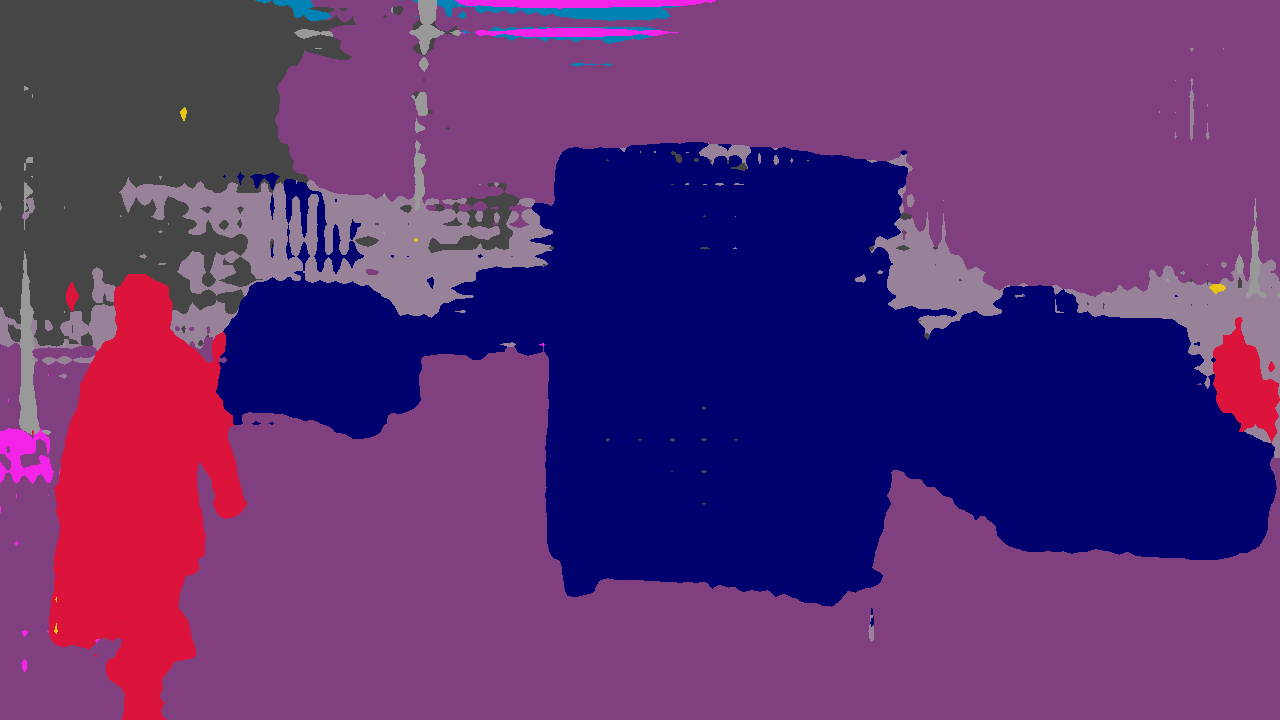}
\end{subfigure}%
\begin{subfigure}{\imgWidth}
\includegraphics[width=\textwidth]{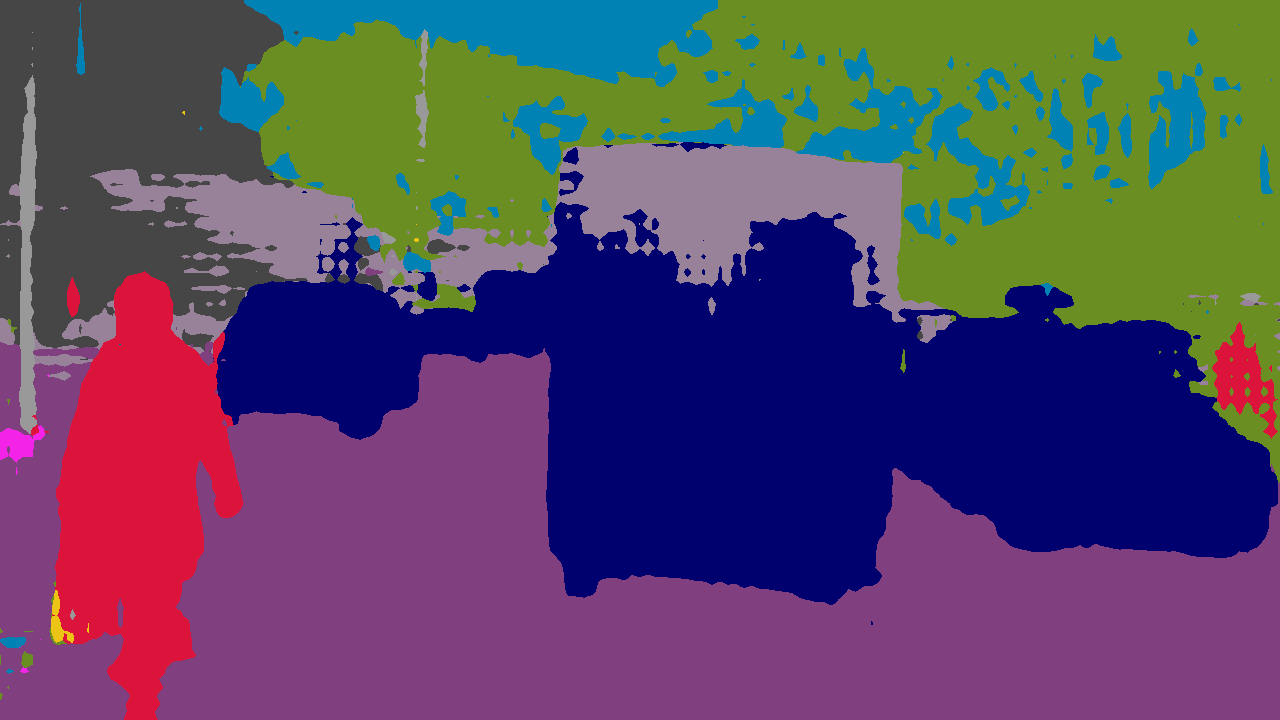}
\end{subfigure}%
\end{subfigure}

\begin{subfigure}{\textwidth}
    \tiny
    \begin{tabularx}{\textwidth}{YYYYYYYYY}
    \cellcolor{road} \textcolor{white}{road} & \cellcolor{sidewalk} \textcolor{white}{sidewalk} & \cellcolor{vegetation} \textcolor{white}{vegetation} &
    \cellcolor{pole} \textcolor{white}{pole} & \cellcolor{signage} \textcolor{black}{signage} & \cellcolor{building} \textcolor{white}{building} 
\end{tabularx}
\begin{tabularx}{\textwidth}{YYYYYYY}
        \cellcolor{barrier} \textcolor{white}{barrier} & \cellcolor{person} \textcolor{white}{person} & \cellcolor{rider} \textcolor{white}{rider} &
        \cellcolor{twowheels} \textcolor{white}{two wheels} & \cellcolor{personaltransport} \textcolor{white}{per. tr.} & \cellcolor{publictransport} \textcolor{white}{pub. tr.} 
\end{tabularx}
\end{subfigure}

\begin{subfigure}{.6em}
\scriptsize\rotatebox{90}{Step 3}
\end{subfigure}%
\begin{subfigure}{\textwidth}
\hspace*{.2em}%
\begin{subfigure}{\imgWidth}
\includegraphics[width=\textwidth]{figures/qualitative_gta_idd/rgb.jpeg}
\end{subfigure}%
\begin{subfigure}{\imgWidth}
\includegraphics[width=\textwidth]{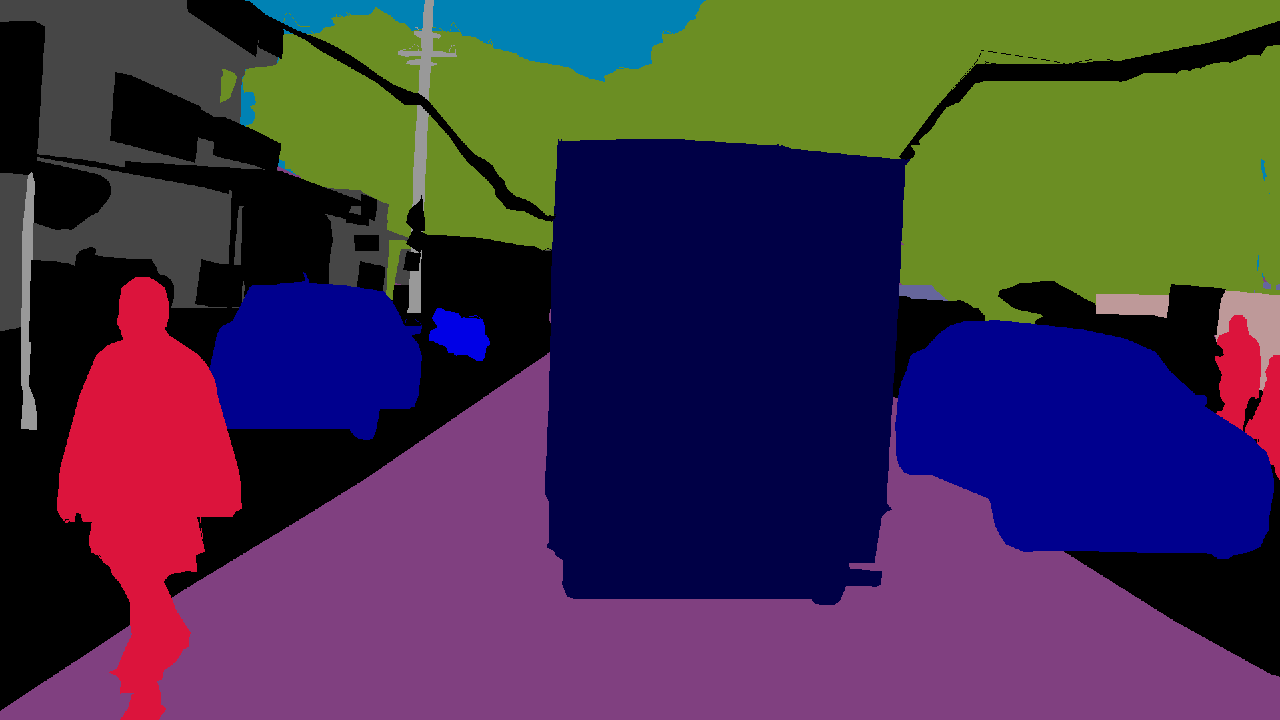}
\end{subfigure}%
\begin{subfigure}{\imgWidth}
\includegraphics[width=\textwidth]{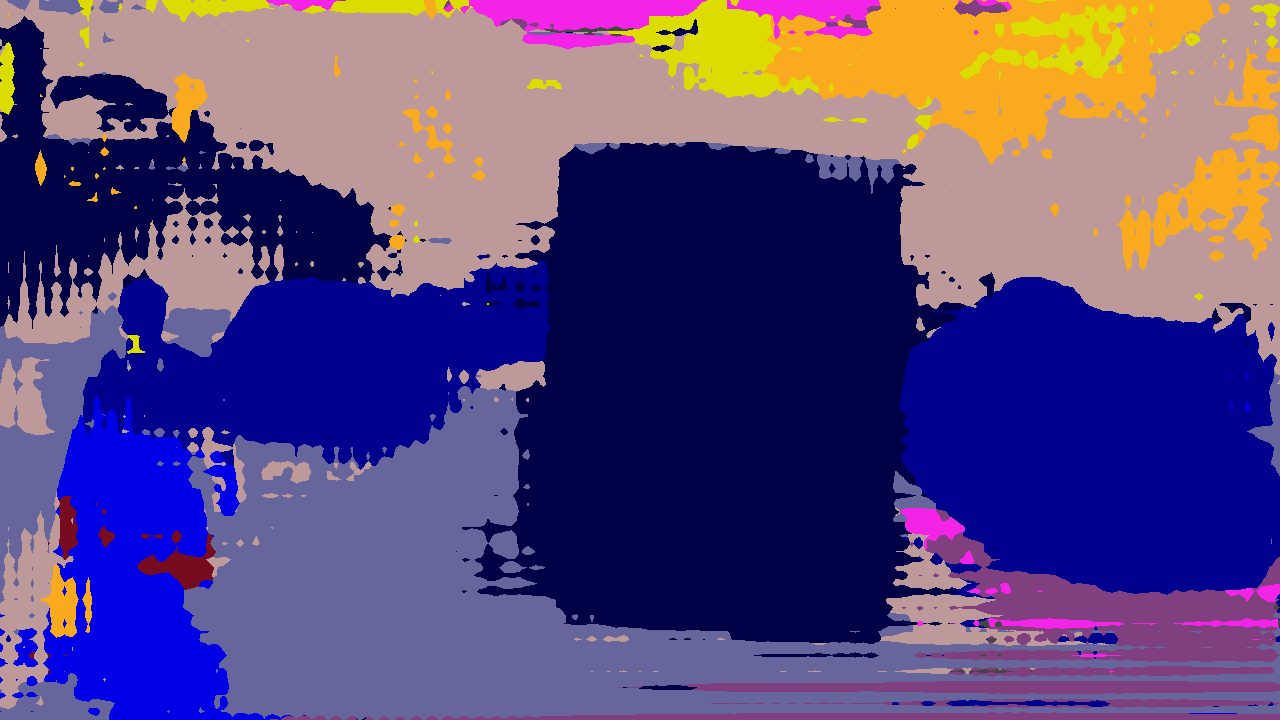}
\end{subfigure}%
\begin{subfigure}{\imgWidth}
\includegraphics[width=\textwidth]{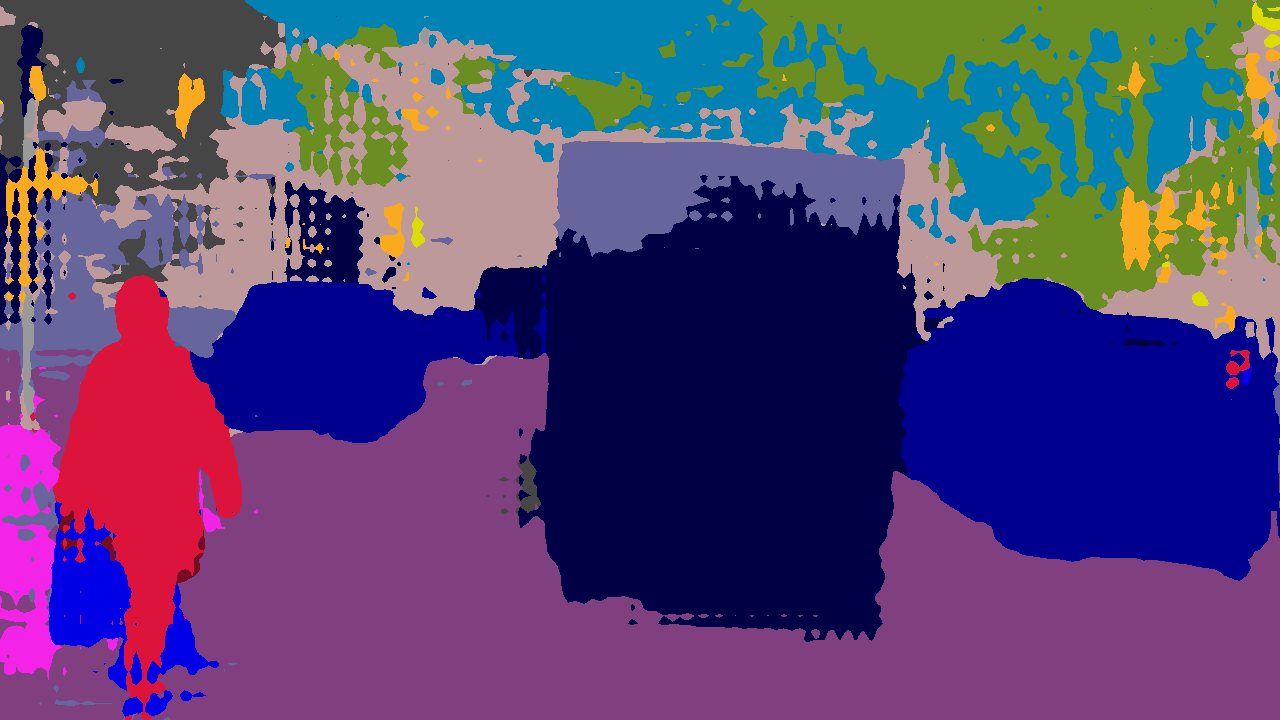}
\end{subfigure}%
\begin{subfigure}{\imgWidth}
\includegraphics[width=\textwidth]{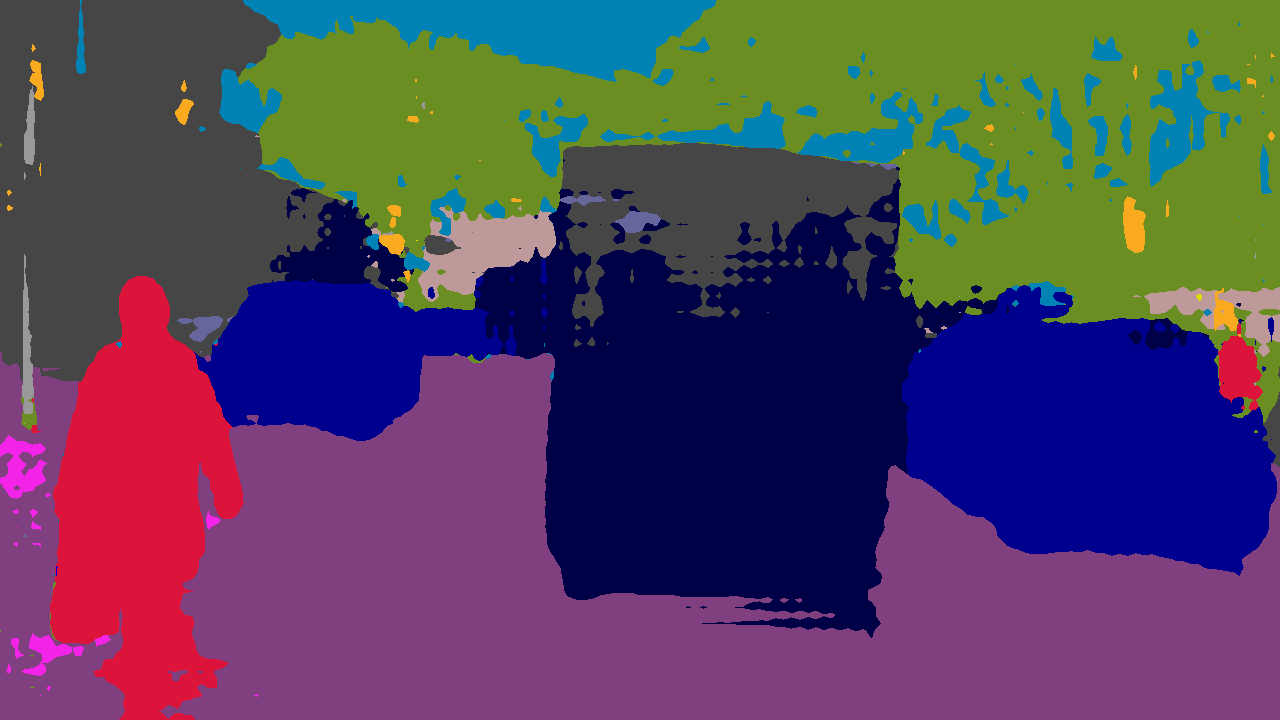}
\end{subfigure}%
\begin{subfigure}{\imgWidth}
\includegraphics[width=\textwidth]{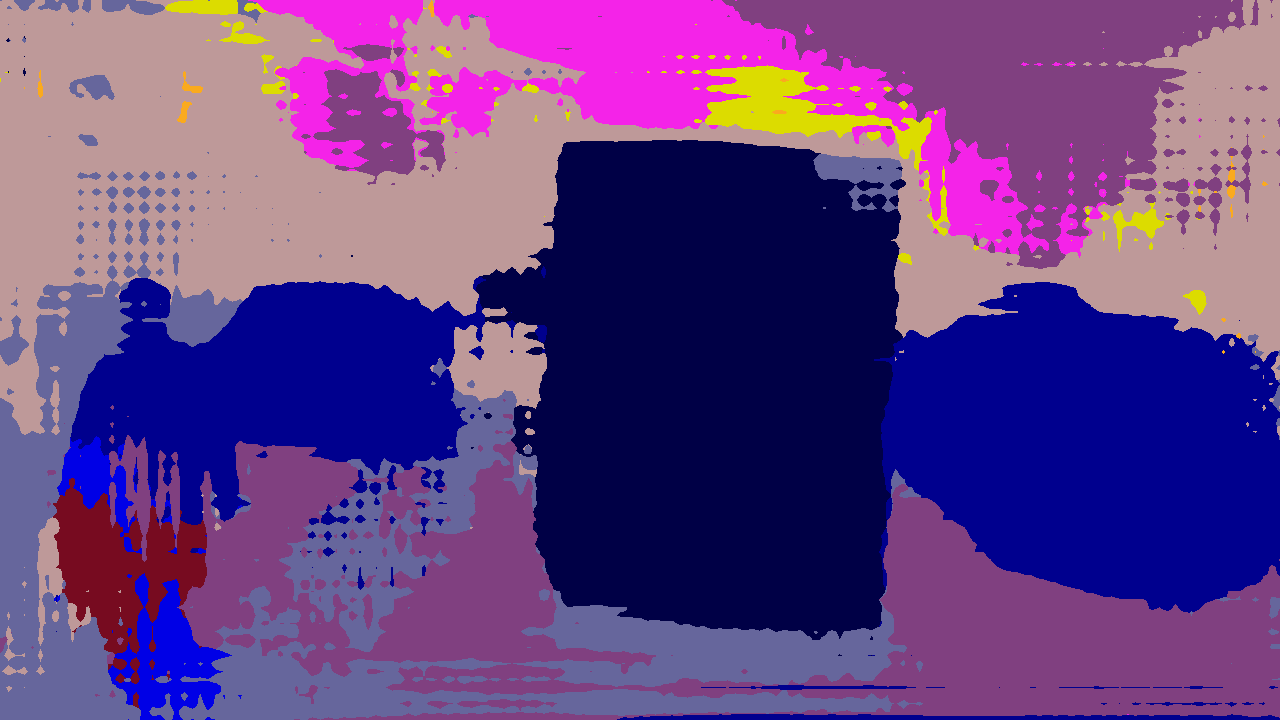}
\end{subfigure}%
\begin{subfigure}{\imgWidth}
\includegraphics[width=\textwidth]{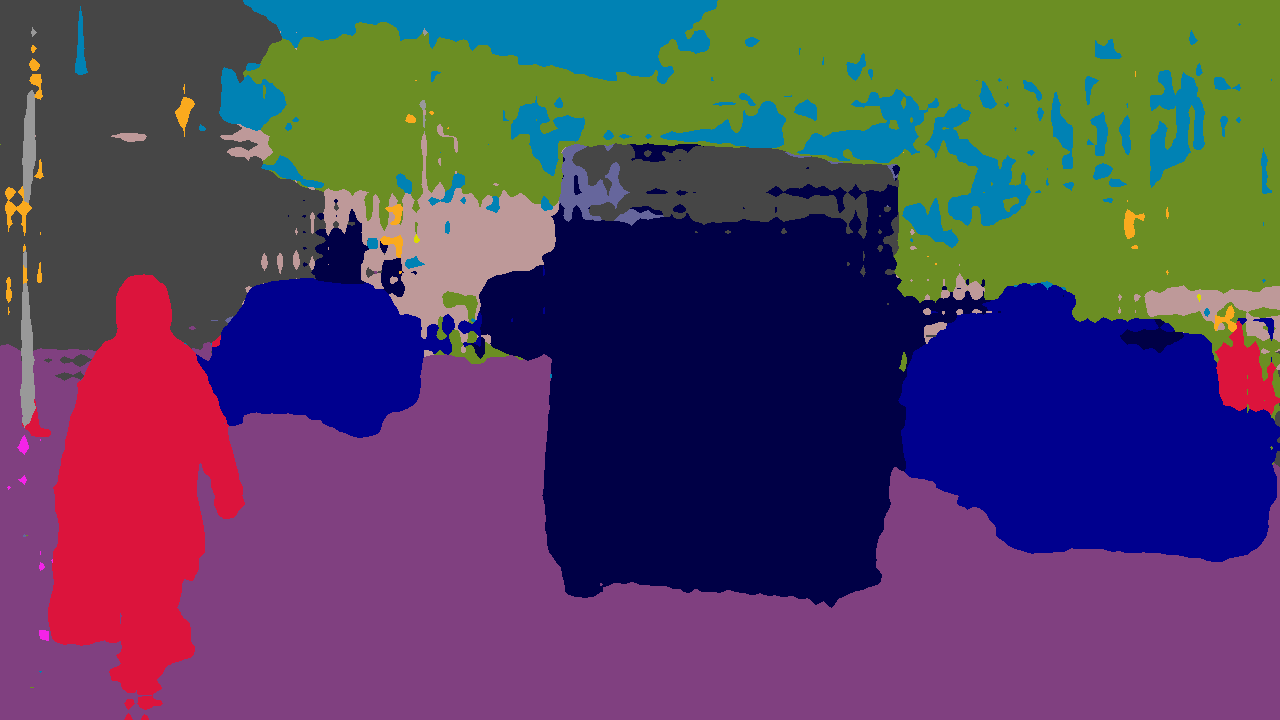}
\end{subfigure}%
\end{subfigure}
\begin{subfigure}{\textwidth}
    \tiny
    \begin{tabularx}{\textwidth}{YYYYYYYYYY}
    \cellcolor{tlight} \textcolor{black}{t. light} & \cellcolor{tsign} \textcolor{black}{t. sign} & \cellcolor{fence} \textcolor{black}{fence} & 
    \cellcolor{wall} \textcolor{white}{wall} & \cellcolor{bicycle} \textcolor{white}{bicycle} & 
    \cellcolor{motorbike} \textcolor{white}{mbike} & \cellcolor{car} \textcolor{white}{car} & \cellcolor{truck} \textcolor{white}{truck} &
    \cellcolor{bus} \textcolor{white}{bus}
    \end{tabularx}
\end{subfigure}
\caption{GTA5$\rightarrow$IDD qualitative results. GT refers to the complete ground truth for the given step.}
\label{fig:quali_idd}
\end{figure}

\begin{table}[t]
    \centering
    \resizebox{\textwidth}{!}{%
    \begin{tabular}{c||cccc|cccccc|ccccccc|c}
    \rotatebox{45}{step} & \rotatebox{90}{road} & \rotatebox{90}{sidewalk} & \rotatebox{90}{sky} & \rotatebox{90}{vegetation}  & \rotatebox{90}{pole} & \rotatebox{90}{traffic light} & \rotatebox{90}{traffic sign} & \rotatebox{90}{building} & \rotatebox{90}{fence} & \rotatebox{90}{wall} & \rotatebox{90}{person} & \rotatebox{90}{rider} & \rotatebox{90}{bicycle} & \rotatebox{90}{motorbike} & \rotatebox{90}{car} & \rotatebox{90}{truck} & \rotatebox{90}{bus} & mIoU \\
    \toprule
    \toprule
    0 & \multicolumn{4}{c|}{93.1} & \multicolumn{6}{c|}{56.8} & \multicolumn{7}{c|}{77.9} & 75.9 \\
    \midrule
    1 & \multicolumn{2}{c}{92.2} & 85.7 & 75.6 & \multicolumn{3}{c}{20.6} & \multicolumn{3}{c|}{51.8}            & \multicolumn{2}{c}{50.6} & \multicolumn{5}{c|}{68.3} & 63.5 \\
    \midrule
    2 & 84.3 & 21.8 & 82.4 & 71.7 & 9.8 & \multicolumn{2}{c}{29.2} & 44.0 & \multicolumn{2}{c|}{21.2} & 21.8 & 18.2 & \multicolumn{2}{c}{31.7} & \multicolumn{2}{c}{62.8} & 24.9 & 40.3  \\
    \midrule
    3 & 78.2 & 16.9 & 79.1 & 70.8 & 7.1 & 2.4 & 35.1 & 25.0 & 5.4 & 19.1 & 19.1 & 2.8 & 14.0 & 39.4 & 74.4 & 43.3 & 28.9 & 33.0 \\
    \toprule
    \end{tabular}%
    }
    \caption{\textit{GTA5}$\rightarrow$\textit{IDD}, CCDA per-class per-step IoU results on the Cityscapes test set.}  
    \label{tab:gta_IDD_quantitative_per_class}
    \end{table}

On the IDD dataset the upper bound given by JTO is $65.9\%$, \ie, $2.7\%$ lower than on Cityscapes (even with two classes less), indicating IDD as a more difficult dataset.  The TNC method achieves a score of $64.2\%$, which is $1.7\%$ lower than JTO, note how such difference is smaller than 
on the Cityscapes dataset  ($1.6\%$ versus $1.3\%$ relative loss, respectively).
Moving to the domain adaptive setup (\textit{GTA5$\rightarrow$IDD})
the Source Only method reaches a mIoU of $5.7\%$, which is increased up to $7.4\%$ by the UDA approach in the MSIW method.
Also in this setup, continual learning strategies have much better performances: MiB reaches $26.8\%$ of mIoU, surpassed by both SKDC ($32.4\%$) and CCDA ($33.0\%$). 
The improvement from MiB to CCDA is similar to the previous case, showing how the performance of the proposed approach are consistent across different datasets ($6.6\%$ for Cityscapes and $6.2\%$ for IDD, respectively).
In this setup the MiB method is the best-performing approach at step $2$. We argue that this is due to the increased number of fine classes in step $1$ due to the collapse of \textit{ground} in \textit{terrain}: this favors the na\"ive knowledge distillation approach of MiB, since there is no coarse-to-fine knowledge transfer. As before, though, the final results show how CCDA is able to preserve much better the knowledge during the third incremental step.
Looking at Table~\ref{tab:gta_IDD_quantitative_per_class} we can see that, even in this setup, one of the causes of decreased performance is the \textit{pole} class, which after being split from \textit{thin object} loses $10.8\%$ mIoU (going from $20.6\%$ to $9.8\%$). Other challenging  classes are \textit{fence} (losing $15.8\%$) and \textit{rider} (with $32.4\%$ loss).

In Figure~\ref{fig:quali_idd} we report some qualitative results for the various methods in this benchmark. Here, we see the significant domain shift present between \textit{GTA5} and \textit{IDD} already from step $0$ (first row): the \textit{truck} (\textit{moving object}) in the foreground is confused for \textit{static object} in all strategies, even when only three classes are considered.
From step $2$ onward, though, the domain shift is overshadowed by the catastrophic forgetting effect: the Source Only and MSIW approaches overfit significantly on the new classes forgetting the old ones, up to the point that even the \textit{sky} is confused as \textit{road}.
On the other hand, as for the \textit{GTA5$\rightarrow$Cityscapes} setting, our  approach (CCDA) is consistently the best approach, maintaining a high precision and tight class borders in all steps.
In particular, when we compare it to the MiB approach, we can see that the latter is overestimating the presence of the class set of step $3$ (last row), for example it misleads part of the  legs of the \textit{person} on the left with a \textit{motorbike}. 

\subsection{Ablation Studies}
\label{subsec:ablation}

In this section we report some ablation results on the {\textit{GTA5}$\rightarrow$\textit{Cityscapes}} benchmark, to better evaluate the components and design choices of our approach and compare its performance to other standard strategies. The section is divided into three parts: additional baseline results are reported in Table~\ref{tab:ablation}, ablation studies on the knowledge distillation loss (see Section~\ref{subsec:c2fkd}) are shown in Table~\ref{tab:kd_ablation}, and ablation studies on the weight initialization rule (see Section~\ref{subsec:c2fuwi}) are finally reported in Table~\ref{tab:weight_ablation}.
\begin{table}[t]
\centering
\setlength\tabcolsep{1.1em}
\renewcommand{\arraystretch}{.8}
\begin{tabular}{ccccc}
\toprule
Model & mIoU\textsubscript{0} & mIoU\textsubscript{1} & mIoU\textsubscript{2} & mIoU\textsubscript{3} \\
\toprule
 Joint Source Only & - & - & - & 30.8 \\
 Joint MaxSquareIW & - & - & - & 38.9 \\
 SNC & 65.0 & 56.2  & 29.0 & 24.8 \\
  \cdashline{1-5}
 TFT & 92.1 & {83.7} & {67.9} & 17.5 \\
 TKDC (ours) & {92.1} & 81.8 & 65.1 & {62.3} \\
 \cdashline{1-5}
  SKDC (ours)  & 65.0 & 65.4 & 34.4 & 30.4 \\
 CCDA (ours)  & {85.4} & {67.9} & 37.2 & {33.1} \\
\bottomrule
\end{tabular}
\caption{Additional baseline experiments on variations of our full approach, conducted on the \textit{GTA5}$\rightarrow$\textit{Cityscapes} setup.}
\label{tab:ablation}
\end{table}

As additional baseline results, in Table~\ref{tab:ablation} we present three UDA-related methods (see Section~\ref{subsec:uda_method}) and two C2F-related methods (see Section~\ref{subsec:c2f_method}). In the first group we placed methods tackling only the domain adaptation task with no incremental learning (\ie, providing them the fully labeled data in a single shot): JSO (Joint Source Only), usually considered as baseline for standard UDA approaches; JMSIW (Joint MSIW), the  upper bound  for our CCDA approach given achieved by applying MSIW on the fully labeled dataset; SNC (Source Non Continual, the same as TNC but with source data instead of the target one), that is upper bound for approaches trained only on source data.\\
In the second group we report: TFT (Target Fine Tuning), baseline for purely continual methods trained  on the target domain, and TKDC, the counterpart of SKDC that is trained on target data. 
From the table we observe how our CCDA strategy achieves a score very close to  JMSIW, with a loss of only $5.8\%$ of mIoU, which is comparable to the performance reduction between JTO and TKDC ($6.3\%$) and the one between JSO and SNC ($6.0\%$).

\begin{table}[t]
\centering
\setlength\tabcolsep{1.1em}
\renewcommand{\arraystretch}{.8}
\begin{tabular}{ccccc}
\toprule
Model & mIoU\textsubscript{0} & mIoU\textsubscript{1} & mIoU\textsubscript{2} & mIoU\textsubscript{3} \\
\toprule
 L1 & \textbf{65.0} & 51.5  & 27.2 & 18.7 \\
 L1 logits & \textbf{65.0} & 61.3  & 31.2 & 23.4 \\
 L2 & \textbf{65.0} & 44.7  & 22.3 & 13.6 \\
 L2 logits &  \textbf{65.0} & 60.6  & 29.2 & 21.6 \\
 MiB \cite{cermelli2020modeling} & \textbf{65.0} & 56.1 & 27.8 & 26.5 \\
 SKDC (ours) & \textbf{65.0} & \textbf{65.4} & \textbf{34.4} & \textbf{30.4} \\
\bottomrule
\end{tabular}
\caption{Ablation experiments on knowledge distillation, performed on the \textit{GTA5}$\rightarrow$\textit{Cityscapes} setup.}
\label{tab:kd_ablation}
\end{table}

In Table~\ref{tab:kd_ablation} we investigate the performance attainable by trying different knowledge distillation strategies as described in detail in Section~\ref{subsec:c2fkd}. We analyze four distance-based expressions and two cross-entropy based ones: as before, all training procedures were performed on the source dataset and tested on the target dataset.
As distance-based strategies we considered the L1 and L2 norms, minimized before and after the \textit{softmax} layer (identified as \textit{logits} in the table). From the experimental results it is clear that matching the logits leads to the best results, gaining $4.7\%$ and $8.0\%$ mIoU in the L1 and L2 versions, respectively.
A better approach, though, is to use cross-entropy based approaches. In this category we show MiB \cite{cermelli2020modeling} and SKDC. The former is reported in the table since it could be considered a simplified case of SKDC where only one coarse class (\ie, the \textit{background}) is considered. This relation is made clear from the experimental results, where MiB loses $3.9\%$ with respect to SKDC due to the missing coarse-to-fine class mapping.

\begin{table}[t]
\centering
\setlength\tabcolsep{0.9em}
\renewcommand{\arraystretch}{.8}
\begin{tabular}{cccccc}
\toprule
Model & Weights Init. & mIoU\textsubscript{0} & mIoU\textsubscript{1} & mIoU\textsubscript{2} & mIoU\textsubscript{3} \\
\toprule
Source Only & Na\"ive & \textbf{65.0} & 56.7 & 25.1 & 4.5 \\ 
Source Only & Unbiased & \textbf{65.0} & 60.4 & 28.2 & 5.5 \\
\cdashline{1-6}
SKDC (ours) & Na\"ive & \textbf{65.0} & 60.1 & 32.6 & 27.9 \\ 
SKDC (ours) & Unbiased & \textbf{65.0} & \textbf{65.4} & \textbf{34.4} & \textbf{30.4} \\

\bottomrule
\end{tabular}
\caption{Ablation experiments on weight initialization, performed on the \textit{GTA5}$\rightarrow$\textit{Cityscapes} setup.}
\label{tab:weight_ablation}
\end{table}

Finally, in Table~\ref{tab:weight_ablation} we investigate the effect of our weight initialization strategy (described in Section~\ref{subsec:c2fuwi}), comparing the score attained by Source Only and SKDC when unbiased weight initialization is enabled or disabled. Recall that the difference lies in whether the bias is rescaled to preserve the original probabilities or not, in which case we fall back to the na\"ive coarse-to-fine weight initialization.
In both cases, the unbiased weight initialization helps to improve mIoU since it reduces the class-confusion and helps the knowledge distillation and cross-entropy objectives.
In particular, 
when the refined weight initialization rule is applied to the SKDC method the performance increases by $2.5\%$.

\section{Conclusion}

In this paper we introduced the novel task of continual coarse-to-fine unsupervised domain adaptation for semantic segmentation, which represents a step closer towards open-world learning. We started by defining in detail the task and we analyzed how it combines the continual learning and UDA tasks. Then, we proposed a new approach (CCDA) to address it. Our approach combines state-of-the-art UDA and continual learning strategies with specific provisions for the new challenging setting, including a novel knowledge distillation strategy and an unbiased weight initialization scheme for the incremental steps. 
We evaluated CCDA in two synthetic-to-real UDA benchmarks, extending them for the use in a continual coarse-to-fine setting. We compare our approach with several methods, including state-of-the-art UDA and CL techniques, outperforming the competitors on both benchmarks.
Further research will be devoted to the introduction of novel coarse-to-fine strategies and to the extension to different experimental scenarios with multiple source and target domains.

\section*{Acknowledgement}
This work was partially supported by the University of Padova Strategic Research Infrastructure Grant 2017: \textit{``CAPRI: Calcolo ad Alte Prestazioni per la Ricerca e l'Innovazione''} and by the  SID project \textit{``Semantic Segmentation in the Wild''}.

\bibliography{strings,refs}

\begin{thebibliography}{10}
\expandafter\ifx\csname url\endcsname\relax
  \def\url#1{\texttt{#1}}\fi
\expandafter\ifx\csname urlprefix\endcsname\relax\def\urlprefix{URL }\fi
\expandafter\ifx\csname href\endcsname\relax
  \def\href#1#2{#2} \def\path#1{#1}\fi

\bibitem{long2015}
J.~Long, E.~Shelhamer, T.~Darrell, Fully convolutional networks for semantic
  segmentation, in: Proceedings of the IEEE Conference on Computer Vision and
  Pattern Recognition, 2015, pp. 3431--3440.

\bibitem{toldo2020unsupervised}
M.~Toldo, A.~Maracani, U.~Michieli, P.~Zanuttigh, Unsupervised domain
  adaptation in semantic segmentation: a review, Technologies 8~(2) (2020).

\bibitem{michieli2019incremental}
U.~Michieli, P.~Zanuttigh, {Incremental Learning Techniques for Semantic
  Segmentation}, in: Proceedings of the International Conference on Computer
  Vision Workshops, 2019.

\bibitem{Chen2019}
M.~Chen, H.~Xue, D.~Cai, Domain adaptation for semantic segmentation with
  maximum squares loss, in: Proceedings of the International Conference on
  Computer Vision, 2019, pp. 2090--2099.

\bibitem{Richter2016}
S.~R. Richter, V.~Vineet, S.~Roth, V.~Koltun, Playing for data: {G}round truth
  from computer games, in: Proceedings of the European Conference on Computer
  Vision, 2016, pp. 102--118.

\bibitem{Cordts2016}
M.~Cordts, M.~Omran, S.~Ramos, T.~Rehfeld, M.~Enzweiler, R.~Benenson,
  U.~Franke, S.~Roth, B.~Schiele, {The Cityscapes dataset for semantic urban
  scene understanding}, in: Proceedings of the IEEE Conference on Computer
  Vision and Pattern Recognition, 2016, pp. 3213--3223.

\bibitem{varma2019idd}
G.~Varma, A.~Subramanian, A.~Namboodiri, M.~Chandraker, C.~Jawahar, Idd: A
  dataset for exploring problems of autonomous navigation in unconstrained
  environments, in: Proceedings of the Winter Conference on Applications of
  Computer Vision, 2019, pp. 1743--1751.

\bibitem{badrinarayanan2017segnet}
V.~Badrinarayanan, A.~Kendall, R.~Cipolla, Segnet: A deep convolutional
  encoder-decoder architecture for image segmentation, IEEE Transactions on
  Pattern Analysis and Machine Intelligence 39~(12) (2017) 2481--2495.

\bibitem{zhao2017}
H.~Zhao, J.~Shi, X.~Qi, X.~Wang, J.~Jia, Pyramid scene parsing network, in:
  Proceedings of the IEEE Conference on Computer Vision and Pattern
  Recognition, 2017, pp. 2881--2890.

\bibitem{chen2017rethinking}
L.~Chen, G.~Papandreou, F.~Schroff, H.~Adam, Rethinking atrous convolution for
  semantic image segmentation, arXiv preprint arXiv:1706.05587 (2017).

\bibitem{chen2018deeplab}
L.-C. Chen, G.~Papandreou, I.~Kokkinos, K.~Murphy, A.~L. Yuille, Deeplab:
  Semantic image segmentation with deep convolutional nets, atrous convolution,
  and fully connected crfs, IEEE Transactions on Pattern Analysis and Machine
  Intelligence 40 (2018) 834--848.

\bibitem{chen2018encoder}
L.~Chen, Y.~Zhu, G.~Papandreou, F.~Schroff, H.~Adam, Encoder-decoder with
  atrous separable convolution for semantic image segmentation, in: Proceedings
  of the European Conference on Computer Vision, 2018, pp. 833--851.

\bibitem{li2021prototypical}
J.~Li, P.~Zhou, C.~Xiong, S.~C. Hoi, Prototypical contrastive learning of
  unsupervised representations, arXiv preprint arXiv:2005.04966 (2020).

\bibitem{yang2021part}
B.~Yang, F.~Wan, C.~Liu, B.~Li, X.~Ji, Q.~Ye, Part-based semantic transform for
  few-shot semantic segmentation, IEEE Transactions on Neural Networks and
  Learning Systems (2021).

\bibitem{mel2020incremental}
M.~Mel, U.~Michieli, P.~Zanuttigh, Incremental and multi-task learning
  strategies for coarse-to-fine semantic segmentation, Technologies 8~(1)
  (2020) 1.

\bibitem{stretcu2020coarse}
O.~Stretcu, E.~A. Platanios, T.~Mitchell, B.~P{\'o}czos, Coarse-to-fine
  curriculum learning for classification, in: Proceedings of the International
  Conference on Learning Representations Workshops, 2020.

\bibitem{stretcu2021coarse}
O.~Stretcu, E.~A. Platanios, T.~M. Mitchell, B.~P{\'o}czos, Coarse-to-fine
  curriculum learning, arXiv preprint arXiv:2106.04072 (2021).

\bibitem{michieli2020gmnet}
U.~Michieli, E.~Borsato, L.~Rossi, P.~Zanuttigh, Gmnet: Graph matching network
  for large scale part semantic segmentation in the wild, in: Proceedings of
  the European Conference on Computer Vision, Springer, 2020, pp. 397--414.

\bibitem{parisi2019continual}
G.~I. Parisi, R.~Kemker, J.~L. Part, C.~Kanan, S.~Wermter, Continual lifelong
  learning with neural networks: A review, Neural Networks (2019).

\bibitem{lesort2019continual}
T.~Lesort, V.~Lomonaco, A.~Stoian, D.~Maltoni, D.~Filliat,
  N.~D{\'\i}az-Rodr{\'\i}guez, Continual learning for robotics: Definition,
  framework, learning strategies, opportunities and challenges, Information
  Fusion 58 (2020) 52--68.

\bibitem{michieli2021unsupervised}
U.~Michieli, M.~Toldo, P.~Zanuttigh, {Unsupervised Domain Adaptation and
  Continual Learning in Semantic Segmentation}, Advanced Methods and Deep
  Learning in Computer Vision, Elsevier (2021).

\bibitem{klingner2020class}
M.~Klingner, A.~B{\"a}r, P.~Donn, T.~Fingscheidt, Class-incremental learning
  for semantic segmentation re-using neither old data nor old labels,
  International Conference on Intelligent Transportation Systems (2020).

\bibitem{michieli2021continual}
U.~Michieli, P.~Zanuttigh, Continual semantic segmentation via
  repulsion-attraction of sparse and disentangled latent representations, in:
  Proceedings of the IEEE Conference on Computer Vision and Pattern
  Recognition, 2021, pp. 1114--1124.

\bibitem{cermelli2020modeling}
F.~Cermelli, M.~Mancini, S.~R. Bul{\`o}, E.~Ricci, B.~Caputo, Modeling the
  background for incremental learning in semantic segmentation, in: Proceedings
  of the IEEE Conference on Computer Vision and Pattern Recognition, 2020, pp.
  9233--9242.

\bibitem{douillard2021plop}
A.~Douillard, Y.~Chen, A.~Dapogny, M.~Cord, Plop: Learning without forgetting
  for continual semantic segmentation, in: Proceedings of the IEEE Conference
  on Computer Vision and Pattern Recognition, 2021, pp. 4040--4050.

\bibitem{maracani2021recall}
A.~Maracani, U.~Michieli, M.~Toldo, P.~Zanuttigh, Recall: Replay-based
  continual learning in semantic segmentation, in: Proceedings of the
  International Conference on Computer Vision, 2021, pp. 7026--7035.

\bibitem{csurka2017domain}
G.~Csurka, Domain adaptation for visual applications: A comprehensive survey,
  arXiv preprint arXiv:1702.05374 (2017).

\bibitem{chen2019crdoco}
Y.-C. Chen, Y.-Y. Lin, M.-H. Yang, J.-B. Huang, Crdoco: Pixel-level domain
  transfer with cross-domain consistency, in: Proceedings of the IEEE
  Conference on Computer Vision and Pattern Recognition, 2019, pp. 1791--1800.

\bibitem{hoffman2018}
J.~Hoffman, E.~Tzeng, T.~Park, J.-Y. Zhu, P.~Isola, K.~Saenko, A.~Efros,
  T.~Darrell, Cycada: Cycle-consistent adversarial domain adaptation, in:
  Proceedings of the International Conference on Machine Learning, 2018, pp.
  1994--2003.

\bibitem{hoffman2016}
J.~Hoffman, D.~Wang, F.~Yu, T.~Darrell, {FCNs in the wild: Pixel-level
  adversarial and constraint-based adaptation}, arXiv preprint arXiv:1612.02649
  (2016).

\bibitem{MurezKKRK18}
Z.~Murez, S.~Kolouri, D.~J. Kriegman, R.~Ramamoorthi, K.~Kim, Image to image
  translation for domain adaptation, in: Proceedings of the IEEE Conference on
  Computer Vision and Pattern Recognition, 2018, pp. 4500--4509.

\bibitem{toldo2020}
M.~Toldo, U.~Michieli, G.~Agresti, P.~Zanuttigh, Unsupervised domain adaptation
  for mobile semantic segmentation based on cycle consistency and feature
  alignment, Image and Vision Computing 95 (2020).

\bibitem{pizzati2020domain}
F.~Pizzati, R.~d. Charette, M.~Zaccaria, P.~Cerri, Domain bridge for unpaired
  image-to-image translation and unsupervised domain adaptation, in:
  Proceedings of the Winter Conference on Applications of Computer Vision,
  2020, pp. 2990--2998.

\bibitem{tranheden2021dacs}
W.~Tranheden, V.~Olsson, J.~Pinto, L.~Svensson, Dacs: Domain adaptation via
  cross-domain mixed sampling, in: Proceedings of the Winter Conference on
  Applications of Computer Vision, 2021, pp. 1379--1389.

\bibitem{toldo2020clustering}
M.~Toldo, U.~Michieli, P.~Zanuttigh, Unsupervised domain adaptation in semantic
  segmentation via orthogonal and clustered embeddings, in: Proceedings of the
  Winter Conference on Applications of Computer Vision, 2021, pp. 1358--1368.

\bibitem{barbato2021latent}
F.~Barbato, M.~Toldo, U.~Michieli, P.~Zanuttigh, Latent space regularization
  for unsupervised domain adaptation in semantic segmentation, in: Proceedings
  of the IEEE Conference on Computer Vision and Pattern Recognition Workshops,
  2021, pp. 2835--2845.

\bibitem{barbato2021road}
F.~Barbato, U.~Michieli, M.~Toldo, P.~Zanuttigh, Road scenes segmentation
  across different domains by disentangling latent representations, arXiv
  preprint arXiv:2108.03021 (2021).

\bibitem{zou2018}
Y.~Zou, Z.~Yu, B.~Vijaya~Kumar, J.~Wang, Unsupervised domain adaptation for
  semantic segmentation via class-balanced self-training, in: Proceedings of
  the European Conference on Computer Vision, 2018, pp. 289--305.

\bibitem{Zou2019}
Y.~Zou, Z.~Yu, X.~Liu, B.~V. Kumar, J.~Wang, Confidence regularized
  self-training, in: Proceedings of the International Conference on Computer
  Vision, 2019, pp. 5982--5991.

\bibitem{zhang2017}
Y.~Zhang, P.~David, B.~Gong, Curriculum domain adaptation for semantic
  segmentation of urban scenes, in: Proceedings of the International Conference
  on Computer Vision, 2017, pp. 2020--2030.

\bibitem{zhang2020curriculum}
Y.~Zhang, P.~David, H.~Foroosh, B.~Gong, A curriculum domain adaptation
  approach to the semantic segmentation of urban scenes, IEEE Transactions on
  Pattern Analysis and Machine Intelligence (2020).

\bibitem{cardace2021shallow}
A.~Cardace, P.~Z. Ramirez, S.~Salti, L.~Di~Stefano, Shallow features guide
  unsupervised domain adaptation for semantic segmentation at class boundaries,
  in: Proceedings of the Winter Conference on Applications of Computer Vision,
  2022, pp. 1160--1170.

\bibitem{tsai2018}
Y.-H. Tsai, W.-C. Hung, S.~Schulter, K.~Sohn, M.-H. Yang, M.~Chandraker,
  Learning to adapt structured output space for semantic segmentation, in:
  Proceedings of the IEEE Conference on Computer Vision and Pattern
  Recognition, 2018, pp. 7472--7481.

\end{thebibliography}
\end{document}